%% file: iclr2026_conference.tex
\documentclass{article} % For LaTeX2e
\usepackage{iclr2026_conference,times}

\input{math_commands.tex}

\usepackage[utf8]{inputenc} % allow utf-8 input
\usepackage[T1]{fontenc}    % use 8-bit T1 fonts
\usepackage{url}            % simple URL typesetting
\usepackage{booktabs}       % professional-quality tables
\usepackage{amsfonts}       % blackboard math symbols
\usepackage{nicefrac}       % compact symbols for 1/2, etc.
\usepackage{microtype}      % microtypography
\usepackage{smiles}
\usepackage{macros}

\usepackage{multirow}
\usepackage{float}
\usepackage[utf8]{inputenc} % allow utf-8 input
\usepackage[T1]{fontenc}    % use 8-bit T1 fonts
\usepackage{url}            % simple URL typesetting
\usepackage{booktabs}       % professional-quality tables
\usepackage{tabularx}
\usepackage{amsfonts}       % blackboard math symbols
\usepackage{nicefrac}       % compact symbols for 1/2, etc.
\usepackage{microtype}      % microtypography
\usepackage{mathtools}
\usepackage{natbib}
\usepackage[dvipsnames]{xcolor}
\usepackage{url}
\usepackage{amsmath}
\usepackage{amssymb}
\usepackage{amsthm}
\usepackage{graphicx}
\usepackage{nicefrac}
\usepackage{enumitem}
\usepackage{xspace}
\usepackage{pifont}
\usepackage{wrapfig,lipsum,booktabs}

\usepackage{amsfonts}

\usepackage{float}  % In preamble for [H] if needed
\usepackage{booktabs}
\usepackage{xcolor}
\usepackage{caption}
\usepackage{tcolorbox}

\usepackage{wrapfig}
\usepackage{caption}
\usepackage{subcaption}
\usepackage{array}
\definecolor{darker}{rgb}{0.15,0.45,0.4}
\definecolor{green}{HTML}{009B55}
\usepackage[colorlinks, urlcolor=darker, citecolor=darker, linkcolor=darker]{hyperref} 
\usepackage{colortbl}

\usepackage[utf8]{inputenc} % allow utf-8 input
\usepackage[T1]{fontenc}    % use 8-bit T1 fonts
\usepackage{url}            % simple URL typesetting
\usepackage{booktabs}       % professional-quality tables
\usepackage{amsfonts}       % blackboard math symbols
\usepackage{nicefrac}       % compact symbols for 1/2, etc.
\usepackage{microtype}      % microtypography
\usepackage{bbding}
\usepackage{threeparttable}
\usepackage{xspace}
\usepackage[ruled,vlined]{algorithm2e}
\usepackage{macros}
\usepackage{wrapfig}
\usepackage{lipsum}
\usepackage{nicefrac}       % compact symbols for 1/2, etc.
\usepackage{colortbl}         % colors
\usepackage{multicol}
\usepackage{multirow}
\usepackage{enumitem}

\usepackage{listings}

\usepackage{titletoc}
\usepackage{mathrsfs}
\usepackage{upquote}
\usepackage[linewidth=0.5pt,leftmargin=1em,rightmargin=1em,innertopmargin=5pt,
    innerbottommargin=5pt,
    innerrightmargin=5pt,
    innerleftmargin=5pt]{mdframed}
\usepackage[toc,page,header]{appendix}

\usepackage{times}
\usepackage{latexsym}
\usepackage{tabularx}
\usepackage{multirow}
\usepackage{booktabs} % To thicken table lines
\usepackage{amsmath}

\usepackage{inconsolata}

\usepackage[utf8]{inputenc} % allow utf-8 input
\usepackage[T1]{fontenc}    % use 8-bit T1 fonts
\usepackage{url}            % simple URL typesetting
\usepackage{nicefrac}       % compact symbols for 1/2, etc.
\usepackage{microtype}      % microtypography
\usepackage{multicol}
\usepackage{transparent}
\usepackage{array}
\usepackage{bm}
\usepackage{caption}

\definecolor{nblue}{cmyk}{0.95,0.0,0.2,0.2}

\newcommand{\method}{\texttt{MedAgentGym}\xspace}
\newcommand{\ours}{\texttt{Med-Copilot}\xspace}
\newcommand{\huggingface}{\raisebox{-1.5pt}{\includegraphics[height=1em]{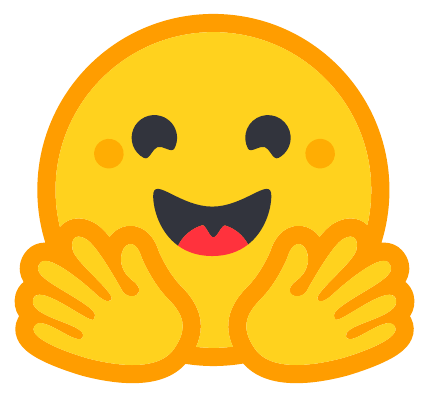}}}
\newcommand{\github}{\raisebox{-1.5pt}{\includegraphics[height=1em]{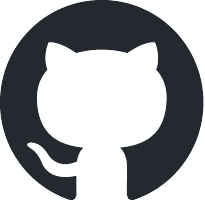}}}

\usepackage{xspace}
\usepackage{colortbl}
\usepackage{listings}
\usepackage{listings}
\usepackage{fancyvrb}
\RecustomVerbatimCommand{\VerbatimInput}{VerbatimInput}{fontsize=\footnotesize,
 frame=single,  
 framesep=0.5em, % separation between frame and text
 labelposition=topline,
}

\lstdefinestyle{prompt}{
  basicstyle=\ttfamily\small,
  breaklines=true,
  frame=none,
  keywordstyle=\bfseries,
  showstringspaces=false,
  literate={~} {$\sim$}{1},
}
\usepackage{tcolorbox}
\tcbuselibrary{listings}

\usepackage{cleveref}
\usepackage{physics}

\theoremstyle{definition}

\theoremstyle{remark}

\numberwithin{equation}{section}

\title{MedAgentGym: A Scalable Agentic Training Environment for Code-Centric Reasoning in Biomedical Data Science}

\author{Ran Xu$^1$\footnotemark[1], Yuchen Zhuang$^2$\thanks{Equal contribution. \dag Correspondence to: \texttt{\small \{Yang.Xie,Wenqi.Shi\}@UTSouthwestern.edu}.}, Yishan Zhong$^2$, Yue Yu$^2$, Zifeng Wang$^3$, Xiangru Tang$^4$,\\
\textbf{Hang Wu$^2$, May D. Wang$^2$, Peifeng Ruan$^5$, Donghan Yang$^5$, Tao Wang$^5$, } \\
\textbf{Guanghua Xiao$^5$, Xin Liu$^6$, Carl Yang$^1$, Yang Xie$^5$\dag, Wenqi Shi$^5$\dag} \\
Emory University$^1$\ Georgia Institute of Technology$^2$ University of Illinois Urbana-Champaign$^3$\\ 
 Yale University$^4$\ UT Southwestern Medical Center$^5$\ University of Washington$^6$\\
\github{} \textbf{\method{}:}\texttt{ \url{https://github.com/wshi83/MedAgentGym}} \\
\huggingface{} \textbf{\method{}:}\texttt{ \url{https://huggingface.co/MedAgentGym}} \\
}

\iclrfinalcopy % Uncomment for camera-ready version, but NOT for submission.
\begin{document}

\maketitle

\input{section/0-abs}
\input{section/1-intro}

\input{section/2-related-works}

\input{section/3-method}

\input{section/4-exp}
\input{section/5-conclusion}

\section*{Ethics Statement}
% We confirm that all authors read and will adhere to the ICLR Code of Ethics. 
This study uses only publicly available or credentialed deidentified datasets (\eg, MIMIC-III and eICU) under their licenses or data use agreements. We do not redistribute data that require credentialed access; instead, we provide scripts to obtain and prepare such data. Licensing and access requirements for all datasets and associated code bases are summarized in Table~\ref{tab:data-access}, and privacy practices are detailed in appendix \ref{app:privacy}. In particular, we followed the PhysioNet Credentialed Health Data Use Agreement for MIMIC-III and eICU and did not transfer any confidential patient data to third-party services. When using Microsoft Azure OpenAI services, we opted out of human review and followed the PhysioNet guidelines for responsible use.

To reduce privacy and security risk, all tasks execute inside isolated Docker containers with pre-installed dependencies, which preserves environment integrity and prevents unintended modification of underlying data. The benchmark verifies solutions by execution outputs rather than raw code, and the released artifacts do not contain protected health information. We emphasize that outputs are research artifacts and must not be used to guide diagnosis or treatment without formal validation and regulatory review. We encourage users of our released resources to conduct additional subgroup and setting-specific audits before any downstream use. We also report compute and token usage, and describe hardware and training footprints, to support transparency about environmental impact.

\section*{Reproducibility Statement}
% \noindent\textbf{Released Artifacts and Environment.} 
We aim for complete reproducibility by providing an anonymized artifact with source code for the benchmark and agents, Dockerfiles and pinned dependency manifests, evaluation harnesses, and scripts that reproduce data preparation for each dataset according to its license. 
The paper specifies the task taxonomy, data sources, and train/test splits, the executable sandbox and interaction interface, and the agent scaffold and action space used across all experiments in \cref{sec:method} and \cref{sec:exp}. Prompt templates for every dataset are included in appendix \ref{app:prompt}; dataset-specific preprocessing and verification logic are documented in appendix \ref{app:data}, including structured JSON formats for inputs and outputs. 

\bibliography{iclr2026_conference}
\bibliographystyle{iclr2026_conference}

\appendix
\input{section/6-appendix}

\end{document}

%% file: math_commands.tex
\usepackage{amsmath,amsfonts,bm}

\def\eqref#1{equation~\ref{#1}}
\def\1{\bm{1}}

\DeclareMathAlphabet{\mathsfit}{\encodingdefault}{\sfdefault}{m}{sl}
\SetMathAlphabet{\mathsfit}{bold}{\encodingdefault}{\sfdefault}{bx}{n}

%% file: section/0-abs.tex
\vspace{-3ex}
\vspace{-0.5ex}
\begin{abstract}
\vspace{-0.5ex}
We introduce \method{}, a scalable and interactive training environment designed to enhance coding-based biomedical reasoning capabilities in large language model (LLM) agents.
\method{} comprises $72,413$ task instances across $129$ categories derived from $12$ authentic real-world biomedical scenarios. 
Tasks are encapsulated within executable sandbox environments, each featuring detailed task specifications, interactive feedback mechanisms, verifiable ground truth annotations, and scalable training trajectory generation.
Extensive benchmarking of $29$ LLMs reveals substantial performance disparities in biomedical data science between commercial and open-source LLMs.
Leveraging efficient multi-threaded and multi-turn trajectory sampling in \method{}, \ours{} achieves performance gains of $+43.02\%$ and $+45.28\%$ from offline and online reinforcement learning, respectively, 
demonstrating \method{} as an effective training ground while establishing itself as a cost-effective, privacy-preserving alternative competitive with proprietary LLMs (\texttt{gpt-4o}).
% Leveraging efficient multi-threaded and multi-turn trajectory sampling in \method{}, we fine-tune a lightweight yet capable biomedical coding agent, \ours{}, through agentic reinforcement learning. 
% Remarkably, \ours{}\texttt{-7B} achieves significant performance gains from both offline ($+43.02\%$) and online ($+45.28\%$) reinforcement learning, demonstrating \method{} as an effective training ground while establishing itself as an affordable, privacy-preserving alternative competitive to commercial LLMs (\texttt{gpt-4o}).
By offering a unified execution environment with a comprehensive benchmark and accessible, extensible training resources, \method{} delivers an integrated platform to develop LLM-based coding assistants for advanced biomedical data science.
\end{abstract}

%% file: section/1-intro.tex
\begin{figure}[h]
	\centering
    \vspace{-3ex}    
    \subfigure[Biomedical coding capabilities with \method{}]{
	\includegraphics[width=0.53\linewidth]{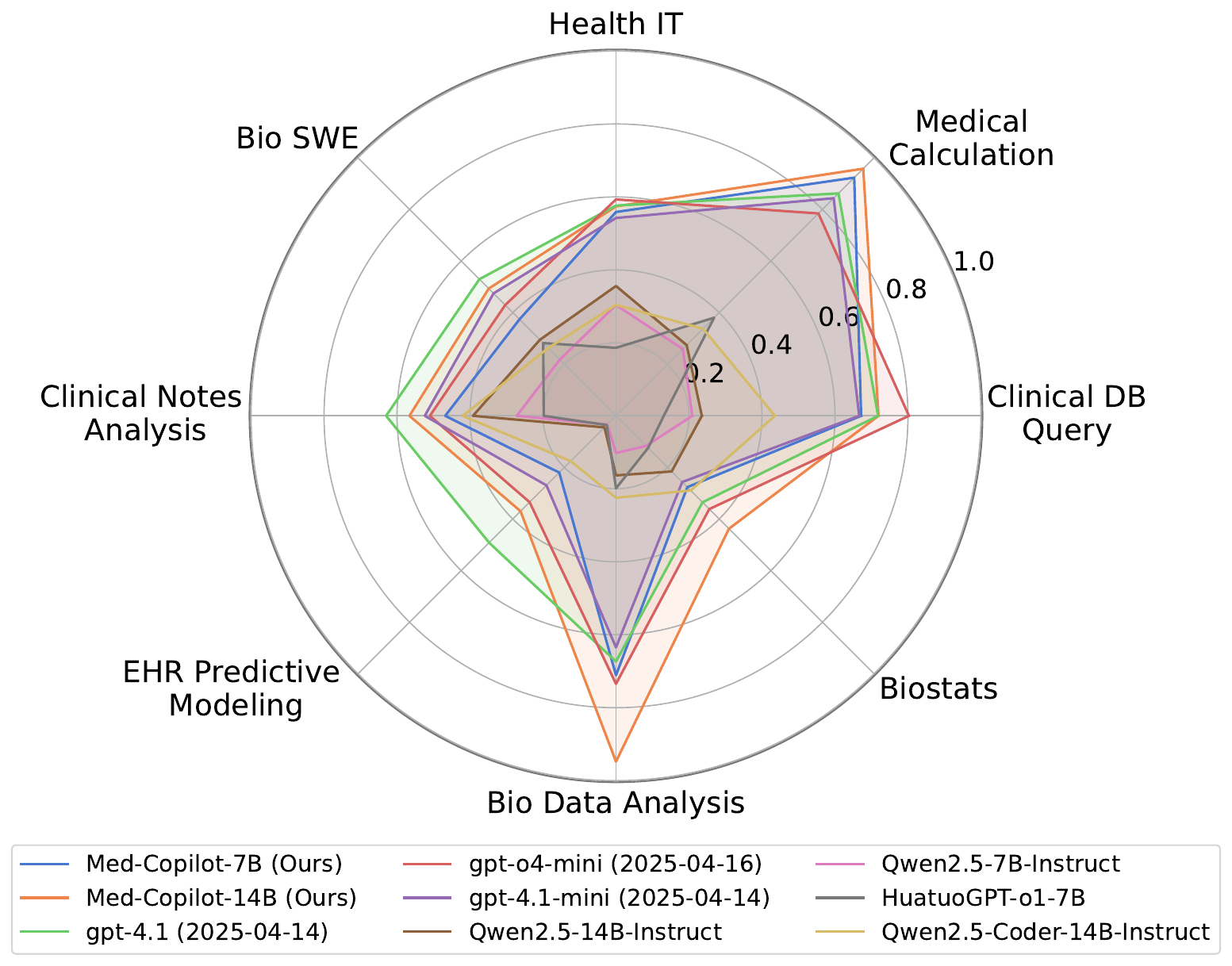}
	\label{fig:radar}
	}
	\subfigure[Overall score of \method{}]{
	\includegraphics[width=0.42\linewidth]{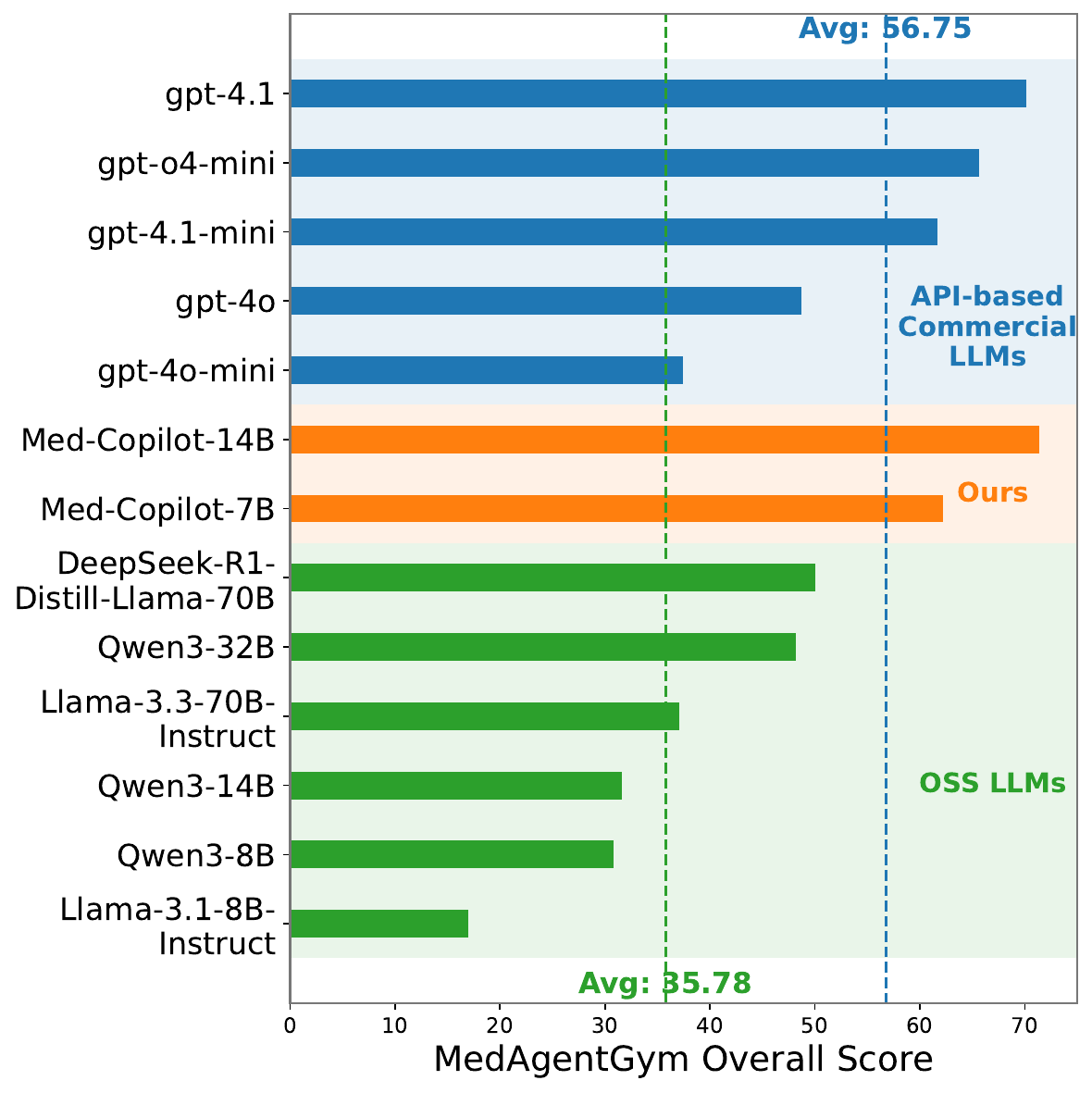}
	\label{fig:leaderboard}
	} 
    \vspace{-2ex}
	\caption{
    Overview of (a) task-specific and (b) overall leaderboard evaluation in \method{}. The results show the (a) performance variations across biomedical data science tasks and (b) large gaps between proprietary and open-source (OSS) LLMs, highlighting the need for continued development of privacy-preserving, affordable LLM agents, especially for complex code-based biomedical reasoning tasks such as biomedical software engineering and predictive modeling. 
    \vspace{-3ex}
    }
\label{fig:teaser}
\end{figure}

% \vspace{-0.5ex}
\section{Introduction}
\label{sec:intro}
% \vspace{-0.5ex}
The exponential growth of healthcare data has fundamentally transformed modern biomedical research, intensifying the need for integration of advanced computational methods with medical domain expertise~\citep{wornow2023shaky,liu2025towards}. 
Biomedical researchers routinely face data science challenges that demand both medical data analysis knowledge and programming proficiency, such as querying large-scale databases, conducting statistical analyses, processing genomic sequences, and building predictive models from electronic health records (EHRs)~\citep{nimmolrat2021patient,lee2022ehrsql,wornow2023ehrshot}. 
While recent advances in large language models (LLMs) have demonstrated significant capabilities in advanced reasoning \citep{o4-mini,guo2025deepseek}, including code generation \citep{alphaevolve} and scientific discovery~\citep{swanson2024virtual,team2025novelseek,yuan2025dolphin}, it remains challenging to translate real-world biomedical data science requirements into executable computational solutions~\citep{wang2024can,wang2025biodsa}.
% Many of these tasks, such as patient triage, risk prediction, and electronic health records (EHRs) querying, increasingly depend on computing systems to accurately process and interpret medical information~\citep{nimmolrat2021patient,lee2022ehrsql,wornow2023ehrshot}.
% However, the combination of deep medical expertise and proficient programming skills remains challenging for many clinicians and biomedical researchers.
% Intelligent systems that automate code generation for medical applications hold substantial promise for bridging this gap and accelerating healthcare AI innovations, yet existing LLMs still fall short of enabling fully automated medical workflows~\citep{wang2024can}.

Developing effective biomedical coding agents poses unique challenges beyond knowledge-intensive medical reasoning~\citep{wang2025medical,wang2025survey} and general-purpose code generation~\citep{zheng2024opencodeinterpreter,jing2025dsbench}.
Within biomedical research and clinical practice, direct deployment of proprietary LLMs remains infeasible due to strict privacy requirements and prohibitive operational costs~\citep{mesko2023imperative,shi2024medadapter}, whereas OSS LLMs exhibit substantial deficiencies in biomedical coding capabilities (Figure~\ref{fig:teaser}).
% Consequently, enhancing the medical coding proficiency of OSS LLMs remains critically important. 
% Medical applications inherently require agents to manage diverse data modalities, such as unstructured clinical notes, structured EHRs, and complex biomedical sequences, while adhering to rigorous standards and regulatory compliance. 
Mitigating this performance disparity calls for addressing two infrastructure gaps: 
(1) \emph{comprehensive, code‑centric biomedical reasoning benchmarks} to diagnose agent limitations and support rigorous, reproducible evaluation; and (2) \emph{specialized, interactive training environments} to develop the complex reasoning and robust coding capabilities required for real‑world biomedical data science.

In this study, we introduce \method{}, a scalable and agentic training environment designed to systematically enhance the coding-centric reasoning capabilities of LLM agents for biomedical data science workflows. Grounded in diverse real-world biomedical scenarios, \method{} provides:
\begin{itemize}[leftmargin=0.6cm]
\item {\textbf{Comprehensive suite of code-centric biomedical reasoning tasks.}}
\method{} encompasses 72,413 biomedical \emph{coding-centric} instances across 129 categories grounded in 12 real-world biomedical scenarios\footnote{We emphasize that \method{} mainly focuses on computational \emph{code generation} for biomedical reasoning, rather than traditional medical coding systems~\citep{soroush2024large} such as ICD-9 or ICD-10.
}. 
We standardize a rich collection of biomedical data science tasks as executable problems with verifiable ground truth, spanning structured medical information retrieval, numerical clinical reasoning, bioinformatics, and machine learning (ML) modeling.
Tasks incorporate diverse data modalities, including EHR tables, clinical notes, genomics, drugs, and biological sequences, which require medical domain-specific reasoning capabilities.
\item {\textbf{Scalable and interactive training infrastructure.}}
\method{} provides an optimized, user-friendly environment to accelerate agent training.
Each instance is encapsulated within \emph{executable, isolated, and reproducible} Docker environments with pre-install dependencies, supporting multi-threading, parallel execution, and sequential sampling.
\method{} ensures efficient trajectory collection and facilitate large-scale automated evaluation compatible with diverse agent scaffolds.
\item {\textbf{Extensive benchmarking and effective agent training for biomedical data science.}}
Through an extensive benchmark of 29 proprietary and open-source LLMs, we identify critical deficiencies in biomedical data analysis and predictive modeling.
\method{} effectively strengthens agentic training: \ours{}-7B achieves gains of +43.02\% and +45.28\% through offline and online reinforcement learning (RL), respectively, and performs comparably to \texttt{gpt-4o} on both in- and out-of-distribution tasks.
We publicly release \method{} and \ours{}, together with high-quality training trajectories and the outcome verifier, to support reproducible benchmarking and continued development of LLM coding agents in biomedical data science.
\end{itemize}

% In summary, \method{} fills a critical gap in developing medical coding agents by offering the first unified, task-rich environment tailored to the demands of healthcare applications. 
% By combining diverse real-world biomedical tasks, an efficient and reproducible infrastructure, and large-scale benchmarking across frontier LLMs, \method{} provides a foundation for advancing LLM coding toward medical applications. 
% We hope \method{} will foster progress in building efficient, reliable, and clinically grounded AI agents to support real-world research and practice in healthcare.

% The remainder of the paper is organized as follows: we review related work in \cref{sec:related}, describe our task curation and environment design in \cref{sec:method}, and present the experimental setup along with benchmark results for \method{} in \cref{sec:exp}. In \cref{sec:medcopilot}, we fine-tune \ours{} to showcase the effectiveness of \method{}. We then conclude the paper in \cref{sec:conclusion}.

%% file: section/2-related-works.tex
% \vspace{-0.5ex}
\section{Related Works}
\label{sec:related}
% \vspace{-0.5ex}

\input{table/tab-related}

\noindent\textbf{Coding-Centric Reasoning in Biomedical Data Science.}
Most existing medical benchmarks primarily evaluate LLMs on knowledge‑intensive, narrative reasoning~\citep{jin-etal-2019-pubmedqa,pal2022medmcqa,tsatsaronis2015overview}. 
Although several efforts target isolated biomedical algorithmic tasks~\citep{tang2024biocoder,medhelm,wang2024can} or simulate portions of clinical workflows~\citep{schmidgall2024agentclinic,li2024mediq,Li2024AgentHospital}, they do not capture a complete set of tasks in the full end‑to‑end lifecycle of biomedical data science, from data extraction~\citep{lee2022ehrsql,ryu2024ehr} to model development~\citep{wornow2023ehrshot,MIMICExtract}. 
Complementing these benchmarks, \method{} emphasizes computation- and coding-intensive tasks that require LLM agents to retrieve, transform, analyze, and compute biomedical data while generating and executing code with pre-installed biomedical libraries and dependencies to produce verifiable solutions.

\noindent\textbf{Scalable and Interactive Training Environment for Biomedical Coding Agents.}
Agentic RL~\citep{guo2025deepseek,schulman2017proximal,shao2024deepseekmath} shifts LLM post-training from passive sequence generation to autonomous agents operating in complex, dynamic settings, including medical reasoning~\citep{xia2025mmedagent,jiang2025meds3medicalsmalllanguage,chen2024huatuogpt,lan2025clinicalgpt,wu2025medreason,wang2025baichuan}. 
Within such a framework, agents interact iteratively with their environment, receiving observations and executing actions, while the environment returns reward signals and state updates~\citep{wang2025ragen,chezelles2024browsergym,shao2024collaborative,nathani2025mlgym}.
However, most biomedical reasoning and data science benchmarks (Table~\ref{tab:related}) are single‑pass evaluations without executable environments or agent‑level interaction signals~\citep{zhu2025medagentboard,healthbench,wu2025medcasereasoning}.
In contrast, \method{} uniquely provides an executable and interactive biomedical coding environment covering comprehensive range of tasks. It also supports efficient multi-turn trajectory sampling through multi‑threaded rollouts, thus enabling scalable and systematic improvement via agentic fine-tuning beyond prompting~\citep{shi2024ehragent,huang2025biomni}.

%% file: table/tab-related.tex
\begin{table}[t]
% \vspace{-3ex}
\caption{Summary of related biomedical reasoning and coding datasets with task details and execution environments.
\method{} is among the first publicly available training environments for LLM agents in biomedical data science, uniquely integrating \emph{executable environments, interactive feedback, and task-isolated run-time facilities} for coding-based reasoning.
``DB'', ``DA'', ``Bioinfo'', and ``ML'' denote ``database'', ``data analytics'', ``bioinformatics'', and ``machine learning'', respectively.}
\label{tab:related}
\vspace{-1ex}
\begin{threeparttable}
\centering 
\renewcommand\arraystretch{0.98}
\resizebox{1.0\linewidth}{!}{ %
\begin{tabular}{@{}l|cc|cccc|cccc|ccc@{}}
\toprule
 & \multicolumn{2}{c|}{\textbf{Domain}} & \multicolumn{4}{c|}{\textbf{Task}} & \multicolumn{4}{c|}{\textbf{Environment \& Facility}} & \multicolumn{3}{c}{\textbf{Scale (\#Instances)}} \\
\cmidrule(lr){2-3} \cmidrule(lr){4-7} \cmidrule(lr){8-11} \cmidrule(lr){12-14} 
\textbf{Datasets ($\downarrow$)} & QA & Coding & DB & DA & Bioinfo & ML &
Execution & Interaction & Isolation & Training  & {\# Train} & {\# Test} & {\# Traj.}\\ \midrule 
MedMCQA~\citep{pal2022medmcqa} & \textcolor{green}{\CheckmarkBold} & \textcolor{red}{\XSolidBrush} & \textcolor{red}{\XSolidBrush} & \textcolor{red}{\XSolidBrush} & \textcolor{red}{\XSolidBrush} & \textcolor{red}{\XSolidBrush} & \textcolor{red}{\XSolidBrush} & \textcolor{red}{\XSolidBrush} & \textcolor{red}{\XSolidBrush} & \textcolor{red}{\XSolidBrush} & 3K & 4.18K & \textcolor{red}{\XSolidBrush} \\
MedQA~\citep{jin2021disease} & \textcolor{green}{\CheckmarkBold} & \textcolor{red}{\XSolidBrush} & \textcolor{red}{\XSolidBrush} & \textcolor{red}{\XSolidBrush} & \textcolor{red}{\XSolidBrush} & \textcolor{red}{\XSolidBrush} & \textcolor{red}{\XSolidBrush} & \textcolor{red}{\XSolidBrush} & \textcolor{red}{\XSolidBrush} & \textcolor{red}{\XSolidBrush} & 11.4K & 1.27K & \textcolor{red}{\XSolidBrush} \\
PubMedQA~\citep{jin-etal-2019-pubmedqa} & \textcolor{green}{\CheckmarkBold} & \textcolor{red}{\XSolidBrush} & \textcolor{red}{\XSolidBrush} & \textcolor{red}{\XSolidBrush} & \textcolor{red}{\XSolidBrush} & \textcolor{red}{\XSolidBrush} & \textcolor{red}{\XSolidBrush} & \textcolor{red}{\XSolidBrush} & \textcolor{red}{\XSolidBrush} & \textcolor{red}{\XSolidBrush} & 450 & 500 & \textcolor{red}{\XSolidBrush} \\
BioASQ~\citep{tsatsaronis2015overview} & \textcolor{green}{\CheckmarkBold} & \textcolor{red}{\XSolidBrush} & \textcolor{red}{\XSolidBrush} & \textcolor{red}{\XSolidBrush} & \textcolor{red}{\XSolidBrush} & \textcolor{red}{\XSolidBrush} & \textcolor{red}{\XSolidBrush} & \textcolor{red}{\XSolidBrush} & \textcolor{red}{\XSolidBrush} & \textcolor{red}{\XSolidBrush} & 745 & 140 & \textcolor{red}{\XSolidBrush} \\
MedAgentsBench~\citep{tang2025medagentsbench} & \textcolor{green}{\CheckmarkBold} & \textcolor{red}{\XSolidBrush} & \textcolor{red}{\XSolidBrush} & \textcolor{red}{\XSolidBrush} & \textcolor{red}{\XSolidBrush} & \textcolor{red}{\XSolidBrush} & \textcolor{red}{\XSolidBrush} & \textcolor{red}{\XSolidBrush} & \textcolor{red}{\XSolidBrush} & \textcolor{red}{\XSolidBrush} & -- & 862 & \textcolor{red}{\XSolidBrush} \\
MIRAGE~\citep{xiong2024benchmarking} & \textcolor{green}{\CheckmarkBold} & \textcolor{red}{\XSolidBrush} & \textcolor{red}{\XSolidBrush} & \textcolor{red}{\XSolidBrush} & \textcolor{red}{\XSolidBrush} & \textcolor{red}{\XSolidBrush} & \textcolor{red}{\XSolidBrush} & \textcolor{red}{\XSolidBrush} & \textcolor{red}{\XSolidBrush} & \textcolor{red}{\XSolidBrush} & -- & 7.66K & \textcolor{red}{\XSolidBrush} \\
HealthBench~\citep{healthbench} & \textcolor{green}{\CheckmarkBold} & \textcolor{red}{\XSolidBrush} & \textcolor{green}{\CheckmarkBold} & \textcolor{red}{\XSolidBrush} & \textcolor{red}{\XSolidBrush} & \textcolor{red}{\XSolidBrush} & \textcolor{red}{\XSolidBrush} & \textcolor{red}{\XSolidBrush} & \textcolor{red}{\XSolidBrush} & \textcolor{red}{\XSolidBrush} & -- & 5K & \textcolor{red}{\XSolidBrush} \\
\midrule
EHRSQL~\citep{lee2022ehrsql} & \textcolor{red}{\XSolidBrush} & \textcolor{green}{\CheckmarkBold} & \textcolor{green}{\CheckmarkBold} & \textcolor{red}{\XSolidBrush} & \textcolor{red}{\XSolidBrush} & \textcolor{red}{\XSolidBrush} & \textcolor{red}{\XSolidBrush} & \textcolor{red}{\XSolidBrush} & \textcolor{red}{\XSolidBrush} & \textcolor{red}{\XSolidBrush} & 15.5K & 1.73K & \textcolor{red}{\XSolidBrush} \\
MedCalcBench~\citep{khandekar2024medcalc} & \textcolor{red}{\XSolidBrush} & \textcolor{red}{\XSolidBrush} & \textcolor{red}{\XSolidBrush} & \textcolor{green}{\CheckmarkBold} & \textcolor{red}{\XSolidBrush} & \textcolor{red}{\XSolidBrush} & \textcolor{red}{\XSolidBrush} & \textcolor{red}{\XSolidBrush} & \textcolor{red}{\XSolidBrush} & \textcolor{red}{\XSolidBrush} & 10.1K & 1.05K & \textcolor{red}{\XSolidBrush} \\
MedAgentBench~\citep{jiang2025medagentbench} & \textcolor{red}{\XSolidBrush} & \textcolor{green}{\CheckmarkBold} & \textcolor{red}{\XSolidBrush} & \textcolor{green}{\CheckmarkBold} & \textcolor{red}{\XSolidBrush} & \textcolor{red}{\XSolidBrush} & \textcolor{green}{\CheckmarkBold} & \textcolor{green}{\CheckmarkBold} & \textcolor{red}{\XSolidBrush} & \textcolor{red}{\XSolidBrush} & -- & 300 & \textcolor{red}{\XSolidBrush} \\
BioCoder~\citep{tang2024biocoder} & \textcolor{red}{\XSolidBrush} & \textcolor{green}{\CheckmarkBold} & \textcolor{red}{\XSolidBrush} & \textcolor{green}{\CheckmarkBold} & \textcolor{green}{\CheckmarkBold} & \textcolor{red}{\XSolidBrush} & \textcolor{green}{\CheckmarkBold} & \textcolor{red}{\XSolidBrush} & \textcolor{green}{\CheckmarkBold} & \textcolor{red}{\XSolidBrush} & -- & 1.24K & \textcolor{red}{\XSolidBrush} \\
BioDSBench~\citep{wang2024can} & \textcolor{red}{\XSolidBrush} & \textcolor{green}{\CheckmarkBold} & \textcolor{red}{\XSolidBrush} & \textcolor{green}{\CheckmarkBold} & \textcolor{green}{\CheckmarkBold} & \textcolor{red}{\XSolidBrush} & \textcolor{green}{\CheckmarkBold} & \textcolor{green}{\CheckmarkBold} & \textcolor{red}{\XSolidBrush} & \textcolor{red}{\XSolidBrush} & -- & 128 & \textcolor{red}{\XSolidBrush} \\
EHRSHOT~\citep{wornow2023ehrshot} & \textcolor{red}{\XSolidBrush} & \textcolor{green}{\CheckmarkBold} & \textcolor{red}{\XSolidBrush} & \textcolor{red}{\XSolidBrush} & \textcolor{red}{\XSolidBrush} & \textcolor{green}{\CheckmarkBold} & \textcolor{red}{\XSolidBrush} & \textcolor{red}{\XSolidBrush} & \textcolor{red}{\XSolidBrush} & \textcolor{red}{\XSolidBrush} & -- & 15 & \textcolor{red}{\XSolidBrush} \\
\rowcolor{RoyalPurple!12} \textbf{\method{} (Ours)} & \textcolor{red}{\XSolidBrush} & \textcolor{green}{\CheckmarkBold} & \textcolor{green}{\CheckmarkBold} & \textcolor{green}{\CheckmarkBold} & \textcolor{green}{\CheckmarkBold} & \textcolor{green}{\CheckmarkBold} & \textcolor{green}{\CheckmarkBold} & \textcolor{green}{\CheckmarkBold} & \textcolor{green}{\CheckmarkBold} & \textcolor{green}{\CheckmarkBold} &  \textbf{59.2K}	& \textbf{13.2K} & \textbf{6.7K}\\
\bottomrule
\end{tabular}}
\end{threeparttable}
% \vspace{-3ex}
\end{table}

%% file: section/3-method.tex
\section{MedAgentGym: A Scalable and Interactive LLM Agent Training Environment for Code-Centric Biomedical Reasoning}
\label{sec:method}

\subsection{Problem Formulation}
We formulate coding-based reasoning as a structured problem-solving task:
given a problem description $x \in \mathcal{X}$, the goal is to generate a code snippet $c \in \mathcal{C}$ that produces an output $y \in \mathcal{Y}$. Each instance $(x, y)$ is paired with a ground truth output $y^*$, and the correctness is verified using $\mathcal{E}: \mathcal{C} \times \mathcal{Y} \rightarrow\{0,1\}$, where $\mathcal{E}= \mathbb{I}(y = y^*)$.
Existing biomedical reasoning datasets typically provide only question-answer pairs ($x, y^*$) without code solutions $c$ or only include a single predefined code solution per task. 
To address this, \method{} enables scalable generation and sampling of multiple coding trajectories $c^{(0)},c^{(1)},\cdots,c^{(k)}$ with corresponding executions $y^{(0)},y^{(1)},\cdots,y^{(k)}$ through parallel execution of LLM agents. 
Each trajectory is either single-turn or multi-turn, depending on task complexity and user requirements. 
Crucially, \method{} captures both \emph{positive} trajectories $\{c^{(i)}|y^{(i)}=y^*\}$ that succeed
and \emph{negative} trajectories $\{c^{(i)}|y^{(i)}\neq y^*\}$ including error messages as learning signals.
% The details of the data and the interactive environment are in \cref{sec:method-data} and \cref{sec:method-env}.

\subsection{Data Construction: From Individual Datasets to Unified Benchmark} 
\label{sec:method-data}

\input{table/tab-data}

\textbf{Task and Data Identification.} 
\method{} focuses on verifiable biomedical data science tasks that benefit from code-based solutions (\ie, code-centric biomedical reasoning).
\emph{Clinically}, we prioritize tasks originating from real-world healthcare settings and validated by a multidisciplinary panel of healthcare experts. 
For example, \method{} involves MIMIC-III and eICU in EHRSQL~\citep{lee2022ehrsql} collected from 222 hospital staff members and annotated by human programmers. 
\emph{Computationally}, we integrate diverse coding tasks, ranging from \emph{structured medical information retrieval} to \emph{open-ended biomedical research}, ensuring comprehensive coverage and task diversity. 

\textbf{Verifiable Instances Preparation.}
To standardize tasks across various sources, each instance in \method{} is structured with: (1) a problem description, (2) verifiable ground-truth outputs, and (3) optional data resources (\eg, EHRs). 
Additionally, standardized system and user prompts are designed to initiate the problem-solving process (see appendix \ref{app:prompt}). 
\method{} is highly flexible, easily accommodating new tasks that include clear descriptions and verifiable ground-truth outputs.
For coding-centric tasks that provide only reference code implementations (\eg, BioCoder~\citep{tang2024biocoder}), we validate task correctness based on the execution output of these reference solutions, generating definitive output signatures. 
This transformation is necessary because multiple valid code implementations may yield identical execution results, making the execution outcome--rather than the code itself--a more reliable and consistent verification signal.
For tasks involving additional data resources (\eg, EHRSQL~\citep{lee2022ehrsql}), we include metadata on data access and sources. 
Detailed task overview and task-specific preparation are documented in appendix \ref{app:data}.

\textbf{Data Statistics.}
\method{} is a unified training environment built upon a large-scale, high-quality dataset comprising approximately 72,000 task instances across 129 categories from 12 real-world biomedical scenarios. 
Notably, with \method{}, we collect large-scale agent trajectories to support coding agent development (\cref{sec:medcopilot}). 
To ensure reproducible and robust evaluation, we define clear train/test splits, separate internal and external validation sets, and perform $n$-gram ($n=10$) string match to eliminate the data contamination issue.
Table~\ref{tab:data} provides statistics for \method{}. 
To accommodate diverse research needs, we offer two versions of \method{}: (1) a comprehensive, full-scale dataset for extensive exploration and detailed analysis, and (2) a balanced, lightweight subset for efficient leaderboard training and evaluation.

\subsection{Coding Environment: From Static Benchmark to Interactive Interface}
\label{sec:method-env}

\begin{figure}
    \centering
    % \vspace{-3ex}
    \includegraphics[width=0.99\linewidth]{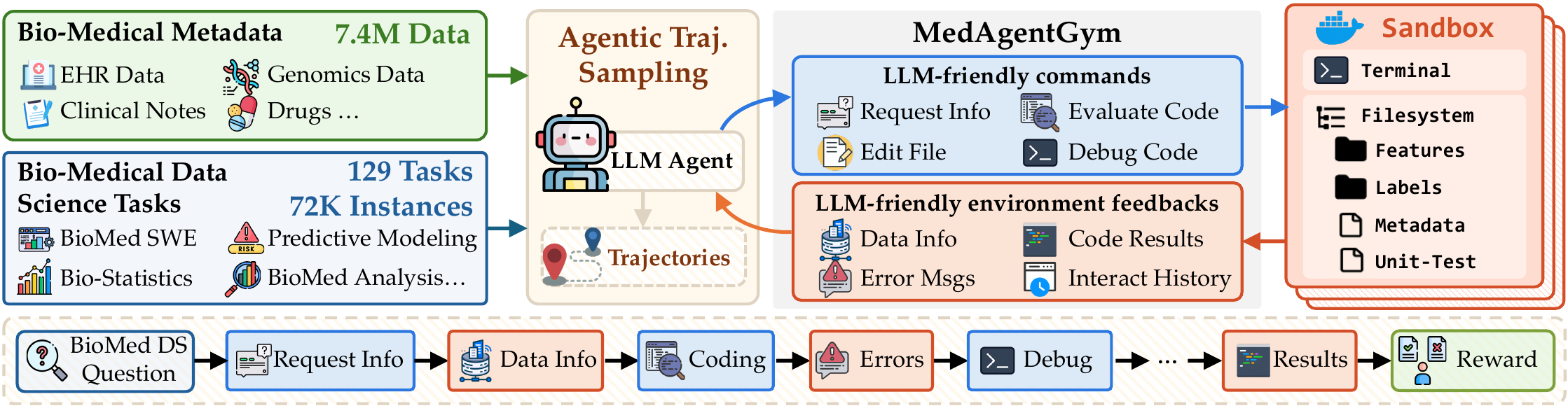}
    % \vspace{-1ex}
    \caption{Overview of \method{}. \method{} contains a comprehensive suite of coding-centric biomedical data science tasks with an interactive execution environment for LLM agents.}
    % \vspace{-2ex}
    \label{fig:overview}
\end{figure}

\textbf{Isolated and Executable Sandbox Environment.} 
To ensure robust and reproducible coding-based biomedical reasoning, \method{} provides isolated executable coding environments (\ie, sandbox) through Docker containers tailored to each task (Figure~\ref{fig:overview}). 
These containers come pre-installed with all required dependencies, including specialized biomedical packages (\eg, \texttt{AlignIO} in BioCoder~\citep{tang2024biocoder}), facilitating reliable task execution.
To address critical data safety concerns, each Docker environment guarantees: (1) \emph{environmental integrity}, where isolation prevents contamination or data corruption potentially caused by LLM-generated code, preserving both the computational environment and the underlying data systems~\citep{yang2024swe}; 
(2) \emph{medical data security}, where secure containerization enforces compliance with medical data usage policies, safeguarding sensitive patient information.
Additionally, \method{} supports extensive flexibility for integrating new tasks, where users can define customized Docker environments through configuration files. 
If certain packages are not initially available, a terminal tool allows LLM agents to dynamically install the required dependencies within their isolated environments.

\textbf{Interactive Feedback.} 
\method{} incorporates interactive feedback mechanisms, effectively bridging LLMs with coding interpreters:
(1) \emph{robust parsing:} To begin, the output generated by LLMs is formatted in structured JSON, facilitating straightforward parsing and code execution. In cases of execution errors, iterative JSON regeneration is employed to maximize successful code execution rates.
(2) \emph{debugging and error grounding:} Compile-time and runtime error messages are systematically translated into a unified natural language format, making them more accessible to LLMs and significantly improving debugging efficiency and interpretability.

\textbf{Efficient Trajectory Collection.} 
Each task in \method{} is packaged in a reproducible Docker image with built-in support for \emph{multi-threading}, \emph{parallel execution}, and \emph{sequential sampling}. 
Specifically, we integrate two widely used multi-threading backend engines, \texttt{Ray}\footnote{\url{https://github.com/ray-project/ray}} and \texttt{Joblib}\footnote{\url{https://joblib.readthedocs.io/en/stable/}}, to accelerate trajectory sampling.
This infrastructure ensures efficient and scalable trajectory collection, supporting both extensive experimentation and systematic evaluation across multiple scenarios.
% Furthermore, \method{} easily adapts tasks without ground-truth code solutions into coding-based tasks with verifiable outputs, significantly enhancing the diversity and complexity of evaluation scenarios.

\textbf{Plug-and-Play.}
A key strength of \method{} lies in its flexible and modular architecture, which readily supports the integration of new biomedical coding tasks. 
This inherent extensibility enables \method{} to continually adapt to evolving advancements in biomedical sciences and artificial intelligence methodologies. Additionally, its trajectory sampling approach allows the straightforward transformation of traditional, non-executable biomedical reasoning tasks into coding-based scenarios with verifiable outputs, significantly broadening the scope and complexity of tasks that can be systematically evaluated. Moreover, users can define custom Docker environments through configuration files, and, if specific software packages are initially absent, a built-in terminal tool facilitates dynamic installation within each isolated execution environment, further improving \method{} in runtime adaptability and user-friendliness.

%% file: table/tab-data.tex
\begin{table}[t]
    \centering
    \caption{
    Dataset statistics for \method{} and its lightweight subset for leaderboard evaluation. $^\star$For open-ended tasks without explicit ground truth (\eg, ML coding in EHRSHOT and MIMIC-Extract), we follow standard RL settings by using the same dataset for training and evaluation. 
    % $^+$Only the test set is utilized for external evaluation, while the corresponding training data remains accessible. Similarly, we provide a subset of training data in leaderboard for efficiency and convenience, without restricting the incorporation of supplementary training resources.
}
    \label{tab:data}
    \begin{threeparttable}
    \renewcommand\arraystretch{0.98}
\resizebox{\linewidth}{!}{ %
    \begin{tabular}{l|cccc|c|ccc|ccc}
    \toprule
& \multicolumn{4}{c|}{\textbf{Data Sources}} & \multicolumn{4}{c|}{\textbf{Task Instances (all)}} & \multicolumn{3}{c}{\textbf{Tasks ({leader-board})}} \\
\cmidrule(lr){2-5} \cmidrule(lr){6-9} \cmidrule(lr){10-12} 
\textbf{Dataset} & {Type} & {\#Patients} & {\#Table} & {\#Elements} & {Category} & {\#Train} & {\#Test} & {\#Total} & {\#Train} & {\#Test} & {\#Total} \\
\midrule
\multicolumn{11}{l}{\textit{Training and Internal Validation (In-Distribution)}} \\ \midrule
{MIMIC-III}~\citep{johnson2016mimic} & Tabular & <1K &	17 & 1.4M	& 9	& 9,318 & 1,122 & 10,440 & 552 & 581 & 1,133 \\
{eICU}~\citep{pollard2018eicu} & Tabular	& <1K	& 10 & 1.5M	& 9 &	6,213	& 611	& 6,824 &	559 &	610 &	1,169\\
{TREQS}~\citep{wang2020text} & Tabular & 100 &	5 & 2.5M	& 4	& 8,988	& 996 &	9,984 &	897 &	995 &	1,892 \\ 
{MedCalcBench}~\citep{khandekar2024medcalc} & Text	& 1K	& -- &--	& 55	& 10,053	& 1,047 & 11,100	& 1,005 &	1,046 &	2,051\\
{MedAgentBench}~\citep{jiang2025medagentbench} & Tabular& 100	&-- &700K	& 10 &	433 &	109 &	542 &	239 &	59	& 298 \\
{BioCoder}~\citep{tang2024biocoder} & Text	& --	& -- &--	& 8	& 981	&157 &	1,138	& 981 &	156	& 1,137 \\
{EHRSHOT}~\citep{wornow2023ehrshot} & Tabular &	63K	& 31 & 1.2M	& 15	& 15	& 15 & 15$^\star$ &	15	& 15 & 15$^\star$ \\
{BioDSBench}~\citep{wang2024can} & Text	& --& -- & -- &12 & 50 & 49 &	99 &	50 & 49	& 99 \\
\rowcolor{RoyalPurple!6} \textbf{\method{} (Internal)} & --	& 65K & 63 & 7.3M & 113	& 36,036	& 4,106	& 40,142	& 4,283	& 3,511	& 7,794 \\
\midrule
\multicolumn{11}{l}{\textit{External Validation (Out-of-Distribution)$^+$only the test set for external evaluation; training data remains accessible}} \\ \midrule
{EHR-SeqSQL}~\citep{ryu2024ehr} & Tabular &<1K	& 17 & 1.4M	& 4	& 18,950	& 7,913 & 26,863 &1,000	& 500 &	1,500 \\
{EHRCon}~\citep{kwon2024ehrcon}  & Tab\&Text	& 46K &13 &--	& 3 &	3,229	& 976	&4,205&	1,000 &	500 &	1,500 \\
{MIMIC-Extract}~\citep{MIMICExtract} & Tabular	& 35K &4 &35K	&3 &	3 &	3 &	3$^\star$ &	3 &	3 &	3$^\star$\\
% {InspectEHR}~\citep{inspectehr} & Tabular	& 19K &	32 &23K &2	&2	&2	&2$^\star$ &	2 &	2	& 2$^\star$\\
{N-PowerAI}~\citep{ruan2025n} &  Text	& -- &-- &--&6	&960& 240&	1200	&960& 240&	1200\\
\rowcolor{RoyalPurple!6} \textbf{\method{} (External)} & --	& 82K & 34 & 1.4M & 16	& 23,142	& 9,132	& 32,271	& 2,963 & 1,243	& 4,203\\ \midrule
\multicolumn{11}{l}{\textit{Overall}} \\ \midrule
\rowcolor{RoyalPurple!12} \textbf{\method{}} & --	& \textbf{146K} & \textbf{80} & \textbf{7.4M} & \textbf{129} &	\textbf{59,175}	& \textbf{13,238} &	\textbf{72,413}	 & \textbf{7,243}	& \textbf{4,754}	& \textbf{11,997}\\
\bottomrule
\end{tabular}
}
\end{threeparttable}
\end{table}

%% file: section/4-exp.tex
\section{Evaluating LLMs as Medical Coding Agents with MedAgentGym}
\label{sec:exp}

\subsection{Experiments Setup}\label{subsec:setup}
\textbf{Agent Scaffolds.}  
Following CodeAct~\citep{wang2024executable}, we establish a default agent scaffold for systematically evaluating coding-based biomedical reasoning. Interactions within \method{} are modeled as a Partially Observable Markov Decision Process (POMDP), focusing on sampled biomedical data science tasks $p \in \mathcal{P}$. 
At each timestep $t$, the agent observes $o_t \in \mathcal{O}$ and samples an action $a_{t+1} \in \mathcal{A}$ from the current policy $\pi_{t}$ based on interaction history.
We define four primary action types:
(a) \texttt{request\_info}: retrieve relevant data from sources such as EHRs;
(b) \texttt{terminal}: manage dependencies or local files within isolated Docker environments.
(c) \texttt{code\_execution}: execute code generated by LLMs through an integrated interpreter; and
(d) \texttt{debugging:} translate code execution errors into natural language explanations enriched with detailed error information for LLM comprehension.

\textbf{Tasks and Datasets.}
Building upon \method{}, we train and evaluate \ours{} on 7,794 \emph{coding-based biomedical reasoning} tasks across 8 datasets:
(1) {MIMIC-III}~\citep{johnson2016mimic} and
(2) {eICU}~\citep{pollard2018eicu} from EHRSQL~\citep{lee2022ehrsql},
(3) {TREQS}~\citep{wang2020text},
(4) {MedCalcBench}~\citep{khandekar2024medcalc}, 
(5) {MedAgentBench}~\citep{jiang2025medagentbench},
(6) {BioCoder}~\citep{tang2024biocoder},
(7) {EHRSHOT}~\citep{wornow2023ehrshot}, and
(8) {BioDSBench}~\citep{wang2024can}.
Moreover, we conduct experiments for \emph{out-of-distribution} evaluation on 4,203 tasks from the following 4 datasets:
(9) {EHR-SeqSQL}~\citep{ryu2024ehr},
(10) {EHRCon}~\citep{kwon2024ehrcon},
(11) {MIMIC-Extract}~\citep{MIMICExtract}, and
(12) {N-PowerAI}~\citep{ruan2025n}.
Note that we do not consider knowledge-intensive medical question-answering tasks~\citep{jin-etal-2019-pubmedqa,pal2022medmcqa,jin2021disease}, as they are orthogonal to coding-aided reasoning.
We include detailed task and dataset information in appendix \ref{app:data}.

\textbf{Baselines.}
We extensively benchmark the following state-of-the-art LLMs on \method{}:
(i) \emph{API-based proprietary LLMs}, including  
gpt-4o-mini~\citep{hurst2024gpt},
gpt-4o~\citep{hurst2024gpt},
gpt-4.1-mini~\citep{gpt-4-1},
gpt-4.1~\citep{gpt-4-1},
gpt-o4-mini~\citep{o4-mini}, and codex-mini~\citep{chen2021evaluating};
(ii) \emph{OSS LLMs}, including  
gemma-3~\citep{team2025gemma},
Qwen3~\citep{qwen3},
Qwen2.5~\citep{yang2024qwen2},
Llama-3~\citep{dubey2024llama},
Ministral~\citep{Ministral-8b}, and
DeepSeek-R1~\citep{guo2025deepseek};
(iii) \emph{coding LLMs}, including 
codex-mini~\citep{chen2021evaluating}, Qwen2.5-Coder-7B-Instruct and -14B-Instruct~\citep{hui2024qwen2};
and (iv) \emph{medical reasoning LLMs} or medical domain-specific LLMs, including 
medgemma-4b-it (\texttt{gemma-3-4b-pt})~\citep{medgemma-hf},
HuatuoGPT-o1-7B (\texttt{Qwen2.5-7B-Instruct})~\citep{chen2024huatuogpt},
m1-7B-23K (\texttt{Qwen2.5-7B-Instruct})~\citep{huang2025m1}, 
MedReason-8B (\texttt{Llama-3.1-8B-Instruct})~\citep{wu2025medreason}, 
and Baichuan-M1-14B-Instruct~\citep{wang2025baichuan}.
Additional model details are available in appendix \ref{app:baseline}.

\input{table/tab-main}

\textbf{Evaluation Metrics.} 
We adopt \emph{success rate (SR)} as the primary evaluation metric. 
For \emph{database, data science, and bioinformatics} tasks with explicit ground truths, we compare LLM-generated code execution outputs with reference solutions using exact match. 
For open-ended \emph{ML} tasks in clinical decision support, we measure performance using \emph{accuracy (Acc)} across test cases. 
% Note that these code generation tasks inherently have infinite solution spaces, unlike traditional classification problems with bounded solution spaces (\eg, even random guessing can yield around 50\% accuracy in binary classification).
% The \emph{overall score} is computed by averaging performance across tasks in test sets of \method{} (leaderboard), providing a comprehensive evaluation of coding-based medical reasoning capabilities within \method{}.
See appendix \ref{app:implementation} for implementation details and \ref{app:codequlity} for additional evaluation on code quality and efficiency. 

% \textbf{Implementation Details.}
% We limit interactions to a maximum of $15$ turns per session, providing agents full access to interaction histories and constraining runtime to $120$ seconds per session. 
% Input tokens are capped at $32,768$, with output limited to $8,192$ tokens per round.
% We use Python $3.10$ as the primary language for agent-code execution due to its modular design and suitability for biomedical computations. 
% To enable interactive feedback (\cref{sec:method-env}), we employ a rule-based parser converting LLM outputs to JSON, facilitating seamless code execution, and utilize \texttt{gpt-4.1-mini} to translate execution errors into grounded explanations. 
% We configure all baseline LLMs following established best practices for reproducibility. 
% Specifically, instruction-following LLMs are configured with a temperature of zero, while reasoning models use a temperature of 0.6. 
% For all experiments with \texttt{Qwen-3} series, we switch to thinking mode for optimal performance under complex reasoning scenarios (\eg, logic, math, and coding).
% See \cref{app:implementation} for additional implementation details. 

\subsection{Results: Benchmarking LLMs and Reasoning Models with MedAgentGym}

Table~\ref{tab:main_results} benchmarks the state-of-the-art LLMs on \method{}. We summarize key observations from our zero-shot leaderboard evaluation as follows:
$\diamond$ \textbf{Significant Performance Gap Between Commercial API-based and OSS LLMs.} 
% Commercial API-based LLMs substantially outperform OSS models across all task categories.
% Specifically, \texttt{o4-mini} excels in structured medical information retrieval tasks, while \texttt{gpt-4.1} demonstrates superior performance in complex computational tasks, such as medical calculation and bioinformatics analysis. 
% In comparison, even strong OSS models such as \texttt{Qwen3-32B} lag significantly behind with more than a 20\% gap on performance.
% Interestingly, scaling OSS models from 4B to 14B yields minimal gains, with noticeable improvement only at 32B, indicating that coding-based medical reasoning capability emerges at larger scales.
This evident performance gap highlights the \emph{critical need for continued development} of lightweight OSS LLMs that match commercial performance while addressing real-world privacy and cost constraints.
$\diamond$ \textbf{Task-Specific Performance Variations between Structured and Open-ended Medical Tasks.} 
LLMs consistently perform better on structured tasks (\eg, database queries, medical calculations) compared to open-ended tasks requiring advanced coding and reasoning (\eg, data analysis, ML prediction). 
% This integration lets one study where coding helps (structured retrieval and computation) and where current models still struggle (open‑ended analysis and modeling).
% \texttt{MedCalcBench} further challenges models with tasks requiring precise recall of domain knowledge, such as medical formulas and diagnostic guidelines. 
% This highlights the diverse task composition in \method{}, enabling comprehensive evaluation of coding-based medical reasoning.
% $\diamond$ \textbf{Enhanced Performance in Reasoning Models.} 
% OSS reasoning models, such as \texttt{QwQ-32B} and \texttt{DeepSeek-R1} series, demonstrate superior performance in structured database querying and knowledge-intensive biomedical research tasks, indicating inherently robust reasoning capabilities essential for computational medical reasoning tasks in \method{}. 
$\diamond$ \textbf{Suboptimal Outcomes in Dedicated Coding and Medical Domain-Specific LLMs.} 
Both coding and medical reasoning LLMs deliver suboptimal performance, revealing that \emph{coding-based biomedical reasoning represents a unique capability} not adequately captured by specialization in either coding or medical reasoning.
% Surprisingly, medical reasoning models consistently underperform relative to their base models except for knowledge-intensive tasks (\eg, \texttt{MedCalcBench}, \texttt{BioCoder}), showing that fine-tuning in medical QA may reduce generalization and instruction-following ability. 
% These findings highlight the need for training data integration to jointly support coding skills and medical reasoning, rather than treating them as separate objectives.

\section{Training LLM Agents for Code-Centric Biomedical Reasoning}
\label{sec:medcopilot}

In this section, we leverage \method{} to systematically enhance lightweight OSS LLMs as proficient coding agents (\ours{}) for biomedical reasoning. 
We first explore a two-stage agentic fine-tuning framework (\cref{sec:sft}), followed by a detailed analysis of model scaling behaviors (\cref{subsec:scale}). 
We then introduce self-improvement to further boost agent performance (\cref{subsec:self-improve}) and conduct additional analysis on model generalization, ablation, and error patterns (\cref{sec:exp-agent}).

\subsection{RL Fine-tuning with Trajectory Sampling}
\label{sec:sft}

\begin{figure}[t]
    \centering
    % \vspace{-4ex}
    \includegraphics[width=0.99\linewidth]{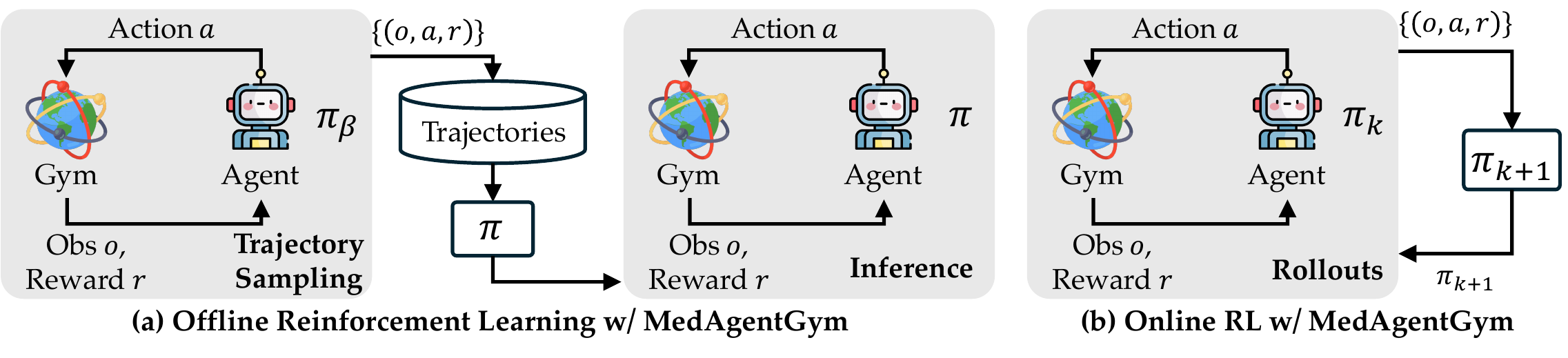}
    % \vspace{-1.5ex}
    \caption{Comparison of (a) offline and (b) online RL paradigms within \method{}. 
    % (a) Offline RL: We first use an arbitrary expert policy $\pi_\beta$ to rollout and sample a fixed dataset of trajectories $\{(s,a,r)\}$ for training the target policy $\pi$. (b) Online RL: The Agent's current policy $\pi_k$ is used to actively collect new rollouts, and the resulting data is immediately used to update and improve the policy $\pi_{k+1}$ to improve the policy in a continuous loop.
    }
    % \vspace{-1ex}
    \label{fig:rl}
\end{figure}

\textbf{Training Setup.}
We select \texttt{Qwen-2.5-Instruct-7B} and \texttt{-14B}~\citep{yang2024qwen2} as our backbones. To enable effective evaluation within \method{}, we utilize a consistent CodeAct-style scaffold, allowing LLM agents to iteratively reason and refine biomedical code through interactive environment feedback. Detailed training setups, including hyperparameters, are provided in appendix \ref{app:implementation}.

\textbf{Trajectory Sampling.} 
\method{} facilitates efficient parallel trajectory sampling using \texttt{ray} and \texttt{joblib} backends. 
Specifically, we roll out (1) 2,137 successful trajectories using \texttt{gpt-4.1-mini} with a temperature of 0 to warm up the fine-tuning for smaller OSS models. 
Each successful trajectory contains 9.25 turns between the LLM and the code interpreter on average. 
In addition to 2,137 positive trajectories for supervised fine-tuning (SFT), we prepare additional trajectory pairs for RL such as direct preference optimization (DPO), including (2) 1,646 offline pairs sampled from \texttt{gpt-4.1-mini}, and (3) 2,939 online pairs. 
For both types, we use the initial prompt interactions as shared context and contrast successful final codes against intermediate erroneous attempts. We release all 6K trajectories above to accelerate coding agent development.
See appendix \ref{app:trajcompo} for detailed trajectories composition.

\textbf{Two-Stage Fine-Tuning.} 
We benchmark two policy improvement methods: (1) SFT directly mimics high-reward trajectories consisting exclusively of successful outcomes, whereas (2) offline or online RL optimizes the policy by favoring selected responses over rejected ones (Figure~\ref{fig:rl}). We further consider a two-stage fine-tuning, initially warming up with SFT and subsequently refining with RL.

\input{table/tab-training}

\textbf{Results: Offline RL (DPO).}
% Figure~\ref{fig:sft} highlights substantial performance gains from SFT across four OSS backbone LLMs of varying sizes. 
Table~\ref{tab:training_results} compares several post-training methods, revealing that simple SFT over successful trajectories significantly boosts performance on structured coding tasks, demonstrating its effectiveness in capturing structured coding patterns. 
Besides, DPO is particularly beneficial for optimizing open-ended task performance. 
% Although DPO alone slightly underperforms compared to SFT, combining an initial SFT warm-up with subsequent DPO further improves overall results by leveraging their complementary strengths.

\textbf{Results: Online RL (PPO and GRPO).}
We further consider online RL methods, including Proximal Policy Optimization (PPO)~\citep{schulman2017proximal} and Group Relative Policy Optimization (GRPO)~\citep{shao2024deepseekmath}, to enable \ours{} to actively explore tasks and dynamically generate higher-quality training data through interaction. 
The evaluation module of \ours{} is employed to provide two reward signals: a correctness reward and a format reward, the latter indicating whether the generated output contains code blocks.
As shown in Table~\ref{tab:training_results}, GRPO achieve markedly stronger performance, suggesting enhanced generalization capabilities in diverse biomedical scenarios compared with offline RL. 
% Given computational resource constraints, we conduct this proof-of-concept only on the 7B model variant; extending these approaches to larger model scales remains a promising direction for future work.

\subsection{Scaling LLM Agent Improvements with MedAgentGym}\label{subsec:scale}

\textbf{Verifier Training Setup.} 
In addition to directly training coding agents, \method{} facilitates the development of an outcome-supervised reward model (ORM) to evaluate generated solutions effectively. Inspired by prior work~\citep{cobbe2021training,pan2024training}, we formalize the verifier task as predicting the probability that a given trajectory successfully solves a coding task. Formally, we represent a trajectory as an interleaved sequence 
$\tau=[o_1,a_1,o_2,a_2,\cdots,o_n,a_n],\ r\in[0,1],$
where each observation $o_k$ comprises elements such as task descriptions, code execution results, and error feedback. 
We fine-tune a \texttt{Qwen2.5-7B-Instruct} model as a verifier with binary predictions `YES' ($l_y$) or `NO' ($l_n$), from which we compute success probability: $r=\exp(l_y)/(\exp(l_y)+\exp(l_n))$.

\textbf{Verifier Training Data.} 
We construct the verifier training dataset by combining two sets of trajectories originally sampled for agent training:
(1) \emph{off-policy trajectories}, consisting of 2,742 samples from \texttt{gpt-4.1-mini}; and 
(2) \emph{on-policy trajectories}, comprising 2,939 samples generated by the agent. 
Combining both on- and off-policy trajectories, we ensure a balanced dataset of successful and unsuccessful trajectories, filtering to fit within a maximum context length of 32k tokens.

\begin{figure}[t]
    % \vspace{-3ex}
	\centering
	\includegraphics[width=\linewidth]{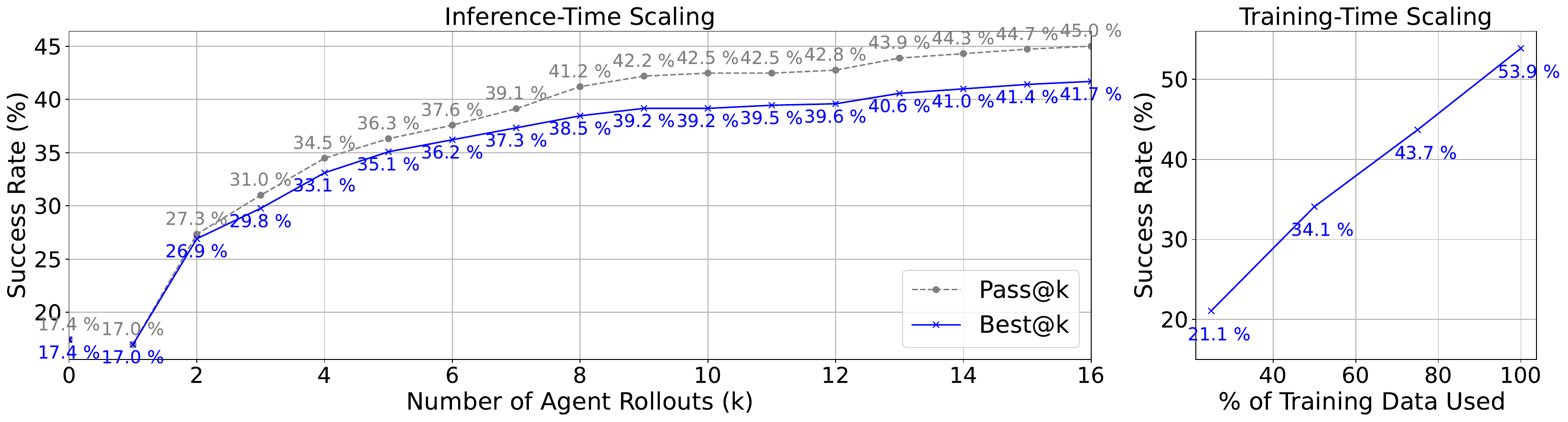}
    \caption{Scalable improvements of LLM agents in \method{}. 
    For inference-time scaling, we employ $T=0$ for the initial rollout and $T=0.6$ for the rest. For train-time scaling, we set $T=0$.
    }
    % \vspace{-2ex}
    \label{fig:scaling}
\end{figure}

\textbf{Results: Inference and Training-Time Scaling.} 
We introduce two additional evaluation metrics: (1) \emph{Pass@K}: the fraction of tasks solved by at least one trajectory from $K$ sampled attempts; and (2) \emph{Best@K}: the fraction of accurately selects successful trajectories. 
Figure~\ref{fig:scaling} (left) illustrates the performance scaling with increasing trajectory sampling. Pass@K significantly improves from 17.0\% at $K=1$ to 45.0\% at 16, while Best@K shows steady advancement from 17.0\% to 41.7\%. 
The relatively small gap between metrics indicates that our trained verifier effectively identifies successful trajectories, unleashing its potential as a reward model for integration into advanced online RL frameworks.
% \textbf{Results: Training-Time Scaling.}
Figure~\ref{fig:scaling} (right) examines agent performance as a function of increased training data volumes in SFT. We observe consistent performance improvements with greater training data availability, suggesting additional computational resources dedicated to sampling further trajectories are likely to yield continued performance gains.

\subsection{Model Performance Scaling with Self-improvement}
\label{subsec:self-improve}

\textbf{Self-Improvement Training Setup.} 
Beyond expert-generated trajectories, we explore self-improvement by refining the model using its own outputs. 
We employ rejection sampling fine-tuning (filtered behavior cloning), using the verifier from~\cref{subsec:scale} to score rollouts. We collect 4,298 trajectory pairs, each comprising the highest-scored correct and lowest-scored incorrect trajectories per prompt.
Starting from \texttt{Qwen2.5-7B-Instruct}, we perform SFT on 1,000 randomly sampled successful trajectories, followed by DPO using eight new rollouts per task and another 4,298 scored pairs. 
% We then execute eight rollouts per task, scoring them with the verifier and selecting another 4,298 pairs for DPO. 
We repeat this DPO step iteratively (iDPO) for further refinement.

\begin{figure}[t]
% \vspace{-3ex}    
\centering    
    \begin{minipage}{0.366\textwidth}
    \centering
    \includegraphics[width=0.99\textwidth]{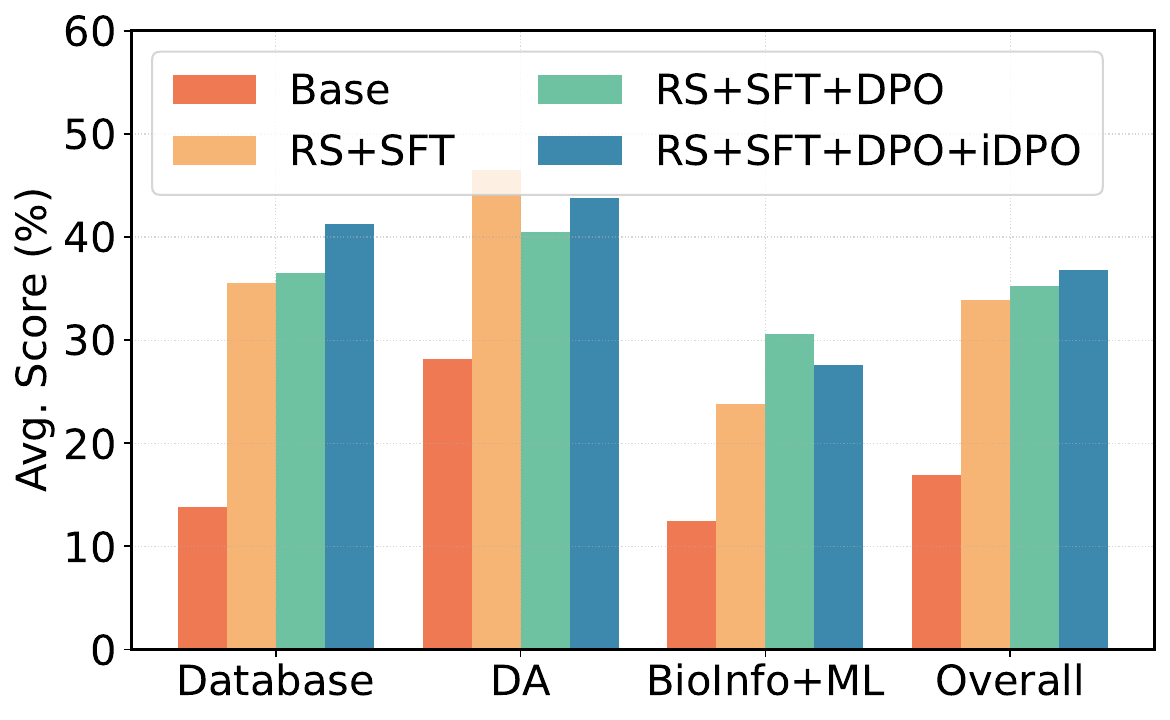}
    % \vspace{-3ex}
    \caption{Self-Improvement \vspace{-0.5ex}}
    \label{fig:self-improvement}
    \end{minipage}
    \hfill
    \begin{minipage}{0.316\textwidth}
    \centering
    \includegraphics[width=0.99\textwidth]{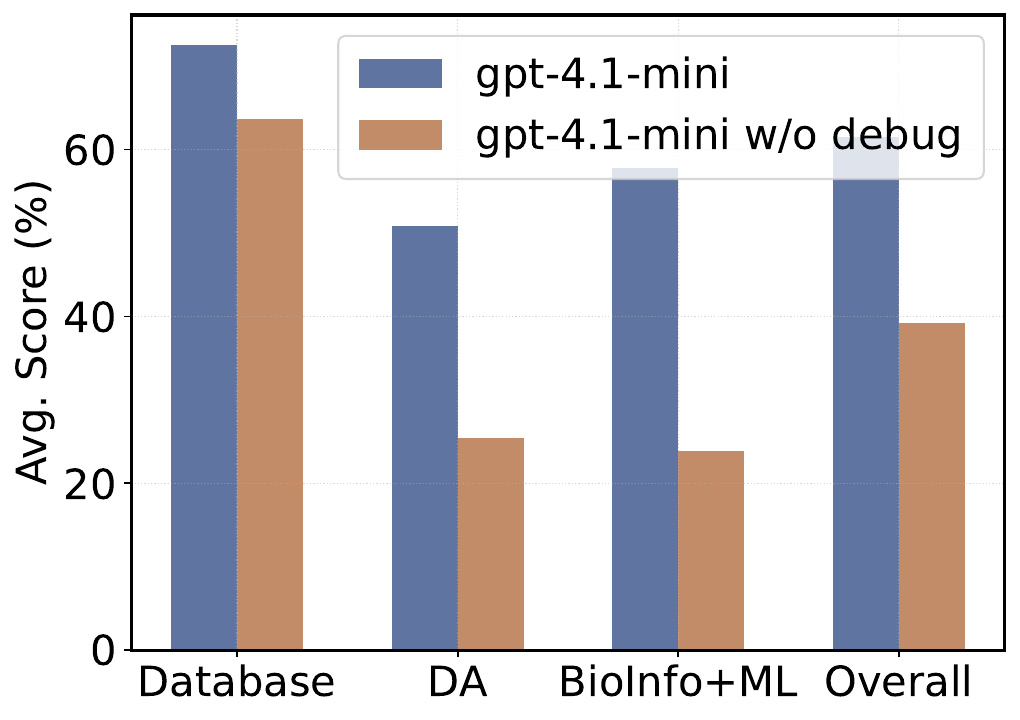}
    \caption{Effect of Debug \vspace{-0.5ex}}
    \label{fig:debug}
    \end{minipage}
    \hfill
    \begin{minipage}{0.25\textwidth}
    \centering
    \includegraphics[width=0.99\textwidth]{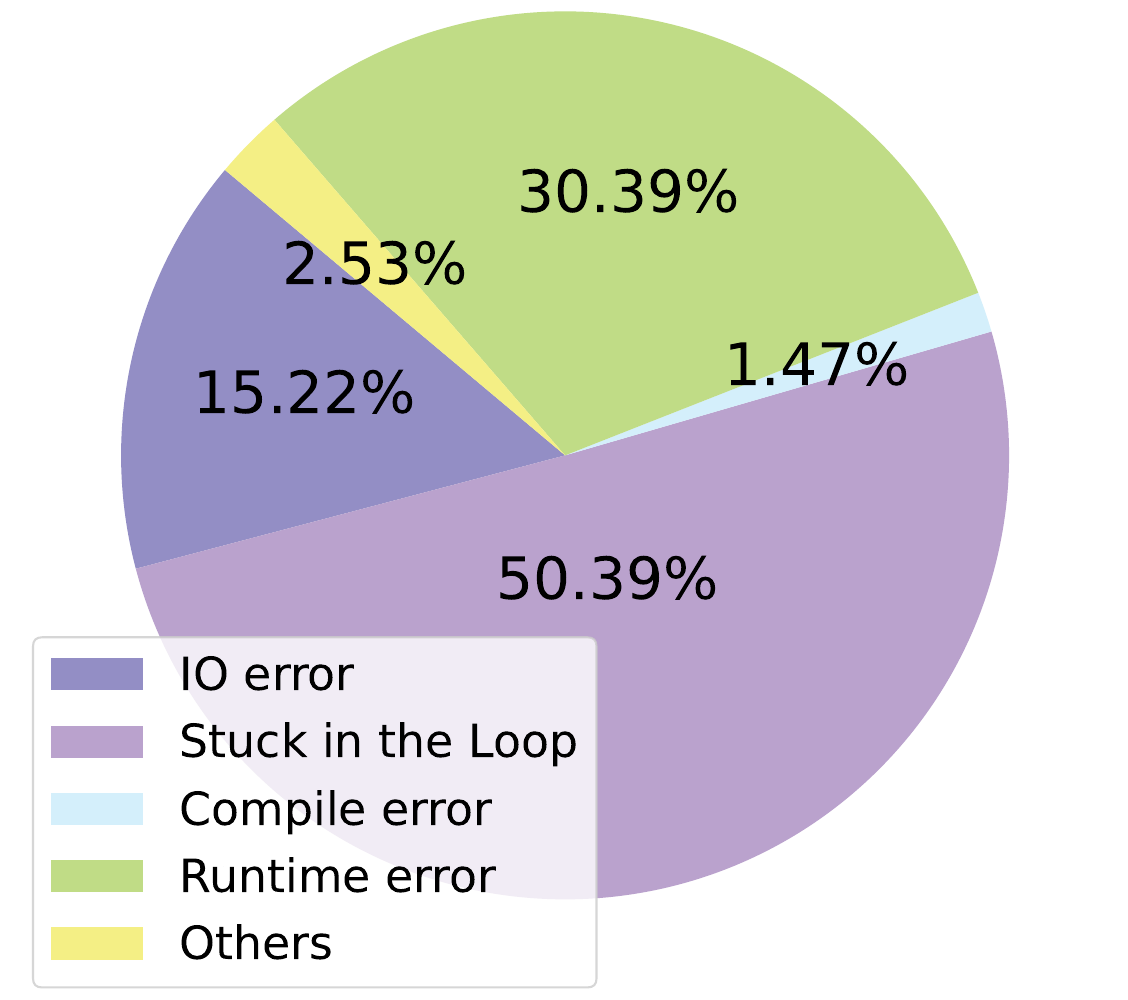}
    \caption{Error Types \vspace{-0.5ex}}
    \label{fig:error}
    \end{minipage}
    \vspace{-1ex}
\end{figure}

\textbf{Results: Rejection Sampling (RS) and iDPO.}
Figure~\ref{fig:self-improvement} illustrates consistent performance gains across one SFT stage and two subsequent DPO stages. 
However, we observe diminishing returns over successive iterations. Initially, rejection sampling SFT significantly boosts performance by effectively capturing successful coding patterns. 
Subsequent DPO stages show smaller incremental improvements, reflecting the model's diminishing exploration space as it tackles increasingly challenging tasks, ultimately converging toward an approximate Nash equilibrium.

% \begin{figure}[t]
%     \vspace{-2ex}
% 	\centering
%     \subfigure[Self-Improvement]{
% 	\includegraphics[width=0.365\linewidth]{figure/self-improvement.pdf}
% 	\label{fig:self-improvement}
% 	}
%     \subfigure[Effect of Debug]{
% 	\includegraphics[width=0.31\linewidth]{figure/debug.pdf}
% 	\label{fig:debug}
% 	} 
% 	\subfigure[Error Types]{
% 	\includegraphics[width=0.25\linewidth]{figure/error.pdf}
% 	\label{fig:error}
% 	}
% 	\caption{Additional studies. For better visualization, we include ML as part of Bio-info. category. \ws{revise}}
%     \vspace{-2ex}
% \label{fig:analysis}
% \end{figure}

\subsection{Generalization, Ablation, and Error Analysis}
\label{sec:exp-agent}

\input{table/tab-ood}
\textbf{Results: External Evaluation.} 
Table~\ref{tab:ood_results} summarizes external evaluation results on \method{}. 
% \ours{} models modestly improve performance on open-ended, reasoning-intensive tasks (\eg, MIMIC-Extract). However, improvements remain limited, with occasional declines in certain domain-specific tasks, indicating challenges in generalizing across specialized biomedical contexts. 
In particular, incorporating online RL optimization techniques, especially GRPO~\citep{shao2024deepseekmath}, can effectively improve performance on unseen, out-of-distribution tasks.

\textbf{Effect of Interactive Coding.} 
Figure~\ref{fig:debug} shows that removing debugging capabilities significantly decreases model performance across all tasks. Interactive coding mechanism in \method{} substantially contributes to successful coding-based medical reasoning by enabling the model to effectively interpret and rectify execution errors.

\textbf{Error Analysis.} 
Figure~\ref{fig:error} summarizes common error types encountered by the strongest evaluated LLM, \texttt{gpt-4.1}. Loop-related issues dominate, accounting for 50.39\% of errors, where agents repeatedly execute the same action in the final turns, indicating difficulty in adapting or exploring alternative strategies. 
This highlights the need to promote effective exploration and enhance robustness in solving complex biomedical reasoning tasks.
Additional experimental results, including cost analysis, case studies, and human studies, are available in appendix \ref{app:exp}.

%% file: table/tab-main.tex
\begin{table}[t]
\centering
\vspace{-3ex}
\caption{Test set results (zero-shot) of LLMs on \method{}. \textbf{Bold} indicates the best result at each scale. 
% $^\dagger$: We only consider Microsoft Azure OpenAI API services due to credentialed health data use agreement. 
$^\ddagger$ and $^\vee$ denote coding LLMs and medical reasoning LLMs, respectively. 
% Notations are consistent across tables. 
% We present the best test-set performance of \ours{} fine-tuned on the \method{} (\cref{sec:medcopilot}) training set for reference only.
}
% \vspace{-1ex}
\resizebox{\linewidth}{!}{ %
\begin{tabular}{l | c c c | c c | c c | c | c}
\toprule
\textbf{Datasets ($\rightarrow$)} &  {\textbf{MIMIC.}} & {\textbf{eICU}} & {\textbf{TREQS}} & {\textbf{MedCalc.}} & \textbf{MedAgent.} & 
 \textbf{BioCoder} & \textbf{BioDS.} & \textbf{EHRSHOT} & \textbf{Avg.}  \\
\textbf{Baselines ($\downarrow$) / Metrics ($\rightarrow$)} & SR & SR & SR & SR & SR & SR & SR & Acc  & Score \\ \midrule
\multicolumn{10}{l}{\textit{API-based Proprietary LLMs$^\dagger$}: We only consider Microsoft Azure OpenAI API services due to credentialed health data use agreement.} \\ \midrule
\texttt{gpt-4o-mini (2024-07-28)}~\citep{hurst2024gpt} & 35.97 &	16.57 &	38.39 &	73.11 &	40.38 &	30.12 &	57.35 &	7.84 &	37.47 \\
\texttt{gpt-4o (2024-08-06)}~\citep{hurst2024gpt} & 43.04 &	43.44 &	53.47 &	73.97 &	54.23 &	30.12 	 & 58.16 &	33.53 &	48.75 \\
\texttt{gpt-4.1-mini (2025-04-14)}~\citep{gpt-4-1} & 62.79 &	63.44 &	69.75 &	84.36 &	54.23 &	47.46 &	63.47 &	48.28 &	61.72 \\
\texttt{gpt-4.1 (2025-04-14)}~\citep{gpt-4-1} & 69.36 &	64.75 &	\textbf{74.97} &	\textbf{86.23} &	57.63 &	\textbf{52.95} &	67.35 &	\textbf{87.93} &	\textbf{70.15} \\
\texttt{gpt-o4-mini (2025-04-16)}~\citep{o4-mini}  & \textbf{76.45} &	\textbf{70.16} &	74.47 &	78.45 &	\textbf{59.32} &	42.94 &	\textbf{73.47} &	50.07 &	65.67 \\
$^\ddagger$\texttt{codex-mini (2025-05-16)}~\citep{chen2021evaluating}	& 67.30 &	64.75 &	74.57 &	82.49 &	58.76 &	48.78 & 67.64 & 58.76 & 65.38 \\
\midrule
\multicolumn{10}{l}{\textit{OSS (Base Size): < 10B parameters}} \\
\midrule
\texttt{Qwen3-1.7B}~\citep{qwen3} & 20.12	& 10.62 &	15.08 &	46.24 & 16.95 &	15.38 &	6.12 &	1.87 & 16.55 \\
\texttt{Qwen3-4B}~\citep{qwen3} & 27.23 &	30.77	& 28.85 &	52.80	& 15.25	& 19.16	&	20.41 & 23.85  &	27.29\\
\texttt{gemma-3-4b-it}~\citep{team2025gemma} & 27.36 &	29.10	& 24.52	& 42.49	& 18.64 &	17.95 &	8.16	& 4.37	& 21.57\\
$^\vee$\texttt{medgemma-4b-it}~\citep{medgemma-hf} & 15.51	& 13.11	& 14.85	& 41.89	&17.62	& 26.74 &	17.82 &	1.33	& 18.61 \\
\texttt{Qwen3-8B}~\citep{qwen3} & \textbf{29.08}	& \textbf{34.53}	& \textbf{37.37}	& \textbf{54.59}	& 20.34	&20.51	&	\textbf{24.49} & \textbf{25.71} & \textbf{30.83} \\
\texttt{Qwen2.5-7B-Instruct}~\citep{yang2024qwen2} & 13.08	& 15.57	& 12.76	& 25.91	& \textbf{30.36} &	21.79	& 10.20	& 5.42	& 17.43\\ 
\texttt{Llama-3.1-8B-Instruct}~\citep{dubey2024llama} & 16.67	& 25.00	&19.17 &	27.53	& 16.95	& 18.59	& 9.19	& 2.36	& 16.97\\
\texttt{Ministral-8B-Instruct-2410}~\citep{Ministral-8b} & 16.70	& 14.92	& 25.39	& 49.81	& 22.03	& 23.72 &	12.24 &	7.79	& 22.27\\ 
% \texttt{Mistral-Nemo-Instruct-2407}~\citep{adler2024nemotron} \\ 
% coding
$^\ddagger$\texttt{Qwen2.5-Coder-7B-Instruct}~\citep{hui2024qwen2} & 9.12	& 10.66 &	15.63 &	24.62 &	18.75 &	10.60 &	17.24 &	10.55 &	14.65\\
% medical
$^\vee$\texttt{HuatuoGPT-o1-7B}~\citep{chen2024huatuogpt} & 4.99 & 7.04	& 7.04 & 38.05 & 18.64 & 28.21 & 19.88	& 5.03 & 16.11 \\
$^\vee$\texttt{m1-7B-23K}~\citep{huang2025m1} & 6.88	& 9.56 &	7.04 &	28.24 &	9.32	& 20.26 &	14.71 &	0.00 & 12.00\\
% $^\vee$\texttt{ClinicalGPT-R1}~\citep{lan2025clinicalgpt} \\
$^\vee$\texttt{MedReason-8B}~\citep{wu2025medreason} & 9.12	 & 9.51	& 9.15	& 43.31 & 21.46	& \textbf{31.42}	& 17.42	& 3.88	& 18.16\\
% $^\vee$\texttt{MedS$^3$}~\citep{jiang2025meds3medicalsmalllanguage} \\
% \rowcolor{Goldenrod!16} \textbf{\ours{} (Qwen3-1.7B)}  \\
% \rowcolor{Dandelion!16} \textbf{\ours{} (Qwen3-4B)}  \\
% \rowcolor{Peach!16} \textbf{\ours{} (Qwen2.5-7B-Instruct)} & 64.13	& 66.91	& 72.02	& 90.06	& 52.54	& 34.62	& 69.39	& 29.55 &	59.90 \\
% \rowcolor{RedOrange!16} \textbf{\ours{} (Qwen3-8B)}  \\
\midrule
\multicolumn{10}{l}{\textit{OSS (Large Size): 10 - 30B parameters}}  \\
\midrule
% \texttt{gemma-3-12b-it}~\citep{team2025gemma} \\
\texttt{Qwen3-14B}~\citep{qwen3} & 31.50 &	31.97	& 30.05	& \textbf{61.38}	& 22.03	& 22.60 & \textbf{26.53} & {26.77}	& 31.60 \\
\texttt{Qwen2.5-14B-Instruct}~\citep{yang2024qwen2} & 17.21	& 14.07	& 16.43 &	27.40	& \textbf{35.59}	& \textbf{29.49}	& 16.33 & 4.45	 &	20.12\\
\texttt{DeepSeek-R1-Distill-Qwen-14B}~\citep{guo2025deepseek} & 35.12	& 38.52 &	32.96	& 48.09 &	32.20	& 21.29	& 24.49	& 11.39	& 30.51\\
% coding
$^\ddagger$\texttt{Qwen2.5-Coder-14B-Instruct}~\citep{hui2024qwen2} & \textbf{41.82} & \textbf{44.26} &	\textbf{35.78}	& 33.75 &	30.42 &	26.28 &	22.45 & \textbf{28.37} &	\textbf{32.89}\\
% $^\ddagger$\texttt{DeepSeek-Coder-V2-Lite-Instruct}~\citep{zhu2024deepseek} \\
% medical
$^\vee$\texttt{Baichuan-M1-14B-Instruct}~\citep{wang2025baichuan} & 4.50	& 12.19	& 7.36	& 1.82 & 21.46 & 16.34 &	17.42 & 0.00 & 10.14 \\
% $^\vee$\texttt{BioMedGPT-R1}~\citep{luo2024biomedgpt} \\
% \rowcolor{SeaGreen!16} \textbf{\ours{} (Qwen2.5-14B-Instruct)} & 64.54	& 63.52 & 76.08 & 92.45 & 54.32 & 43.56 & 92.96 & 43.56 & 66.37\\
% \rowcolor{JungleGreen!16} \textbf{\ours{} (Qwen3-14B)}  \\
\midrule
\multicolumn{10}{l}{\textit{OSS (XL Size): > 30B parameters}}  \\
\midrule
\texttt{Qwen3-32B}~\citep{qwen3} & 52.48	& 60.95	& 53.82	& 63.82 &	45.93	& 32.67 & 28.57		& 47.29	& 48.19 \\
\texttt{Qwen2.5-32B-Instruct}~\citep{yang2024qwen2} & 54.56	& 45.41	& 62.81	& 69.96	& 40.67	& 27.45	& 22.45	& 18.13	& 42.68\\
\texttt{QwQ-32B}~\citep{qwq32b} & 62.31	& 56.72	& \textbf{66.15} & 67.69	& \textbf{47.46}	& \textbf{42.31}	& 14.29	& \textbf{55.05}	& \textbf{51.50}\\
\texttt{DeepSeek-R1-Distill-Qwen-32B}~\citep{guo2025deepseek} & 62.18	& 58.36	& 65.82 &	60.14 &	43.56 &	28.66 & 26.53 &	31.17 & 47.05 \\
\texttt{Llama-3.3-70B-Instruct}~\citep{dubey2024llama} &39.93	&25.08	&24.98	&\textbf{84.99}	& 39.40	& 27.55 & 24.49	& 29.93	& 37.04\\ 
\texttt{DeepSeek-R1-Distill-Llama-70B}~\citep{guo2025deepseek} & \textbf{64.59}	& \textbf{64.92}	& 56.98	& 76.96 &	28.81 &	32.05 & \textbf{42.86} & 33.42 & 50.07\\ 
\bottomrule
\end{tabular}
}
\label{tab:main_results}
% \vspace{-3ex}
\end{table}

%% file: table/tab-training.tex
\begin{table}[t]
\centering
\renewcommand\arraystretch{0.92}
\caption{\ours{} performance on \method{} finetuned with sampled trajectories.
}
% \vspace{-1ex}
\resizebox{1.01\linewidth}{!}{ %
\begin{tabular}{l | c c c| c c| c c| c | c c }
\toprule
% & \multicolumn{9}{c|}{{In-distribution}} & \multicolumn{6}{c}{{Out-of-distribution}} \\
% \cmidrule(lr){2-10} \cmidrule(lr){11-16} 
\textbf{Datasets ($\rightarrow$)} &  {\textbf{MIMIC-III}} & {\textbf{eICU}} & {\textbf{TREQS}} & {\textbf{MedCalc.}} & \textbf{MedAgent.} & 
 \textbf{BioCoder} & \textbf{BioDS.} & \textbf{EHRSHOT} & \textbf{Avg.} & \textcolor{OliveGreen}{\textbf{$\Delta$}} \\
\textbf{Base ($\downarrow$) / Metrics ($\rightarrow$)} & SR & SR & SR & SR & SR & SR  & SR & Acc & \multicolumn{2}{c}{Score} \\ \midrule
% \texttt{Qwen3-1.7B}  \\
% \quad +SFT  & & & & & & & & & & \\
% \texttt{Qwen3-4B}  \\
% \quad +SFT  & & & & & & & & & & \\
% \texttt{Qwen3-8B}  \\
% \quad +SFT  & & & & & & & & & & \\ 
\rowcolor{Gray!16} \texttt{Qwen2.5-7B-Instruct} & 13.08 & 15.57 &	12.76	& 25.91	& 30.36	& 21.79 & 10.20 & 5.42 & 16.89 & --\\
\quad +SFT  & 57.83 & 61.48	& {72.66}	& 89.06	& 50.85	& 28.33	& 55.10	& 15.62 & 53.87 & \textcolor{OliveGreen}{(\emph{+36.98})}\\
% \quad +DPO  & 49.59 & 43.61	& 46.68	& 49.20	& 45.25	& 30.13	& 69.39	& 26.43	& 45.04 & \textcolor{OliveGreen}{(\emph{+28.15})}\\
\quad \quad +DPO  & {64.13} & {66.91} & 72.02 & {90.06} & {52.54} & {34.62} & {69.39} & {29.55}	& {59.90} & \textcolor{OliveGreen}{(\emph{+43.02})}\\
% +iDPO \\
% \midrule
% \hdashline
\quad \quad +PPO  & 66.10	& 67.25	& \textbf{73.88} & 74.52	& 51.33	& 32.71	& 65.47	& 32.40 & 57.96 & \textcolor{OliveGreen}{(\emph{+41.07})}\\
\rowcolor{Peach!16} \quad \quad +GRPO  & \textbf{68.21} & \textbf{68.73} & {70.50}	& \textbf{92.33}	& \textbf{55.87}	& \textbf{37.40}	& \textbf{71.11} & \textbf{33.18} & \textbf{62.17} & \textcolor{OliveGreen}{(\emph{+45.28})}\\
\midrule
\rowcolor{Gray!16} \texttt{Qwen2.5-14B-Instrust} & 17.21 & 14.07	& 16.43	& 27.40	& 35.59	& 29.49	& 16.33	& 4.45	& 20.12 & --\\
\quad +SFT  & 61.45	& 62.46	& {76.38}	& {94.36}	& 52.54	& 39.80	& 89.80	& 34.58 & 63.92 & \textcolor{OliveGreen}{(\emph{+43.80})} \\
% \quad +DPO  & 57.49	 & 59.18 & 70.45 & 71.32 & 47.46 & 42.95 & 91.84 & 41.33 &	60.25 & \textcolor{OliveGreen}{(\emph{+40.13})} \\
\quad \quad +DPO  & {64.54} & {63.52} & 76.08 & 92.45 & {54.32} & {43.56} & {92.96}	& {43.56} &	{66.37} & \textcolor{OliveGreen}{(\emph{+46.25})}\\
% \rowcolor{JungleGreen!16} {(Qwen3-14B)}  \\
% +iDPO \\
% \midrule
% \hdashline
\quad \quad +PPO  & 67.55 & 68.53 & \textbf{78.32} & 94.86 & 53.22 & 45.88 & 91.33 & 56.79 & 69.56  & \textcolor{OliveGreen}{(\emph{+49.44})}\\
\rowcolor{SeaGreen!16} \quad \quad +GRPO & \textbf{68.78} & \textbf{69.34} & {76.84} & \textbf{95.81} & \textbf{57.41} & \textbf{49.32} & \textbf{94.78} & \textbf{59.05} & \textbf{71.42} & \textcolor{OliveGreen}{(\emph{+51.30})}\\
\bottomrule
\end{tabular}%
% \vspace{-6ex}
}
\label{tab:training_results}
\end{table}

%% file: table/tab-ood.tex
\begin{wraptable}[18]{r}{0.62\linewidth}
\centering
\caption{External test set results on \method{}.
}
\resizebox{\linewidth}{!}{ %
\begin{tabular}{l | c c c c | c}
\toprule
\textbf{Datasets ($\rightarrow$)} & \textbf{EHR-SeqSQL} & \textbf{EHRCon}  & \textbf{MIMIC-Extract} & \textbf{N-PowerAI} & \textbf{Avg.} \\
\textbf{Base ($\downarrow$) / Metrics ($\rightarrow$)} & SR & SR & Acc  & SR & Score \\ \midrule
\multicolumn{6}{l}{\textit{API-based Proprietary LLMs$^\dagger$ (for reference)}} \\ \midrule
\texttt{gpt-4o-mini}~\citep{hurst2024gpt} & 50.80	& 23.20	& 2.67 & 16.03 & 26.03\\
\texttt{gpt-4o}~\citep{hurst2024gpt} & 58.40 & 35.79 & 9.82 & 20.71	& 34.69 \\
\texttt{gpt-4.1-mini }~\citep{gpt-4-1} & 70.60	& 52.40	& 5.62 & 25.66	& 43.20	\\
\texttt{gpt-4.1}~\citep{gpt-4-1} & 78.20 & \textbf{63.00} & 10.41 & 33.53 &	51.06	\\
\texttt{gpt-o4-mini}~\citep{o4-mini} & \textbf{100.00} & 51.00 & \textbf{16.88} & \textbf{36.15} & \textbf{53.94}\\
\midrule
\multicolumn{6}{l}{\textit{OSS LLMs}} \\
\midrule
\texttt{Qwen3-1.7B}~\citep{qwen3} & 33.60 & 17.20 & 1.90 & 14.72 & 16.86 \\
\texttt{Qwen3-4B}~\citep{qwen3} & 44.80 & 26.20	& 4.59 & 19.30 & 23.72\\
\texttt{Qwen3-8B}~\citep{qwen3} & 52.00 & 31.40	& 6.82 & 20.12 & 27.59\\
$^\vee$\texttt{HuatuoGPT-o1-7B}~\citep{chen2024huatuogpt} & 33.25 & 19.80 & 2.11 & 12.45 & 16.90 \\
\rowcolor{Gray!16} \texttt{Qwen2.5-7B-Inst}~\citep{yang2024qwen2} & 42.20 & 27.20 &	1.34 & 11.66 & 20.60\\ 
\textbf{\ours{} (SFT, 7B)}  & 42.40 &	28.80	& 1.95	& 10.48 &	20.91\\
% \textbf{\ours{} (DPO, 7B)} & 42.60	& 26.20 &	4.19	& 12.23 &	21.31\\
\textbf{\ours{} (DPO, 7B)} & 43.40 & 23.00 &	2.14 &	14.82 & 20.84 \\
\textbf{\ours{} (PPO, 7B)}  & 45.60	&24.40	& 4.30	& 17.19	&22.87\\
\rowcolor{Peach!16} \textbf{\ours{} (GRPO, 7B)} & 61.25	&46.80	&10.80	&27.65	&36.63\\
\texttt{Qwen3-14B}~\citep{qwen3} & 69.00 & 45.00 & 9.24	& 23.59 &	36.71\\
\texttt{Qwen2.5-Coder-14B-Inst}~\citep{hui2024qwen2} & 52.40 & 42.00 & 6.77 & 28.95 & 32.53 \\
\rowcolor{Gray!16} \texttt{Qwen2.5-14B-Inst}~\citep{yang2024qwen2} & 46.40	& 39.20 &	4.51 & 21.57 & 27.92\\
\textbf{\ours{} (DPO, 14B)} & 42.20 & 40.80 &	2.75 & 25.89	& 27.91\\
\textbf{\ours{} (PPO, 14B)} & 66.40 & 43.70 & 7.15 & 32.01 & 37.32\\
\rowcolor{SeaGreen!16} \textbf{\ours{} (GRPO, 14B)} & \textbf{72.80} & \textbf{56.60} & \textbf{14.91} & \textbf{43.77} & \textbf{47.02} \\
\texttt{R1-Dis-Qwen-14B}~\citep{guo2025deepseek} & 56.00 &	40.80	& 2.37 & 17.60 & 29.19\\
\texttt{Qwen3-32B}~\citep{qwen3} & 64.80 & 54.40 & 12.17 & 31.26	& 42.16\\
\bottomrule
\end{tabular}
}
\label{tab:ood_results}
\end{wraptable}

%% file: section/5-conclusion.tex
\section{Conclusion}
\label{sec:conclusion}

We present \method{}, an executable, privacy‑preserving, and extensible training environment for scaling code‑based biomedical reasoning in LLM agents. 
% \method{} provides (1) isolated, reproducible execution with domain‑specific dependencies, (2) interactive feedback that converts compiler and runtime errors into actionable debugging signals, and (3) scalable trajectory generation that supports both offline and online reinforcement learning. 
With 72K task instances across 129 categories, \method{} enables comprehensive benchmarking of 29 proprietary and OSS LLMs for biomedical data science within a modular, decoupled architecture that supports flexibility and extensibility.
\ours{} further demonstrates that systematic training and trajectory sampling with \method{} improve coding proficiency for biomedical data science tasks. 
% By unifying diagnostic evaluation with an interactive training ground, \method{} serves both as a tool to reveal limitations in medical computation and as a platform to train more capable, privacy‑preserving biomedical coding agents.
\method{} has the potential to accelerate progress from structured medical information retrieval tasks toward more open‑ended computational research questions in clinical research and biomedical discovery.

%% file: section/6-appendix.tex
\newpage
\section{Limitations and Broader Impacts}
\label{app:limitation}
\subsection{Limitations}
% \noindent \textbf{Resource Limitations.} 
Although \method{} demonstrates strong empirical performance improvement in a wide range of coding-aided biomedical reasoning tasks, several limitations remain.
Firstly, \method{} requires substantial computational resources for trajectory sampling, model fine-tuning, and iterative self-improvement procedures. Although we achieve significant improvements with relatively lightweight OSS LLMs, further scaling and advanced RL methods require increased computing infrastructures, limiting accessibility for resource-constrained research groups.
Secondly, our current dataset size and trajectory collection are primarily constrained by computational budget rather than data availability, potentially limiting the full exploration of model scaling behavior.
Thirdly, \method{} primarily supports text and structured data modalities. Future extensions will incorporate multimodal biomedical data (\eg, medical imaging, EEG, audio or video signals), enabling a richer and more comprehensive evaluation of multi-modal reasoning capabilities. Achieving effective multi-modal integration, however, presents significant challenges in data collection, curation, and standardized evaluation frameworks.

% \noindent \textbf{Multi-turn Reinforcement Learning.}
% Our current RL framework primarily addresses single-turn or limited multi-turn interactions. In practice, realistic clinical workflows often involve extended multi-turn reasoning. Extending the framework to robustly support complex multi-turn interactions will require more advanced RL methods (\eg, PPO or GRPO), improved exploration strategies, and carefully designed reward structures to manage long-term dependencies effectively.

\subsection{Broader Impacts}
\noindent \textbf{Potential Positive Societal Impacts.}
\method{} can significantly enhance the development of accessible, affordable, and privacy-preserving AI tools for clinical decision-making. Improved coding-based biomedical reasoning capabilities in open-source LLM agents (\eg, \ours{}) have the potential to democratize access to advanced computational healthcare assistance, benefiting clinicians, researchers, and healthcare systems globally, particularly in resource-limited settings. The plug-and-play architecture also allows continuous adaptation to new medical knowledge and practices, fostering sustainable and community-driven innovation in healthcare technology.

\noindent \textbf{Potential Negative Societal Impacts.}
Despite the benefits, the introduction and widespread deployment of sophisticated computational frameworks like \method{} may unintentionally widen existing healthcare inequities. Institutions with limited computational resources (including both Microsoft Azure API service and high-performance computing clusters) or inadequate data infrastructure may struggle to access or fully benefit from these technological advancements, potentially exacerbating disparities in healthcare capabilities across regions or socioeconomic groups. Moreover, reliance on publicly available datasets may perpetuate existing biases due to uneven data representation, potentially disadvantaging underrepresented patient populations and rare disease conditions.

\subsection{Privacy Statements}
\label{app:privacy}
\input{table/tab-appendix-data-access}
\noindent \textbf{Data Privacy and Licensing.} We carefully curated \method{} with strict adherence to ethical standards, using publicly available datasets or datasets with appropriate privacy protections and anonymizations. Table \ref{tab:data-access} lists the access requirements for the 12 datasets in \method{} and the code base for data processing or task implementation. We explicitly designed isolated Docker environments to ensure data privacy and security. Nevertheless, ethical usage of our methods and models in clinical settings requires rigorous validation, transparency in limitations, and close collaboration with healthcare professionals. We encourage responsible deployment, emphasizing human oversight, continuous evaluation, and clear communication of model capabilities and uncertainties to mitigate ethical and practical risks.

\noindent \textbf{LLM Usage Statement.}
In compliance with the PhysioNet Credentialed Health Data Use Agreement (version 1.5.0)\footnote{\url{https://physionet.org/about/licenses/physionet-credentialed-health-data-license-150/}}, we strictly prohibit transferring confidential patient data (\eg, MIMIC-III and eICU) to third-party entities, including external online services and APIs. To responsibly utilize the Azure OpenAI Service, we adhere closely to PhysioNet's guidelines on responsible GPT usage\footnote{\url{https://physionet.org/news/post/gpt-responsible-use}}. Specifically, we have opted out of the human review process by completing the Azure OpenAI Additional Use Case Form\footnote{\url{https://aka.ms/oai/additionalusecase}}, thereby ensuring no third-party entity accesses or processes sensitive patient information. We consistently monitor our data handling practices and strictly adhere to applicable guidelines and privacy regulations, maintaining the highest ethical standards in our research and operations.

\input{section/2-related-works-v2}

\input{appendix/app-data}
\input{appendix/app-baseline}
\input{appendix/app-implementation}

\input{appendix/app-exp}

\input{appendix/app-prompt}

%% file: table/tab-appendix-data-access.tex
\begin{table}[ht]
    \centering
    \renewcommand\arraystretch{0.98}
    \caption{Data Access and License Information of 12 datasets in \method{}. ``Custom'' represents additional dataset- or task-specific license and data access requirements (\eg, DUA or credentials).}
    \resizebox{1.01\linewidth}{!}{ %
    \begin{tabular}{l|cccc}
        \toprule
       \textbf{Dataset} &  \textbf{Data License} & \textbf{Data Access} & \textbf{Code License} & \textbf{Code Access}\\
        \midrule
        \multicolumn{5}{l}{\textit{Training and Internal Validation (In-Distribution)}} \\ \midrule
        MIMIC-III~\citep{johnson2016mimic,lee2022ehrsql}  & Custom & \href{https://physionet.org/content/mimiciii/1.4/}{MIMIC-III on PhysioNet}  & CC-BY-4.0 & \href{https://github.com/glee4810/EHRSQL}{MIMIC-III on EHRSQL} \\
        eICU~\citep{pollard2018eicu,lee2022ehrsql}  & Custom &	\href{https://physionet.org/content/eicu-crd/2.0/}{eICU on PhysioNet} & CC-BY-4.0 & \href{https://github.com/glee4810/EHRSQL}{eICU on EHRSQL}  \\
        TREQS~\citep{wang2020text} & Custom & \href{https://physionet.org/content/mimiciii/1.4/}{MIMIC-III on PhysioNet} & MIT & \href{https://github.com/wangpinggl/TREQS}{TREQS on GitHub}\\
        MedCalcBench~\citep{khandekar2024medcalc} & CC-BY-SA 4.0 &  \href{https://github.com/ncbi-nlp/MedCalc-Bench/tree/main}{MedCalcBench} & Public & \href{https://github.com/ncbi-nlp/MedCalc-Bench/tree/main}{MedCalcBench on GitHub}\\ 
        MedAgentBench~\citep{jiang2025medagentbench} &	MIT & \href{https://hub.docker.com/r/jyxsu6/medagentbench}{MedAgentBench (FHIR Server)} & MIT  & \href{https://github.com/stanfordmlgroup/MedAgentBench}{MedAgentBench on GitHub} \\
        BioCoder~\citep{tang2024biocoder}  & CC-BY-4.0	& \href{https://huggingface.co/datasets/lilbillbiscuit/biocoder_public}{BioCoder on Huggingface} & N/A & \href{https://github.com/gersteinlab/BioCoder}{BioCoder on GitHub}\\
        BioDSBench~\citep{wang2024can} &	MIT & \href{https://github.com/RyanWangZf/BioDSBench}{BioDSBench} & MIT & \href{https://github.com/RyanWangZf/BioDSBench}{BioDSBench on GitHub} \\
        EHRSHOT~\citep{wornow2023ehrshot}  & Custom &	\href{https://redivis.com/datasets/53gc-8rhx41kgt}{EHRShot (Standford)} & Apache & \href{https://github.com/som-shahlab/ehrshot-benchmark}{EHRSHOT on Github}  \\ \midrule

        \multicolumn{5}{l}{\textit{External Validation (Out-of-Distribution)}} \\ \midrule
        EHR-SeqSQL~\citep{ryu2024ehr} & Custom & \href{https://physionet.org/content/mimiciii/1.4/}{MIMIC-III on PhysioNet} & N/A & \href{https://github.com/seonhee99/EHR-SeqSQL}{EHR-SeqSQL on GitHub}\\
        EHR-Con~\citep{kwon2024ehrcon} & Custom & \href{https://physionet.org/content/mimiciii/1.4/}{MIMIC-III on PhysioNet} & MIT & \href{https://github.com/dustn1259/EHRCon}{EHR-Con on GitHub} \\
        MIMIC-Extract~\citep{MIMICExtract} & Custom & \href{https://mimic.mit.edu/docs/gettingstarted/cloud/}{MIMIC-III on PhysioNet } & MIT & \href{https://github.com/MLforHealth/MIMIC_Extract}{MIMIC-Extract on GitHub}\\
        N-PowerAI~\citep{ruan2025n} & N/A & \href{https://www.biorxiv.org/content/10.1101/2025.02.06.636776v1.supplementary-material}{N-Power AI Supp. Mat.} & N/A & \href{https://ai.swmed.edu/N-PowerAI/}{N-Power AI on Webpage}\\
        \bottomrule
    \end{tabular}
    }
    \label{tab:data-access}
\end{table}

%% file: section/2-related-works-v2.tex
\section{Additional Related Works}
\label{app:related}

\begin{wrapfigure}{r}{0.36\textwidth}
    \centering
    \includegraphics[width=\linewidth]{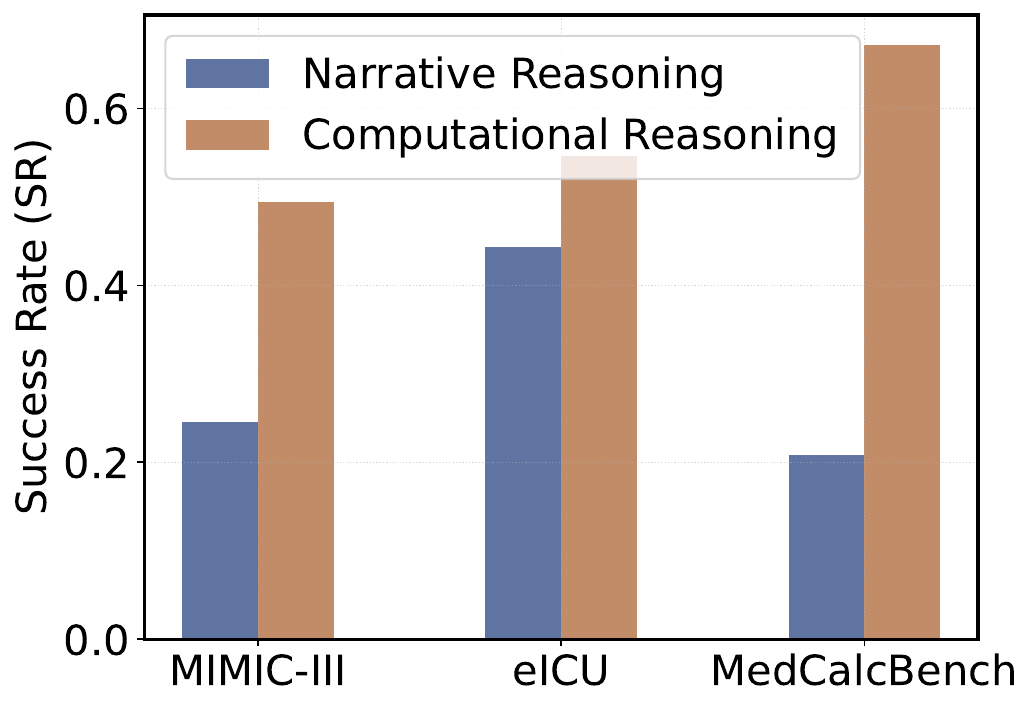}
    \caption{{Coding empowers computational medical reasoning (w/ \texttt{gpt-4-turbo}).\vspace{-1ex}}}
    \label{fig:medcode}
\end{wrapfigure}

\textbf{Medical Agents (Coding).}
Recent advances have demonstrated that LLMs exhibit strong capabilities in medical reasoning and planning leveraging extensive biomedical knowledge~\citep{singhal2023large,moor2023foundation,lievin2024can}, fueling increased interest in developing LLM-based autonomous agents tailored specifically for medical tasks~\citep{jin2024agentmd,gao2025txagent,li2024mmedagent,liao2024reflectool,tang2024medagents,Kim2024MDAgents}. 
In particular, LLM-based agents have shown promise in specialized computational tasks, including querying EHR databases~\citep{shi2024ehragent}, performing bio-statistical calculations~\citep{ruan2025n}, and conducting bioinformatics analyses~\citep{tang2024biocoder,wang2024can,tayebi2024large}.
As shown in Figure~\ref{fig:medcode}, integrating coding capabilities into LLM-based agents further enhances performance on tasks traditionally approached through natural language reasoning (\eg, MIMIC-III, eICU~\citep{lee2022ehrsql}), as well as numerical and rule-based medical reasoning (\eg, MedCalcBench~\citep{khandekar2024medcalc}).
However, existing coding-based medical agents rely primarily on prompt engineering without systematic improvement, limiting their robustness and scalability when addressing complex and diverse coding tasks in real-world biomedical scenarios.
In contrast, \method{} specifically targets reasoning-intensive coding tasks by introducing a unified, scalable, and interactive training environment that systematically improves the coding-based medical reasoning capabilities of LLM agents.

\textbf{Medical Reasoning Models.}
Recent advancements have substantially improved biomedical reasoning capabilities of LLMs through RL~\citep{huang2025m1,lai2025med,zhang2025med,jiang2025meds3medicalsmalllanguage,wu2025medreason,chen2024huatuogpt,lan2025clinicalgpt,wang2025baichuan,li2025aligning,zhang2025med,miao2025paper2agent,jin2025stella,yu2025medreseacher,zhi2025medgr,liu2025beyond}.
For example, M1~\citep{huang2025m1} improves by distilling knowledge from the reasoning traces generated by DeepSeek-R1~\citep{guo2025deepseek}.
MedS3~\citep{jiang2025meds3medicalsmalllanguage} employs Monte Carlo Tree Search (MCTS) to generate rule-verifiable reasoning trajectories and employs process-reward models to select optimal reasoning paths during inference. 
Similarly, HuatuoGPT-o1~\citep{chen2024huatuogpt} and ClinicalGPT-R1~\citep{lan2025clinicalgpt} integrate domain-specific verifiers to guide RL fine-tuning processes for improved clinical reasoning. 
Extending beyond language modeling, Med-R1~\citep{lai2025med} and MedXpertQA~\citep{zuo2025medxpertqa} adapt RL methodologies to vision-language models, effectively addressing medical visual question answering tasks. 
Despite these developments, current medical reasoning models predominantly target natural language-based reasoning, with limited attention given to coding-intensive scenarios common in biomedical research and clinical practice.

% % Huatuo-o1~\citep{chen2024huatuogpt}
% % Baichuan-M1~\citep{wang2025baichuan}
% % github: BioMedGPT-R1~\citep{luo2024biomedgpt}
% % m1~\citep{huang2025m1}
% % MedS3~\citep{jiang2025meds}
% % ClinicalGPT-R1~\citep{lan2025clinicalgpt}
% % medreason~\citep{wu2025medreason}: knowledge graph
% % ALFA~\citep{li2025aligning}
% % VLMs:
% % Med-R1~\citep{lai2025med} 
% % Med-RLVR~\citep{zhang2025med}
% % MedXpertQA~\citep{zuo2025medxpertqa}

\textbf{Medical Reasoning Benchmarks.}
% Table~\ref{tab:related} summarizes representative medical reasoning benchmarks, many of which primarily assess LLM performance through closed-form medical QA tasks. 
Most existing medical reasoning benchmarks focus primarily on evaluating LLM performance through closed-form medical QA tasks~\citep{pal2022medmcqa,jin2021disease,jin-etal-2019-pubmedqa,tsatsaronis2015overview,tang2025medagentsbench,xiong2024benchmarking,healthbench}. 
In addition, AgentClinic~\citep{schmidgall2024agentclinic} further evaluates diagnosis prediction within simulated clinical scenarios, while MedHELM~\citep{medhelm} provides comprehensive evaluations in various medical NLP tasks.
Despite these extensive benchmarking efforts, existing benchmarks -- including recent concurrent works such as MedAgentBoard~\citep{zhu2025medagentboard}, HealthBench~\citep{healthbench}, and MedCaseReasoning~\citep{wu2025medcasereasoning} -- typically focus on evaluation scenarios, with limited emphasis on dedicated training environments aimed at systematically improving medical reasoning capabilities~\citep{thapa2025disentangling}, especially within coding-intensive and interactive medical scenarios.

\textbf{Medical Agent Training Environments.}
To advance medical agents with narrative reasoning, AgentClinic~\citep{schmidgall2024agentclinic} and AgentHospital~\citep{Li2024AgentHospital} simulate hospital workflows focused on diagnostic tasks, while MediQ~\citep{li2024mediq} offers interactive simulations designed for medical information retrieval. 
Beyond medicine, specialized environments have emerged for systematically evaluating and improving LLM agents across diverse tasks~\citep{zhao2025sirius,wang2025ragen}, such as software engineering~\citep{pan2024training,yang2024swe,yang2025swesmith}, reasoning \citep{reasoninggym}, web browsing~\citep{drouin2024workarena}, agent planning and collaboration~\citep{xi2024agentgym,shao2024collaborative}, data science~\citep{guo2024dsagent,jing2025dsbench,zhang2025datascibench,zhang2024benchmarking}, machine learning engineering~\citep{nathani2025mlgym,huang2023mlagentbench,chan2024mle,tang2023ml}, automated research~\citep{kang2024researcharena,schmidgall2025agentrxiv,schmidgall2025agentlaboratoryusingllm}, and scientific discovery~\citep{team2025novelseek,yuan2025dolphin}. 
Inspired by these interactive training frameworks, \method{} uniquely targets real-world biomedical scenarios, aiming to rigorously benchmark and systematically enhance coding-based biomedical reasoning capabilities of LLM agents.

% SWE: SWE-Agent, SWE-Bench, SWE-Gym, SWE-smith
% DS: DataSciBench, DSBench, DSEval
% MLE: MLAgentBench, ML-Bench, MLE-bench, MLGYM, MLE-dojo
% Research: ResearchArena, Agentrxiv, Agent laboratory, coelho2025deepresearchgym
% Scientific discovery: novelseek, dolphin,

%% file: appendix/app-data.tex
\section{Task and Data Details}
\label{app:data}

\subsection{Overview}
We refer a task as coding‑based biomedical reasoning when LLM agents write and run code whose execution yields a verifiable outcome in biomedical data science. This definition allows us to objectively verify the results while preserving the steps that agents actually take, allowing for training and analysis at the trajectory level. 

\begin{wrapfigure}{r}{0.4\textwidth}
    \centering
    \vspace{-2ex}
    \includegraphics[width=0.99\linewidth]{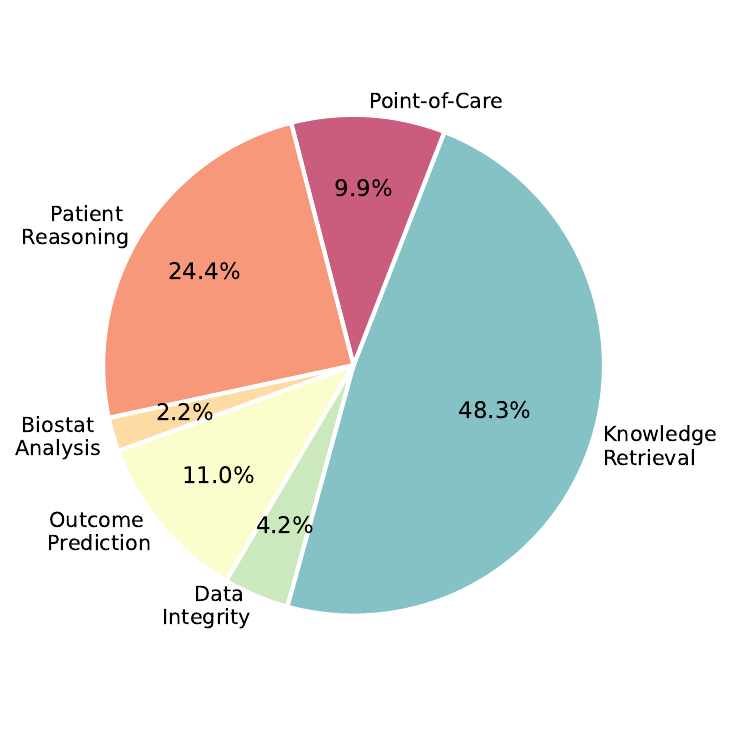}
    \caption{Diversity analysis.\vspace{-1ex}}
    \label{fig:data-clin}
\end{wrapfigure}
\noindent \textbf{Biomedical Application Category.} \method{} spans multiple biomedical subdomains, including \emph{Database queries} (DB, including MIMIC-III, eICU, TREQS, EHR-SeqSQL, and EHRCon), \emph{Data Analytics} (DA, including MedCalcBench and MedAgentBench), \emph{Bioinformatics} (Bioinfo, including BioCoder, BioDSBench, N-PowerAI), and \emph{Machine Learning} (ML, including EHRSHOT and MIMIC-Extract). 
% In several empirical analysis, we include ML as part of Bioinfo. category for better visualization.

Figure~\ref{fig:data-clin} illustrates the diverse task distribution within \method{}. Consider a clinician identifying patients at risk for sepsis from EHR data, a task requiring not only understanding of sepsis criteria but also SQL queries to extract relevant laboratory values, temporal logic to track patient trajectories, and statistical methods to validate findings. Similarly, researchers analyzing multi-omics data must integrate biological knowledge with bioinformatics algorithms and computational pipelines. These scenarios exemplify the core challenge of biomedical data science: operationalizing medical expertise through executable code, where domain knowledge alone proves insufficient without corresponding computational implementation.

% \begin{wrapfigure}{r}{0.24\textwidth}
%     \centering
%     \vspace{-2ex}
%     \includegraphics[width=\linewidth]{figure/in-out-dis.png}
%     \caption{Inter-distribution similarity.\vspace{-2ex}}
%     \label{fig.in-out-dis}
% \end{wrapfigure}

\noindent \textbf{Computational Task Category.} \emph{Structured tasks} primarily include database query scenarios, such as those from MIMIC-III, eICU, TREQS, EHR-SeqSQL, EHRCon, and MedCalcBench (rule- or equation-based), which require precise formulation of executable queries against structured EHR data. \emph{Open-ended tasks} include biomedical data analysis and medical coding scenarios drawn from datasets such as MedAgentBench, BioCoder, BioDSBench, EHRSHOT, MIMIC-Extract, and N-PowerAI, demanding nuanced and flexible code generation for complex analysis, statistical reasoning, or clinical decision-making. 

Specifically, we evaluate LLMs across eight biomedical coding domains: (1) clinical database querying (MIMIC-III, eICU, TREQS, EHRseqSQL), (2) clinical note analysis (EHRcon), (3) medical computation (MedCalcBench), (4) health information technology (MedAgentBench), (5) biomedical software engineering (Biocoder), (6) biomedical data analysis (BioDSBench), (7) biostatistics (NPowerAI), and (8) ML-based predictive modeling (EHRSHOT, MIMIC-Extract).

% Tasks span structured and unstructured medical information retrieval~\citep{lee2022ehrsql,ryu2024ehr,wang2020text,kwon2024ehrcon,pollard2018eicu}, numerical medical reasoning~\citep{khandekar2024medcalc}, health informatics and bioinformatics~\citep{jiang2025medagentbench,tang2024biocoder,wang2024can,ruan2025n}, and machine learning (ML)-based predictive modeling~\citep{wornow2023ehrshot,MIMICExtract}. 

\begin{wrapfigure}{r}{0.5\textwidth}
	\centering
    \vspace{-3ex}
 %    \subfigure[Biomedical Tasks]{
	% \includegraphics[width=0.34\linewidth]{figure/task-v3.pdf}
	% \label{fig:data-clin}
	% }
    \subfigure[Intra-Category]{
	\includegraphics[width=0.44\linewidth]{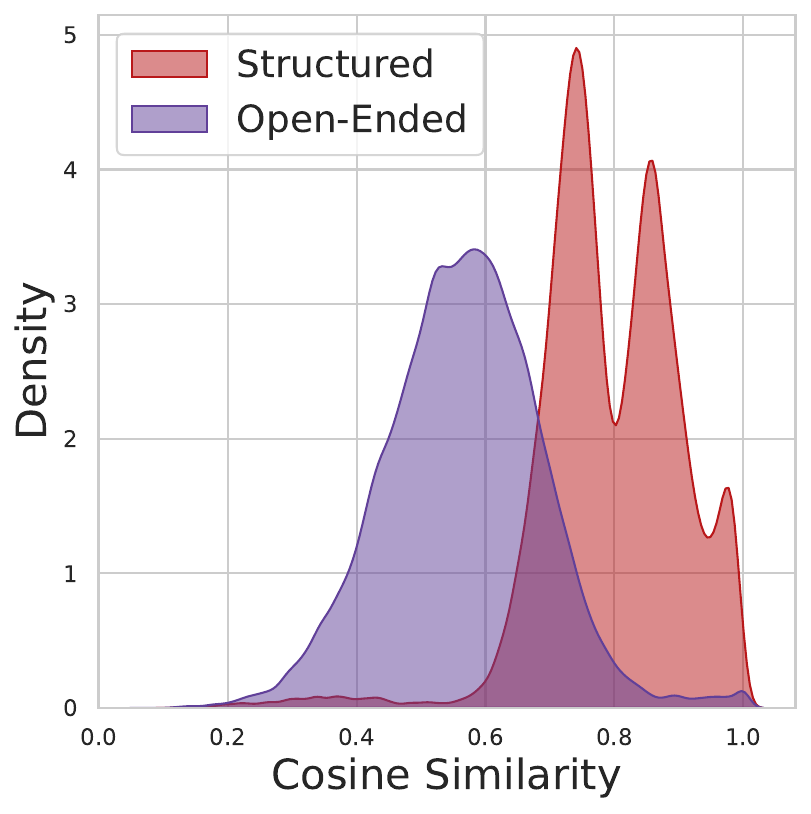}
	\label{fig:intra}
	} 
	\subfigure[Inter-Distribution]{
	\includegraphics[width=0.45\linewidth]{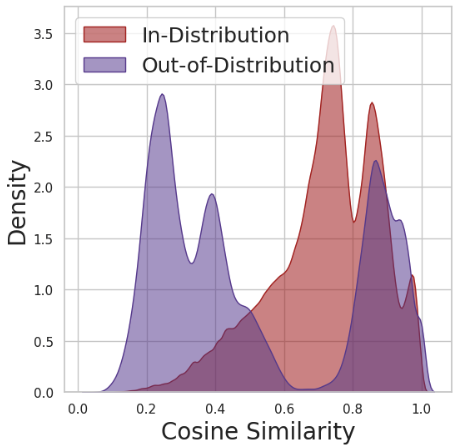}
	\label{fig:in-out-dis}
	}
    \vspace{-2ex}
	\caption{Similarity analysis}
\label{fig:dis}
\end{wrapfigure}
\noindent \textbf{In- \& Out-of-Distribution.} We further categorize tasks in \method{} into \emph{in-} and \emph{out-of-distribution}, facilitating a rigorous evaluation of model generalization and adaptability. To highlight intrinsic differences between these distributions, Figure~\ref{fig:in-out-dis} shows the distribution of sampled code trajectories. The resulting visualization demonstrates significant divergence in trajectory complexity, interaction frequency, and required code refinement steps between in-distribution and out-of-distribution tasks, underscoring the challenges posed by novel biomedical reasoning contexts. 

\subsection{Training and Internal Testing (In-Distribution) Dataset Details}
\noindent \textbf{EHRSQL: MIMIC-III and eICU.}
% \noindent \textbf{Task Description.}
EHRSQL~\citep{lee2022ehrsql} comprises text-to-SQL tasks that leverage electronic health records from MIMIC-III~\citep{johnson2016mimic} and eICU~\citep{pollard2018eicu}. They evaluate the ability of LLMs (and agents) to translate clinical questions posed by healthcare professionals into executable SQL queries. This includes handling complex queries involving temporal logic and conditional abstention.

% \noindent \textbf{License and Data Access.}
% The query-question dataset is publicly available under a Creative Commons Attribution 4.0 (CC BY 4.0) license. However, the underlying datasets, MIMIC-III (v1.4) and eICU (v2.0), require credentialed access from PhysioNet, which requires the completion of the training of human subjects of the CITI and the adherence to the PhysioNet Credentialed Health Data Use Agreement (DUA).

\noindent \textbf{TREQS.}
% \noindent \textbf{Task Description.}
TREQS~\citep{wang2020text} is a text-to-SQL benchmark tailored specifically to clinical question answering using the MIMIC-III dataset. It emphasizes generating accurate SQL queries from template-based natural language questions against a simplified schema comprising five core tables, with an emphasis on large result-set handling.

% \noindent \textbf{License and Data Access.}
% The TREQS dataset and associated implementation are released under an MIT License. However, practical usage necessitates credentialed access to the MIMIC-III database via PhysioNet, including human-subject research certification and acceptance of the data use agreement.

\noindent \textbf{MedCalcBench.} 
% \noindent \textbf{Task Description.}
MedCalcBench~\citep{khandekar2024medcalc} provides a structured evaluation of clinical calculation capabilities in LLMs. Each instance poses a patient-specific clinical scenario requiring precise medical calculations such as clinical scores or medication dosages, accompanied by expert-curated stepwise solutions for validation.

% \noindent \textbf{License and Data Access.}
% MedCalcBench is publicly accessible under a Creative Commons Attribution-ShareAlike 4.0 International (CC BY-SA 4.0) license. The dataset is openly available via HuggingFace without restrictions beyond proper attribution requirements.

\noindent \textbf{MedAgentBench.}
% \noindent \textbf{Task Description.}
MedAgentBench~\citep{jiang2025medagentbench} is a simulated EHR environment designed to evaluate LLM-driven clinical workflows. It features realistic patient scenarios across ten task categories, requiring agents to perform clinical reasoning, EHR querying via FHIR interfaces, and clinical decision support.

% \noindent \textbf{License and Data Access.}
% MedAgentBench, including its synthetic patient records and task definitions, is openly available under an MIT License. No real patient data is involved, thus obviating the need for credentialed access or data use agreements.

\noindent \textbf{BioCoder.}
% \noindent \textbf{Task Description.}
BioCoder~\citep{tang2024biocoder} assesses the capability of LLMs to generate accurate bioinformatics code solutions. It comprises practical coding challenges derived from authentic bioinformatics software, requiring the generation and verification of functionally correct Python methods.

% \noindent \textbf{License and Data Access.}
% BioCoder is publicly available under the Creative Commons Attribution 4.0 International (CC BY 4.0) license, accessible via GitHub and HuggingFace repositories without specific restrictions.

\noindent \textbf{BioDSBench.}
% \noindent \textbf{Task Description.}
BioDSBench~\citep{wang2024can} evaluates LLM proficiency in biomedical data science coding tasks, involving the generation of Python or R code to replicate analytical workflows derived from actual biomedical research studies. Tasks span statistical analyses, data manipulations, and visualization routines.

% \noindent \textbf{License and Data Access.}
% The benchmark materials are provided under an MIT License, with task descriptions openly accessible via GitHub and HuggingFace. Underlying datasets, sourced primarily from publicly available platforms (\eg, cBioPortal), require no specialized credentialing beyond standard citation compliance.

\noindent \textbf{EHRSHOT.}
% \noindent \textbf{Task Description.}
EHRSHOT~\citep{wornow2023ehrshot} benchmarks LLMs on few-shot clinical prediction tasks leveraging real-world, longitudinal, deidentified EHR data. It focuses on rapid adaptation to tasks such as risk prediction and forecasting clinical outcomes given limited labeled examples.

% \noindent \textbf{License and Data Access.}
% EHRSHOT dataset is released under the PhysioNet Credentialed Health Data License (version 1.5.0). Access is restricted to credentialed researchers who have completed the necessary human-subject research training and signed the PhysioNet data use agreement.

\subsection{External Evaluation (Out-of-Distribution) Dataset Details}

\noindent \textbf{EHR-SeqSQL.}
% \noindent \textbf{Task Description.}
EHR-SeqSQL~\citep{ryu2024ehr} extends text-to-SQL evaluation to sequential, multi-turn interactions, emulating realistic clinical dialogues. Tasks require maintaining context across multiple SQL queries, assessing LLM capability in handling compositional and contextual reasoning.

% \noindent \textbf{License and Data Access.}
% The question-SQL pairs are openly accessible for research purposes, though no explicit license is stated, presumed aligned with CC BY 4.0 based on predecessor datasets. However, meaningful utilization necessitates credentialed access to MIMIC-III and eICU databases from PhysioNet.

\noindent \textbf{EHRCon.}
% \noindent \textbf{Task Description.}
EHRCon~\citep{kwon2024ehrcon} involves assessing clinical note consistency with structured EHR records, focusing on identifying discrepancies. It serves as a verification task requiring precise alignment between unstructured clinical text and corresponding database entries.

% \noindent \textbf{License and Data Access.}
% The EHRCon data set is released via PhysioNet under the Credentialed Health Data License (version 1.5.0), requiring credentialed access, completion of CITI training, and adherence to the strict data use agreement of PhysioNet.

\noindent \textbf{MIMIC-Extract.}
% \noindent \textbf{Task Description.}
MIMIC-Extract\citep{MIMICExtract} provides structured, preprocessed time-series patient data derived from the MIMIC-III dataset, used in clinical predictive modeling such as mortality risk or intervention prediction, enabling standardized assessments of time-series reasoning capabilities.

% \noindent \textbf{License and Data Access.}
% The preprocessing pipeline code is released under an MIT License. However, actual extracted data remains governed by the PhysioNet Credentialed Health Data License, necessitating credentialed PhysioNet access for practical dataset utilization.

\noindent \textbf{N-PowerAI.}
% \noindent \textbf{Task Description.}
N-PowerAI~\citep{ruan2025n} evaluates LLM capabilities in performing statistical sample-size and power analyses for clinical trial design. It requires multi-step statistical reasoning and the generation of precise numeric results corresponding to various clinical scenarios.

% \noindent \textbf{License and Data Access.}
% N-PowerAI is available under the Creative Commons Attribution-NonCommercial-NoDerivatives 4.0 International (CC BY-NC-ND 4.0) license. For ground-truth generation, the N-PowerAI software is accessible via an online interface or source-code repository without any patient data involved, thus requiring no credentialed access or special agreements.

\subsection{Train-Test Set Split}
For datasets that provide predefined training, validation, and test splits, we combine the training and validation subsets into a single unified training set and retain the original test subset exclusively for evaluation. In cases where datasets lack predefined splits, we randomly allocate 50\% of the instances to training, assigning the remaining 50\% to the test set. 
For tasks containing more than $1000$ samples in both training and test sets, we create a lighter subset through downsampling to support efficient leaderboard-based training and evaluation. Specifically, we leverage task-specific metadata to perform uniform sampling within each fine-grained category, thereby maintaining diversity, ensuring balanced representation, and preserving the original data distribution.

\subsection{Data Pre-processing Details}
\subsubsection{Structured Tasks}

For database querying related datasets, including \textbf{MIMIC-III}, \textbf{eICU}, \textbf{TREQS}, and \textbf{EHR-SeqSQL}, each task instance is structured into a JSON format comprising: (1) the contextual description and the corresponding natural-language query, (2) the ground-truth SQL query, and (3) the resulting answer from the database execution. Instances yielding null results upon SQL execution, indicating the absence of a valid answer, are excluded from the dataset.

For \textbf{EHRCon}, we organize the data into structured databases that link patient records through hospital admission IDs, complemented by a separate database containing associated clinical notes. Each task is formulated as a JSON object consisting of: (1) admission ID, (2) relevant medical terminology, (3) count of detected inconsistencies, and (4) a binary indicator denoting the presence or absence of inconsistencies.

For \textbf{MedCalcBench}, each instance initially consists of a patient note, a specific medical calculation query, a ground-truth answer, and a detailed step-by-step solution. To accurately evaluate the coding capabilities of LLM agents without direct guidance, we remove all intermediate calculation hints, presenting only the patient note and the calculation query for model inference.

For \textbf{N-PowerAI}, statistical analysis tasks are augmented through attribute substitution. Specifically, each original instance is expanded 100-fold by systematically replacing an attribute with a randomly chosen equivalent from a predefined valid range, preserving the integrity and interpretability of the statistical context. Each augmented instance includes recalculated values for sample size (N) and statistical power, stored systematically within JSON-formatted records.

\subsubsection{Open-ended Tasks}

\textbf{MedAgentBench} instances require LLM agents to follow natural-language instructions to perform tasks within a FHIR-compliant interactive medical environment. We retain original instructions, solutions, and Medical Record Numbers (MRNs). To derive verifiable evaluation signals, we execute the provided ground-truth on the server-side environment to obtain authoritative reference answers.

\textbf{BioCoder} tasks require implementing biostatistics algorithms or addressing scientific programming challenges. Each instance comprises a problem description, context-specific code, test cases, and expected outputs. While evaluation datasets already contain all necessary components, training instances initially lack context-specific code and test cases. To address this gap, we employ the \texttt{o3-mini} model to auto-generate relevant context code and corresponding test cases based on provided ground-truth functions. Generated functions undergo rigorous validation via a code interpreter, retaining only verified, error-free instances. Additionally, we exclusively utilize the Python-based subset of BioCoder, deferring the JavaScript subset for subsequent integration.

\textbf{BioDSBench} instances involve biomedical data analysis tasks derived from real-world datasets. Features are systematically organized into directories by task, with each task's description and reference Python implementation captured within JSON structures.

For datasets dedicated to predictive model development (\eg, \textbf{EHRSHOT} and \textbf{MIMIC-Extract}), initial features are provided in pre-processed form but necessitate additional table joining, filtering, and integration to produce final training inputs. While labels accompany these tasks, explicit reference Python implementations are not provided, as evaluation metrics directly measure the accuracy of model predictions on predefined test subsets. Distinct subsets of training, validation, and testing data and labels are explicitly maintained and separately utilized for both training and evaluation phases.

\input{table/tab-traj}

\subsection{Sampled Trajectory Details}
\label{app:trajcompo}

Table~\ref{tab:traj} details the proportion of action types (\cref{subsec:setup}) in trajectories.
Structured tasks predominantly involve data retrieval (over 50\%) from databases or resources, complemented by coding and debugging steps. 
In contrast, open-ended tasks require significant coding and debugging efforts due to diverse question types, often necessitating terminal interactions to install specialized biomedical packages.
Although \method{} contains extensive training data and allows repeated sampling, the current trajectory count primarily reflects computational budget constraints.
Specifically, Figure~\ref{fig:scaling} (right) demonstrates consistent performance improvements with increasing training data volume, indicating that expanded trajectory sampling through additional computational resources would yield further gains.

%% file: table/tab-traj.tex
\begin{wraptable}{r}{0.45\linewidth}
    \centering
    \renewcommand\arraystretch{0.98}
    \caption{Trajectory Composition (\%).}
    \resizebox{\linewidth}{!}{ %
    \begin{tabular}{l|cccc}
        \toprule
       \textbf{Actions ($\rightarrow$)} & \textbf{request info}	& \textbf{terminal} & \textbf{code} & \textbf{debug} \\
        \midrule
        % \multicolumn{5}{l}{\textit{Template-based Tasks}} \\ \midrule
        MIMIC-III & 71.07 &	0 & 28.84 &	0.08  \\
        eICU & 72.17 &	0 &	27.13 &	0.70  \\
        TREQS &	64.27 &	0 & 35.54 & 0.19 \\
        MedCalc. &	0	& 0 & 	74.91	& 25.09  \\ \midrule
        \rowcolor{RoyalPurple!6} \textbf{Structured} & \textbf{51.88} &	\textbf{0} &	\textbf{41.61}	& \textbf{6.52}  \\
        \midrule
        % \multicolumn{5}{l}{\textit{Open-ended Tasks}} \\ \midrule
        MedAgent. &	0 &	0 & 100 &	0  \\
        BioCoder& 0	& 0.29	& 96.11	& 3.60\\
        BioDS. &	0 &	6.30 &	87.60 &	6.90 \\
        EHRSHOT &	0	& 0.43 &	59.43	& 40.14  \\ \midrule
        \rowcolor{RoyalPurple!6} \textbf{Open-ended}  & \textbf{0}& \textbf{1.76} &	\textbf{85.79} &	\textbf{12.46}  \\
        \midrule
        \rowcolor{RoyalPurple!12} \textbf{\method{}} & \textbf{32.71}	& \textbf{0.14} &	\textbf{57.11}	& \textbf{10.04} \\
        \bottomrule
    \end{tabular}
    }
    \label{tab:traj}
\end{wraptable}

%% file: appendix/app-baseline.tex
\section{Baseline Details}
\label{app:baseline}

We include additional details of the coding and medical domain-specific LLMs:

% \subsection{Coding LLMs}
\begin{itemize}[leftmargin=0.6cm]
    \item \textbf{Qwen2.5-Coder-Instruct}~\citep{hui2024qwen2} is derived from the Qwen2.5 series and further fine-tuned explicitly on large-scale coding datasets and coding-specific instruction sets. This targeted training substantially enhances their capabilities in code generation, debugging, and programmatic reasoning, outperforming general-purpose models of similar scale on coding tasks.
    \item \textbf{medgemma-4b-it} (\texttt{gemma-3-4b-pt})~\citep{medgemma-hf} is a medical-domain variant based on gemma architecture and fine-tuned specifically on medical QA and instruction datasets, which provide strong capabilities for medical reasoning and question answering.
    \item \textbf{HuatuoGPT-o1-7B} (\texttt{Qwen2.5-7B-Instruct})~\citep{chen2024huatuogpt}, built on the Qwen2.5-7B architecture, is extensively fine-tuned in clinical reasoning datasets via PPO with verifier-based rewards to enhance complex reasoning capabilities. Specifically, it incorporates a medical-specific verifier model that guides the generation of complex reasoning trajectories. HuatuoGPT-o1-7B excels in medical reasoning tasks by explicitly generating intermediate reasoning steps that facilitate iterative refinement and introspective evaluation. 
    \item \textbf{m1-7B-23K} (\texttt{Qwen2.5-7B-Instruct})~\citep{huang2025m1} is fine-tuned on approximately 23,000 rigorously curated medical QA examples, significantly enhancing its domain-specific knowledge and reasoning capabilities.
    \item \textbf{MedReason-8B} (\texttt{Llama-3.1-8B-Instruct})~\citep{wu2025medreason} is fine-tuned for medical questions-answering and clinical reasoning tasks. Its training emphasizes the generation of step-by-step rationales, enabling robust performance on medical reasoning and diagnostic tasks.
    \item \textbf{Baichuan-M1-14B-Instruct}~\citep{wang2025baichuan} is a 14B medical LLM pre-trained from scratch on approximately 20 trillion tokens of medical domain-specific content and high-quality general text. It integrates specialized modeling across over 20 medical specialties with advanced architectural modifications enhancing context understanding and long-sequence reasoning. 
\end{itemize}

%% file: appendix/app-implementation.tex
\section{Implementation Details}
\label{app:implementation}

\noindent \textbf{Evaluation Metrics.} 
Following existing agent benchmarks~\citep{liu2023agentbench}, we adopt  \emph{success rate (SR)} as the primary evaluation metric. 
For \emph{database, data science, and bioinformatics} tasks with explicit ground truths, we compare LLM-generated code execution outputs with reference solutions using exact match. 
For open-ended \emph{ML} tasks in clinical decision support, we measure performance using \emph{accuracy (Acc)} across provided test cases. 
Note that these code generation tasks inherently have infinite solution spaces, unlike traditional classification problems with bounded solution spaces (\eg, even random guessing can yield around 50\% accuracy in binary classification).
The \emph{overall score} is computed by averaging performance across tasks in test sets of \method{} (leaderboard), providing a comprehensive evaluation of coding-based biomedical reasoning capabilities within \method{}.

\noindent \textbf{Experimental Setup Details.}
We limit interactions to a maximum of $15$ turns per session, providing agents full access to interaction histories and constraining runtime to $120$ seconds per session. 
Input tokens are capped at $32,768$, with output limited to $8,192$ tokens per round.
We use Python $3.10$ as the primary language for agent-code execution due to its modular design and suitability for biomedical computations. 
To enable interactive feedback (\cref{sec:method-env}), we employ a rule-based parser converting LLM outputs to JSON, facilitating seamless code execution, and utilize \texttt{gpt-4.1-mini} to translate execution errors into grounded explanations. 
We configure all baseline LLMs following established best practices for reproducibility. 
Specifically, instruction-following LLMs are configured with a temperature of zero, while reasoning models use a temperature of 0.6. 
For all experiments with \texttt{Qwen-3} series, we switch to thinking mode for optimal performance under complex reasoning scenarios (\eg, logic, math, and coding).

\noindent \textbf{SFT.}
For SFT experiments, smaller models (up to 8B parameters) are trained using eight NVIDIA A100 GPUs, whereas the 14B-parameter model is trained on eight NVIDIA H200 GPUs. We utilize the AdamW optimizer~\citep{loshchilov2017decoupled} with a learning rate of $1e-4$. The training batch size is set to 8, and the maximum input token length per batch is configured to 40,000 tokens.

\noindent \textbf{DPO.}
DPO experiments are conducted using the same hardware configurations as SFT experiments. We employ the AdamW optimizer with a reduced learning rate of $5e-6$. Training utilizes a batch size of 64 and a KL-divergence coefficient ($\beta$) of 0.1 to regulate the divergence from the initial policy.

\noindent \textbf{PPO \& GRPO.}
PPO and GRPO experiments are conducted using the same hardware configurations as SFT experiments.
All online RL experiments are conducted using VeRL framework~\citep{sheng2025hybridflow}.
We integrate the VeRL package and dependencies inside the \ours{} docker image to enable communication between the reward functions and the evaluation module.
PPO and GRPO training is performed with a batch size of 128 and a learning rate of $1\times 10^{-5}$.
The temperature parameter during model rollout is consistently set to 0.6. Throughout training, the coefficient for the KL divergence regularization term is fixed at $\beta=1\times 10^{-3}$. 
% All experiments are conducted with 8 NVIDIA H200 GPUs, each equipped with 141GB of memory.

%% file: appendix/app-exp.tex
\section{Additional Experimental Results}
\label{app:exp}

\subsection{Code Quality and Efficiency}
\label{app:codequlity}
\input{table/tab-appendix-metrics}

For a comprehensive evaluation, we further report additional evaluation metrics on code quality and efficiency, including (1) \textbf{number of turns} for interaction effectiveness, (2) cyclomatic \textbf{complexity} for code complexity, (3) \textbf{maintainability} index for code readability, and (4) \textbf{line-of-code (loc)} and (5) \textbf{logical line-of-code (lloc)} for code efficiency (Table~\ref{tab:metric}). 
Comparing different tasks (take gpt-4.1 for example), we observe that machine learning tasks such as EHRSHOT involve significantly higher complexity and longer code.
Comparing different models (averaged across datasets), we observe that advanced closed-source models generate more complex and longer code; after training, \ours produces structurally efficient and more maintainable code compared to backbone models.

\begin{wrapfigure}[10]{r}{0.4\textwidth}
    \centering
    \vspace{-2ex}
    \includegraphics[width=\linewidth]{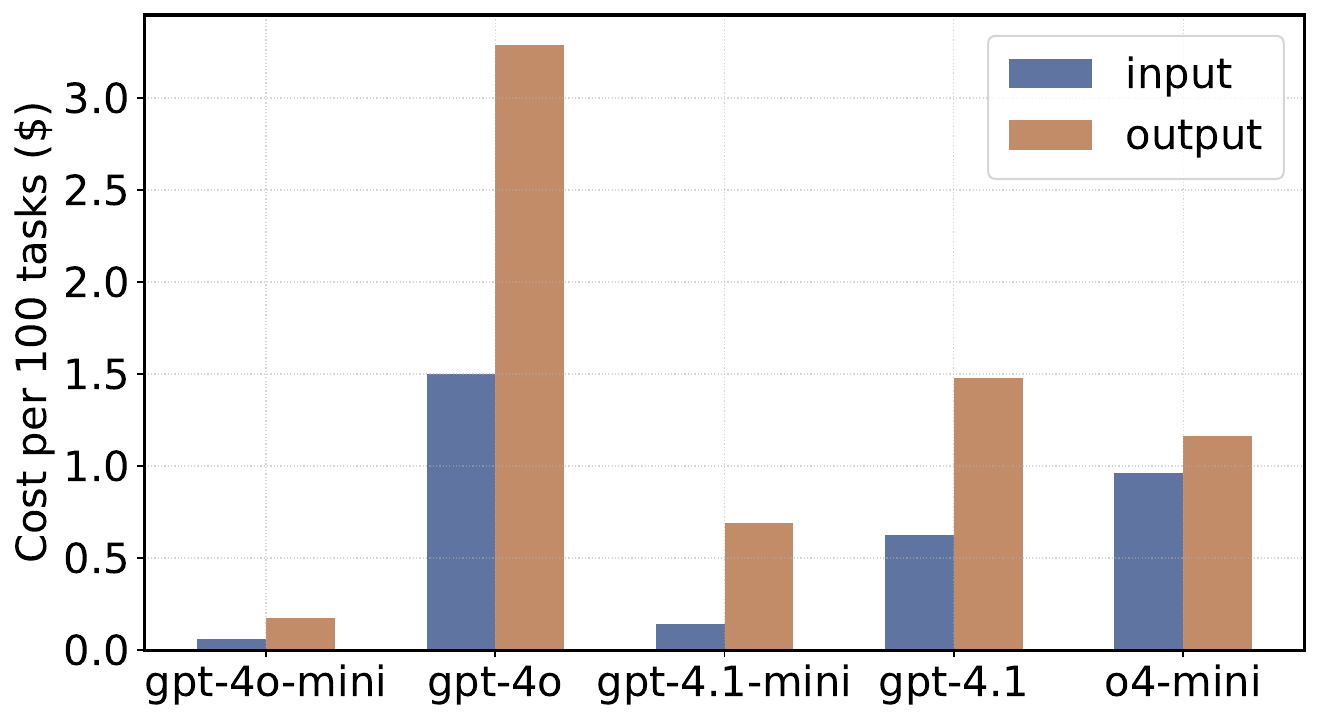}
    \vspace{-4ex}
    \caption{Cost information.}
    \label{fig:cost}
\end{wrapfigure}

\subsection{Cost Analysis}
\input{table/tab-cost}
Table~\ref{tab:cost} summarizes input and output token statistics for various API-based proprietary LLMs evaluated on datasets within \method{}. Notably, the input and output token lengths per query vary significantly across models and tasks. Among these models, \texttt{gpt-4.1-mini} achieves relatively low average input and moderate output token counts, which implies more efficient token utilization during inference compared to larger variants such as \texttt{gpt-4o} and \texttt{gpt-o4-mini}. Conversely, \texttt{gpt-o4-mini} incurs higher average input costs. 
Figure~\ref{fig:cost} presents the API cost per 100 tasks.
Overall, smaller GPT variants (\eg, \texttt{gpt-4.1-mini} and \texttt{gpt-4o-mini}) offer superior token-efficiency, translating into lower computational and API costs without substantial compromise in performance, demonstrating their effectiveness as cost-efficient solutions for large-scale biomedical reasoning applications.

\subsection{Structured and Open-ended Tasks}
\begin{figure}[h]
	\centering
    \subfigure[Structured Tasks]{
	\includegraphics[width=0.31\linewidth]{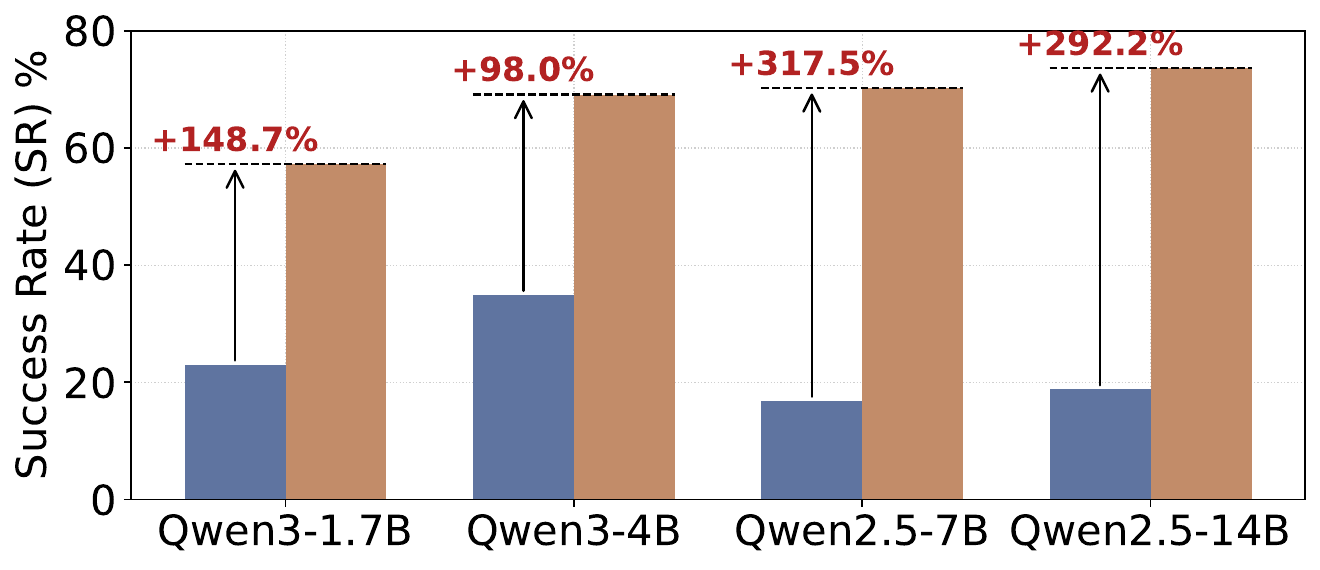}
	\label{fig:sft-tep}
	}
	\subfigure[Open-Ended Tasks]{
	\includegraphics[width=0.31\linewidth]{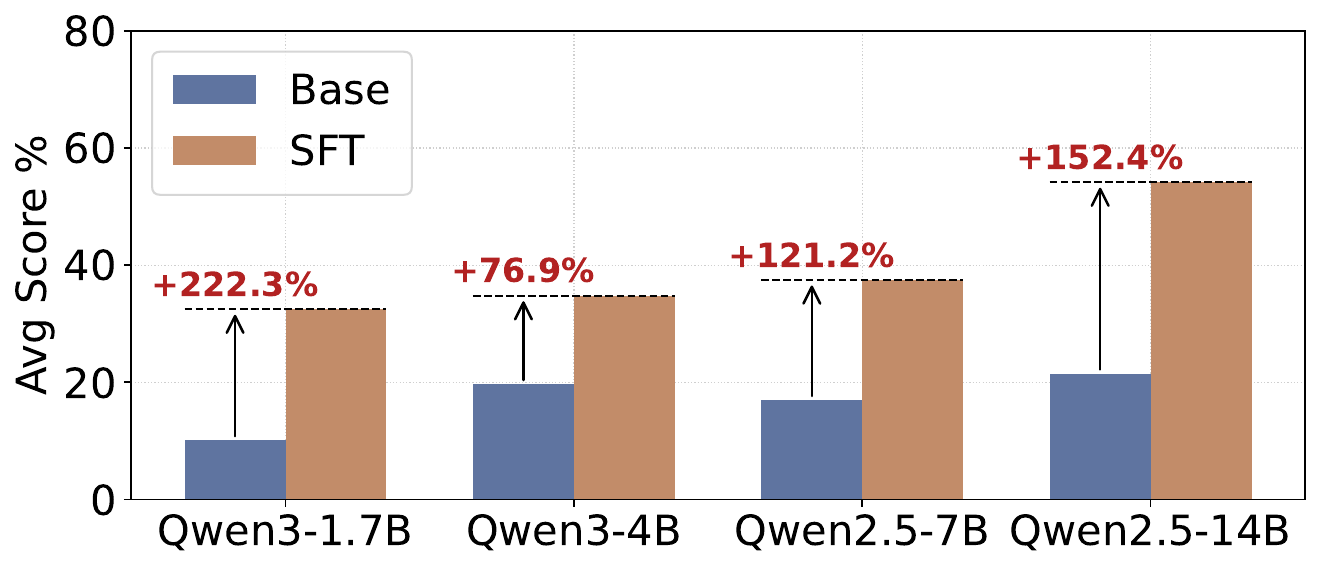}
	\label{fig:sft-open}
	} 
	\subfigure[Overall]{
	\includegraphics[width=0.31\linewidth]{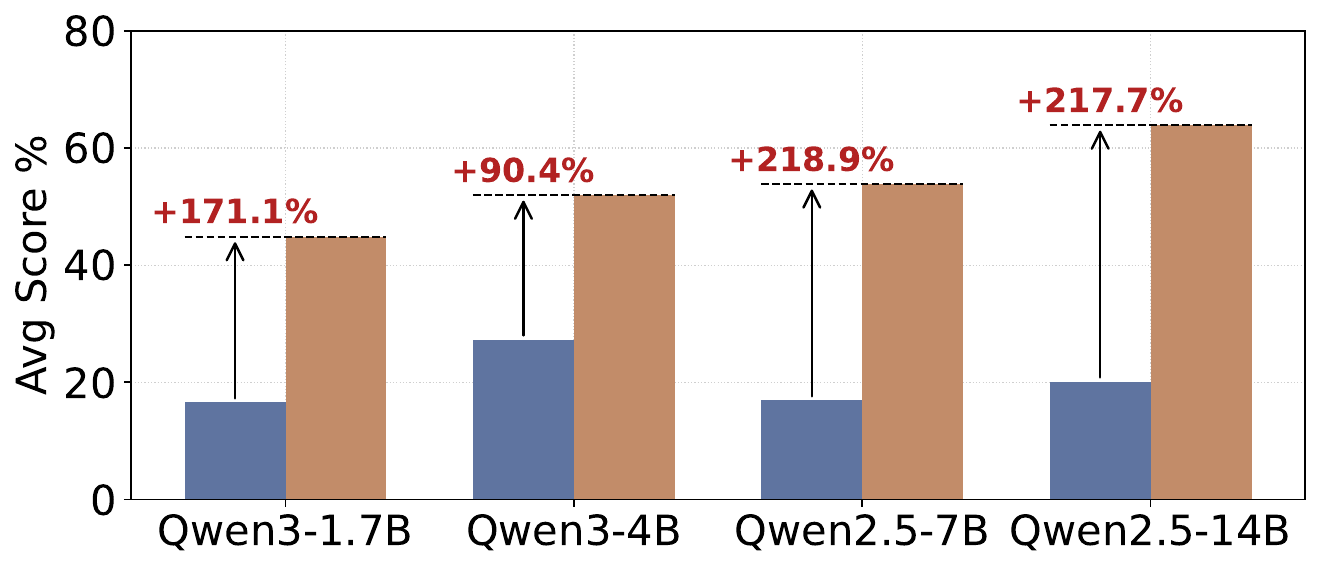}
	\label{fig:sft-avg}
	} 
	\caption{\ours{} SFT performance on \method{} across various backbone LLMs.}
\label{fig:sft}
\end{figure}

Figure~\ref{fig:sft} shows substantial performance gains from SFT across four OSS backbone LLMs of varying sizes. 
Simple SFT on successful trajectories markedly boosts performance on structured coding tasks, indicating its effectiveness in capturing structured coding patterns. 
DPO, in contrast, is particularly effective for optimizing performance on open-ended tasks. 

\begin{wrapfigure}[10]{r}{0.15\textwidth}
    \centering
    \vspace{-8ex}
    \includegraphics[width=\linewidth]{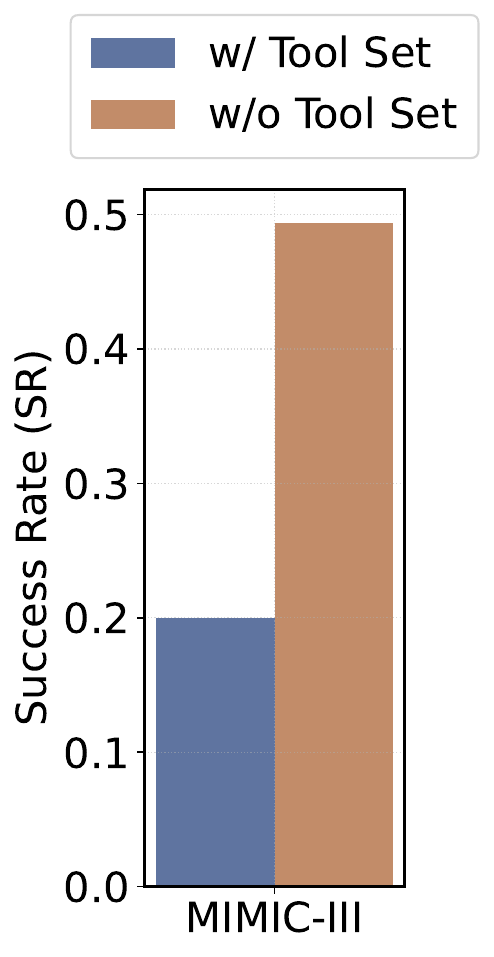}
    \vspace{-10ex}
    \caption{Effect of toolset.\vspace{-12ex}}
    \label{fig:tool}
\end{wrapfigure}

\subsection{Ablation Study: Effect of Pre-defined Toolset}
Figure~\ref{fig:tool} compares the performance of GPT-4-based agents on the MIMIC-III dataset with and without predefined toolsets integrated into our agent scaffold. 
This illustrates our agent scaffold's ability to flexibly accommodate external tools. Interestingly, despite providing a set of predefined tools, including functions for database loading, data filtering, value retrieval, arithmetic calculations, date computations, and SQL execution (see additional details of toolset in \citet{shi2024ehragent}), we observe a surprising decline in agent performance. It suggests that the LLM agent inherently generates more flexible and contextually appropriate code when unencumbered by predefined function constraints, aligning with the observations reported by \citep{qian2025smart,qiu2025alitageneralistagentenabling}.

\subsection{Ablation Study: Effect of Warm-up Stage }
\input{table/tab-sft-abs}

Table~\ref{tab:sft-abs} shows the effect of the initial SFT stage during agentic RL finetuning. Although DPO alone slightly underperforms compared to SFT, combining an initial SFT warm-up with subsequent DPO further improves overall results by leveraging their complementary strengths.

\subsection{Case Study}

\begin{figure}[h]
    \centering
    \includegraphics[width=0.8\linewidth]{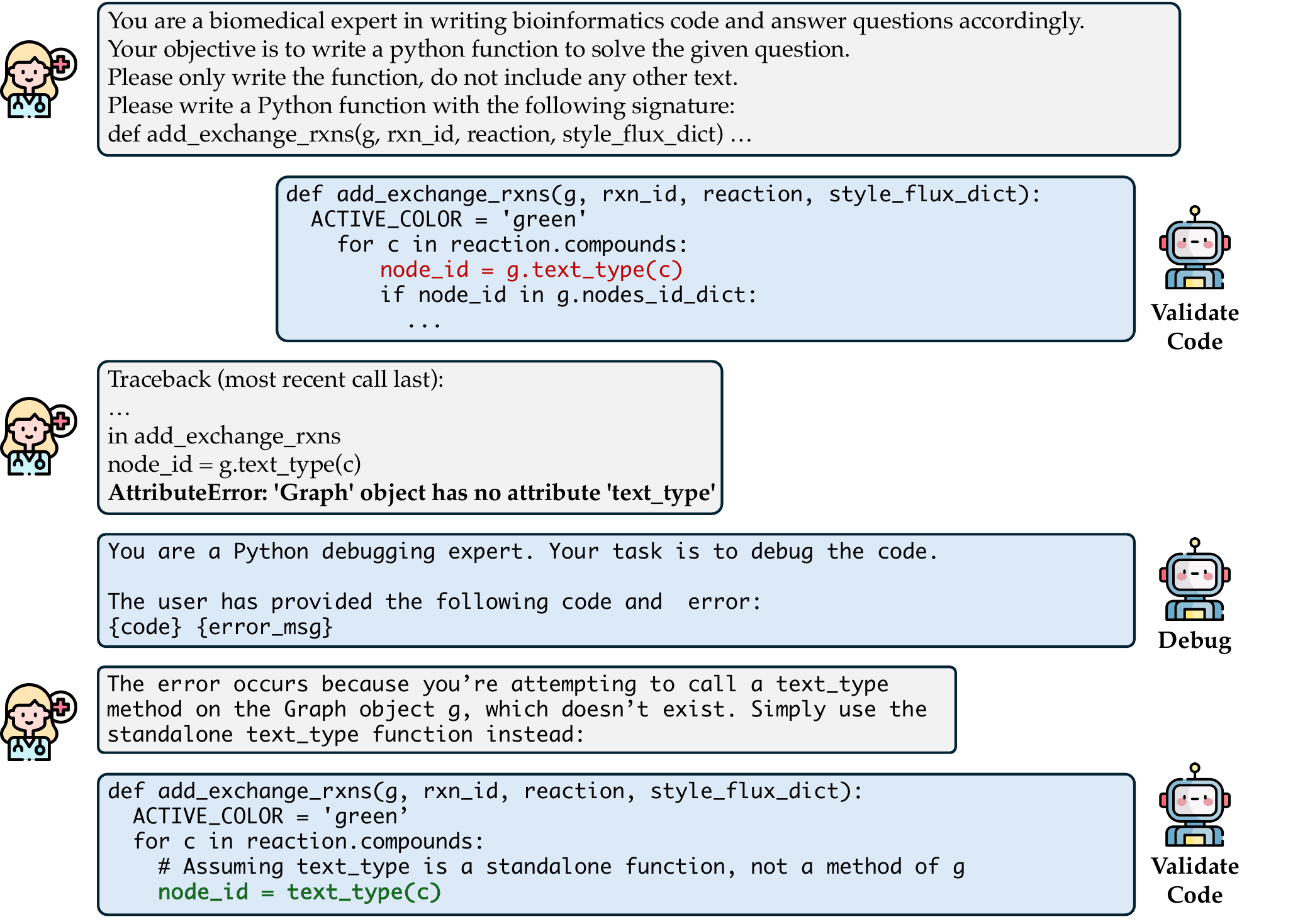}
    \caption{Case study of \texttt{gpt-4.1-mini} on BioCoder.}
    \label{fig:case}
\end{figure}

To illustrate the practical utility of interactive coding mechanism, we conduct a detailed case study involving a typical bioinformatics coding task in Figure~\ref{fig:case}. Specifically, the task requires writing a Python function (\texttt{add\_exchange\_rxns}) that modifies biochemical reaction graphs by integrating exchange reactions. Initially, the LLM agent-generated solution encountered an attribute error, mistakenly invoking a non-existent \texttt{text\_type} method on a \texttt{Graph} object. Upon receiving explicit debugging feedback, the LLM agent effectively identified and corrected the mistake by utilizing the standalone \texttt{text\_type} function rather than incorrectly calling it as a method of the graph instance. This case highlights the capability of debugging in \method{} environment to provide targeted, actionable debugging feedback, enabling iterative code refinement and significantly enhancing agent-generated solutions for complex biomedical programming tasks.

\begin{figure}[h]
    \centering
    \includegraphics[width=0.99\linewidth]{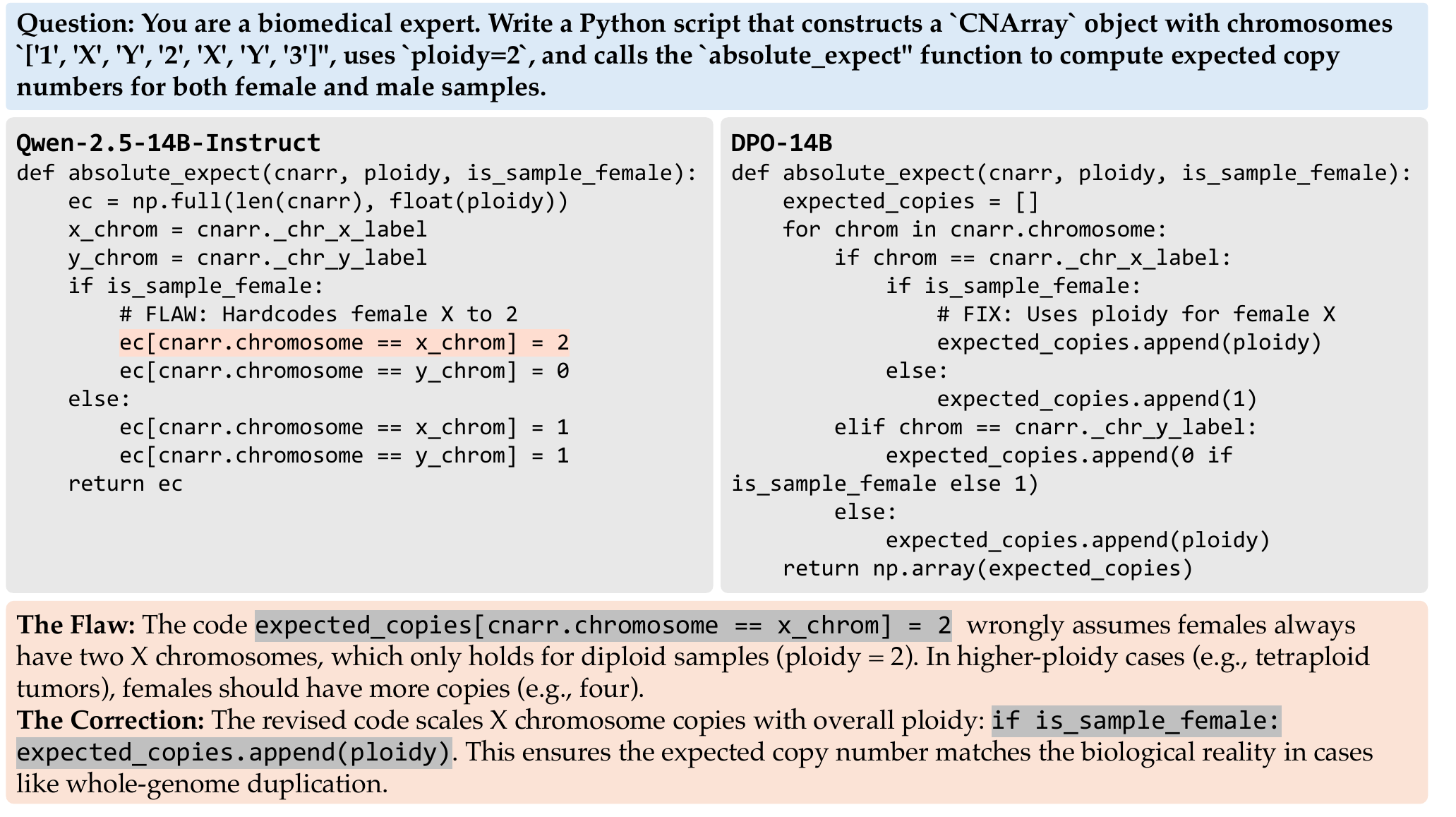}
    \caption{
    Domain-specific code generation error in a biomedical task from BioCoder~\citep{tang2024biocoder}. The task requires implementing a Python function to compute chromosome copy numbers based on ploidy. The baseline model (Qwen-2.5-14B-Instruct, left) incorrectly hardcodes the female X chromosome count to 2, failing to account for non-diploid scenarios such as tetraploid tumor cells. Our DPO-trained model (DPO-14B, right) correctly implements dynamic scaling of X chromosome copy numbers proportional to the ploidy parameter, demonstrating improved understanding of domain-specific biological constraints.
    % \ycz{check} An example of a subtle, domain-specific flaw in code generation for a biomedical task from BioCoder~\cite{tang2024biocoder}. The prompt (top) asks for a Python function to compute chromosome copy numbers based on ploidy. The baseline model, Qwen-2.5-14B-Instruct (left), incorrectly hardcodes the female X chromosome copy number to 2. This logic fails in non-diploid cases (e.g., tetraploid tumors). In contrast, our model, DPO-14B (right), correctly generalizes the function, dynamically scaling the expected X chromosome count with the provided ploidy parameter.
    }
    \label{fig:case1}
\end{figure}

\begin{figure}[h]
    \centering
    \includegraphics[width=0.99\linewidth]{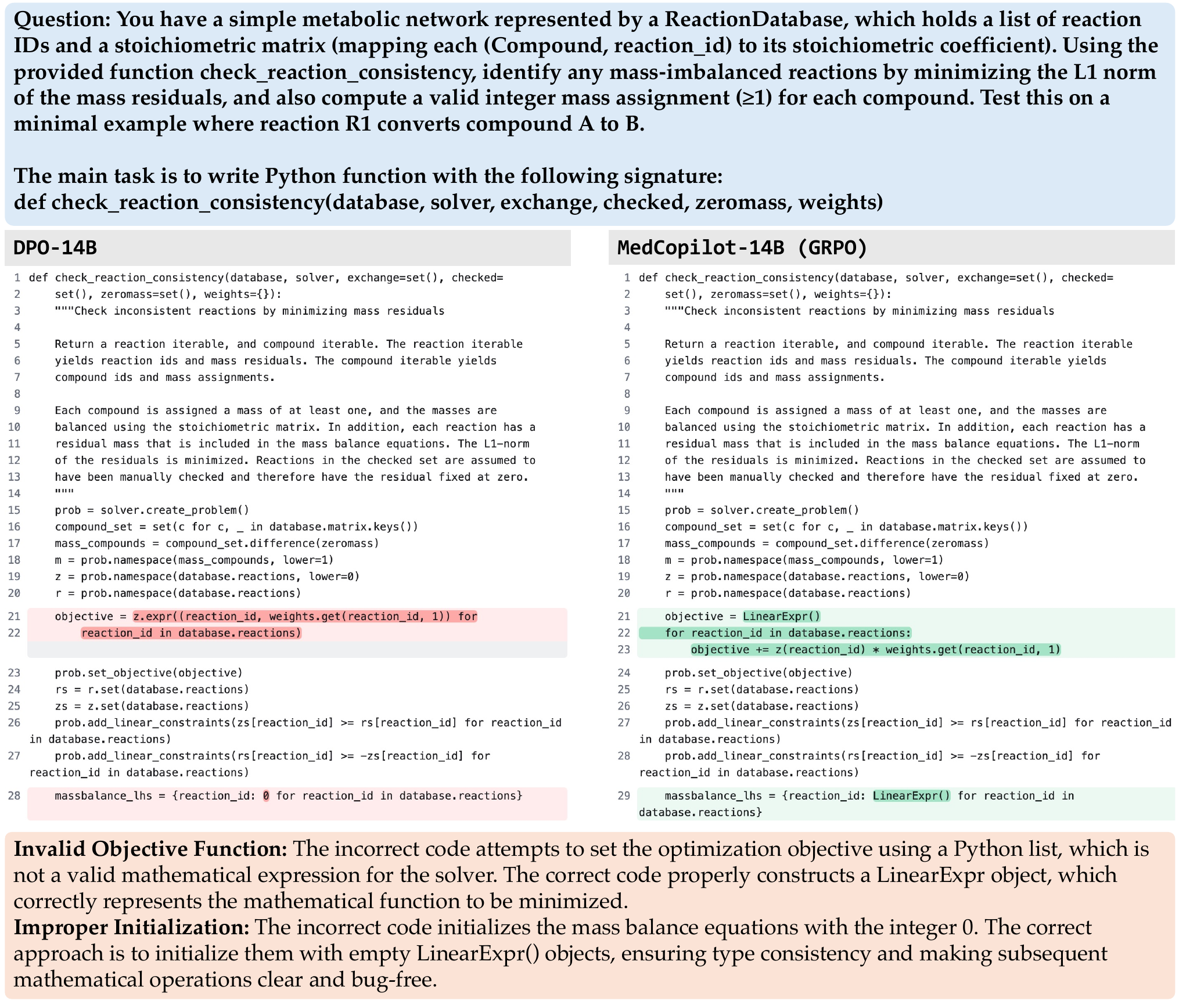}
    \caption{
    Qualitative comparison of code generation for a complex optimization task from BioDSBench~\citep{wang2024can}. The task requires implementing a linear program to verify mass conservation in metabolic networks. The baseline model (DPO-14B, left) generates syntactically plausible but semantically incorrect code with two critical errors: (1) defining the optimization objective using a Python list rather than the required LinearExpr object, and (2) initializing mass balance equations with integer 0 instead of LinearExpr(). In contrast, \ours{}-14B (GRPO, right) correctly employs the LinearExpr class for both objective function construction and mass balance initialization, producing executable code that accurately models the metabolic constraints.
    % \ycz{check} Qualitative comparison of code generation for a complex optimization task from BioDSBench~\cite{wang2024can} The task requires setting up a linear program to check mass consistency in a metabolic network. (Left) The baseline model (DPO-14B) produces syntactically plausible but semantically invalid code. It commits two critical errors: (1) it attempts to define the optimization objective using a Python list instead of a valid LinearExpr object, and (2) it improperly initializes mass balance equations with the integer 0 instead of an empty LinearExpr(). (Right) Our model (\ours{}-14B (GRPO)) correctly identifies the need for the LinearExpr class, using it for both the iterative construction of the objective function and the initialization of the mass balance dictionary, resulting in correct, executable code.
    }
    \label{fig:case2}
\end{figure}

\begin{figure}[h]
    \centering
    \includegraphics[width=0.99\linewidth]{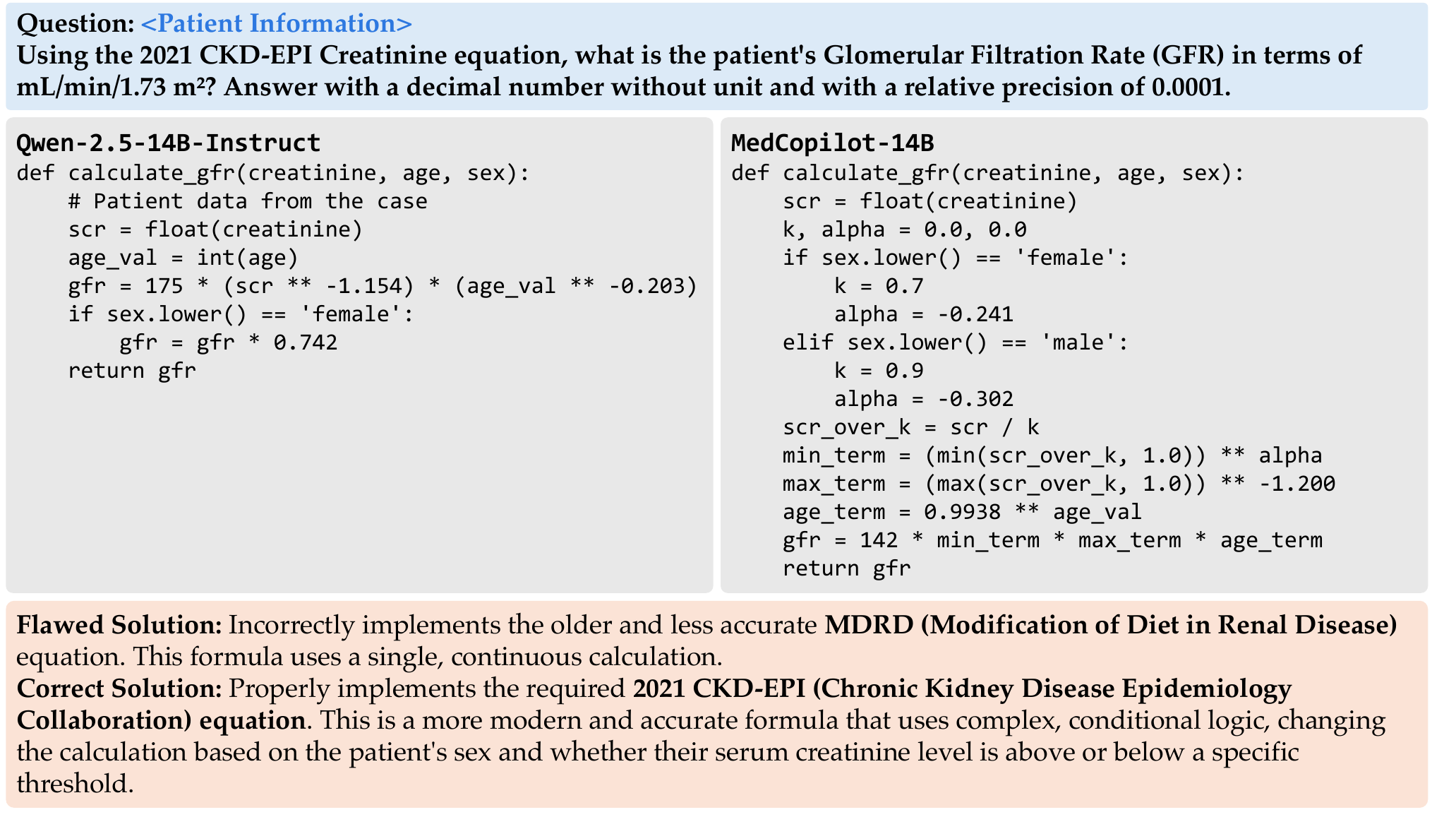}
    \caption{Domain-specific complexity in medical code generation from MedCalcBench~\citep{khandekar2024medcalc}. The task requires implementing the 2021 CKD-EPI equation for Glomerular Filtration Rate (GFR) calculation. The baseline model (Qwen-2.5-14B, left) incorrectly generates a flawed implementation of the outdated MDRD formula instead of the requested 2021 standard. In contrast, \ours{}-14B (right) accurately implements the complex conditional logic specified in the 2021 CKD-EPI guidelines, demonstrating precise adherence to current medical standards.
    % \ycz{check} An illustrative example of domain-specific complexity in code generation from MedCalcBench~\cite{khandekar2024medcalc} When prompted to implement the modern 2021 CKD-EPI equation for Glomerular Filtration Rate (GFR), the generalist model (Qwen-2.5-14B, left) incorrectly provides a flawed implementation of the simpler, outdated MDRD formula. In contrast, our model (\ours{}-14B, right) correctly generates the complex, conditional logic required by the specified 2021 standard, demonstrating a superior understanding of complex medical guidelines.
    }
    \label{fig:case3}
\end{figure}

Case studies with code patterns in Figures~\ref{fig:case1} to \ref{fig:case3} further illustrate how baseline models frequently produce syntactically valid code but incorrect solution in biomedical tasks, from hardcoding biological parameters to misapplying medical formulas, while our fine-tuned models demonstrate accurate implementation of domain-specific constraints and current clinical standards.

\subsection{Difficulty Analysis on External Evaluation Set}
\input{table/tab-appendix-ext}

\method{} includes four challenging unseen out-of-distribution medical coding tasks as external validation sets in \cref{sec:exp-agent}. 
For example, the original MIMIC-Extract task in \method{} intentionally utilizes \textit{raw, unprocessed} data as a \textit{challenging, out-of-distribution} scenario designed specifically to assess model capabilities in feature engineering and data preprocessing.
To illustrate the difference clearly, we additionally evaluate \textit{a fully pre-processed} version of MIMIC-Extract. As demonstrated in Table~\ref{tab:mimic_bixbench}, providing structured data significantly improves model performance, highlighting the distinct difficulty posed by raw data.

To further demonstrate the generalization of \ours, we include an additional evaluation set, BixBench~\citep{mitchener2025bixbench}, a bioinformatics coding dataset comprising over 50 real-world scenarios of practical biological data analysis with nearly 300 associated open-answer questions.
It is designed to measure the ability of LLM-based agents to explore biological datasets, perform long, multi-step analytical trajectories, and interpret the nuanced results of those analyses. 
Exceptional performance in BixBench demonstrates the robustness of \ours and its ability to generalize beyond the specific domain of medical coding to broader scientific analytical tasks. 

\subsection{Human Study}
\input{table/tab-appendix-human-study}
To systematically compare coding styles and performance differences between human programmers and automated agents, we conducted a human evaluation involving 83 tasks randomly selected from the test subsets of the 12 datasets included in \method{}. 
This evaluation set comprises 52 structured and 31 open-ended biomedical coding tasks. The human participants are biomedical engineers and research scientists with over six years of experience in computational biology, relational database querying, HTTP-based interactions, and machine learning development. 
The human evaluation study was conducted under the approval of the Institutional Review Board (IRB). Participants voluntarily contributed to the evaluation and did not receive monetary compensation. 

Table~\ref{tab:human_performance} summarizes the results of human evaluation study conducted to establish reference performance benchmarks across representative structured and open-ended biomedical reasoning tasks from the \method{} benchmark. 
Human experts completed selected instances from each dataset, documenting the number of attempts, correctly solved instances, overall SR, total time spent, and average time per task (in minutes). 
Results indicate that, on average, the human subject required approximately 4.5 times longer to solve open-ended tasks relative to structured tasks, while achieving a 40\% lower success rate, reflecting the increased complexity and cognitive load associated with open-ended biomedical reasoning scenarios.

In addition, we also performed a quantitative analysis on 250+ trajectories (randomly sampled over 10\% of our trajectory collection in \cref{sec:sft}) and confirmed that the vast majority of successful solutions followed a logically sound path, with cases of `correct answer from flawed code' being exceptionally rare (<1\%).

\

%% file: table/tab-appendix-metrics.tex
\begin{table}[ht]
\centering
\renewcommand\arraystretch{0.98}
\caption{Additional evaluation on code quality and efficiency.
}
\resizebox{1.01\linewidth}{!}{ %
\begin{tabular}{l | c c c | c c | c c | c | c}
\toprule
\textbf{Datasets ($\rightarrow$)} &  {\textbf{MIMIC.}} & {\textbf{eICU}} & {\textbf{TREQS}} & {\textbf{MedCalc.}} & \textbf{MedAgent.} & 
 \textbf{BioCoder} & \textbf{BioDS.} & \textbf{EHRSHOT} & \textbf{Avg.}  \\ \midrule
\multicolumn{10}{l}{\textit{gpt-4.1 (2025-04-14)}} \\ \midrule
 \#turns & 25.91 & 26.59 & 20.65 & 10.73 & 17.28 & 22.08 & 21.75 & 8.71 & 19.21 \\ 
 complexity & 0.01 & 0.06 & 0.01 & 4.09 & 0.23 & 7.77 & 0.17 & 20.97 & 4.16 \\
 maintainability & 95.14 & 95.99 & 96.62 & 88.38 & 91.04 & 68.20 & 92.67 & 56.24 & 85.54 \\
 loc & 9.26 & 9.67 & 4.17 & 19.00 & 18.89 & 24.82 & 28.97 & 144.69 & 32.43 \\
 lloc & 5.86 & 6.33 & 3.00 & 15.20 & 10.79 & 21.84 & 16.44 & 110.51 & 23.75 \\ \midrule
\multicolumn{10}{l}{\textit{gpt-4.1-mini (2025-04-14)}} \\ \midrule
 \#turns & 19.66 & 19.90 & 16.35 & 9.18 & 19.20 & 23.08 & 16.53 & 22.60 & 18.31 \\ 
 complexity & 0.02 & 0.04 & 0.01 & 3.51 & 0.03 & 7.30 & 0.26 & 19.85 & 3.88 \\
 maintainability & 95.62 & 96.06 & 98.93 & 87.01 & 94.43 & 69.43 & 92.54 & 57.77 & 86.47 \\
 loc & 16.49 & 14.47 & 6.85 & 23.37 & 13.08 & 25.98 & 28.17 & 171.69 & 37.51 \\
 lloc & 8.05 & 7.22 & 3.68 & 17.58 & 7.92 & 20.78 & 15.40 & 119.58 & 25.03 \\ \midrule
\multicolumn{10}{l}{\textit{Qwen2.5-7B-Instruct}} \\ \midrule
 \#turns & 17.23 & 14.81 & 12.38 & 5.98 & 14.39 & 25.42 & 9.31 & 15.33 & 14.36 \\ 
 complexity & 0.02 & 0.02 & 0.01 & 4.41 & 0.01 & 4.78 & 0.30 & 11.09 & 2.58 \\
 maintainability & 96.54 & 96.02 & 98.58 & 82.65 & 80.20 & 81.67 & 95.66 & 54.69 & 85.75 \\
 loc & 16.81 & 17.07 & 8.72 & 28.54 & 49.09 & 20.81 & 22.00 & 137.85 & 37.61 \\
 lloc & 7.52 & 8.23 & 4.38 & 18.09 & 25.34 & 15.46 & 11.79 & 90.58 & 22.67 \\ \midrule
\multicolumn{10}{l}{\textit{\ours (7B)}} \\ \midrule
 \#turns & 20.74 & 17.80 & 14.31 & 7.86 & 16.24 & 28.97 & 16.80 & 29.73 & 19.06 \\ 
 complexity & 0.01 & 0.01 & 0.01 & 3.81 & 0.01 & 5.08 & 0.04 & 18.66 & 3.45 \\
 maintainability & 94.58 & 95.01 & 98.49 & 83.76 & 82.64 & 81.40 & 97.68 & 62.47 & 87.00 \\
 loc & 21.58 & 19.88 & 12.00 & 25.42 & 53.67 & 24.76 & 17.16 & 141.50 & 39.50 \\
 lloc & 9.95 & 9.10 & 5.73 & 17.74 & 26.26 & 17.82 & 9.11 & 95.97 & 23.96 \\
\bottomrule
\end{tabular}%
}
\label{tab:metric}
\end{table}

%% file: table/tab-cost.tex
\begin{table}[ht]
\centering
\renewcommand\arraystretch{0.98}
\caption{Statistics of input and output tokens per question for API-based commercial LLMs.
}
\resizebox{1.01\linewidth}{!}{ %
\begin{tabular}{l | c c c | c c | c c | c | c}
\toprule
\textbf{Datasets ($\rightarrow$)} &  {\textbf{MIMIC.}} & {\textbf{eICU}} & {\textbf{TREQS}} & {\textbf{MedCalc.}} & \textbf{MedAgent.} & 
 \textbf{BioCoder} & \textbf{BioDS.} & \textbf{EHRSHOT} & \textbf{Avg.}  \\ \midrule
\multicolumn{10}{l}{\textit{Input}} \\ \midrule
\texttt{gpt-4o-mini}~\citep{hurst2024gpt} & 3430.83& 1947.72 &1689.71 &651.92 &9501.86 &5166.50&5068.88&5986.20&4180.45 \\
\texttt{gpt-4o}~\citep{hurst2024gpt} & 4399.87&3122.02&1823.31&739.48&8474.81&5133.71&21077.12&3235.71&6000.75 \\
\texttt{gpt-4.1-mini}~\citep{gpt-4-1} & 1869.37&1691.45&1430.15&834.73&8087.50&2621.79&7369.35&4466.07&3546.30 \\
\texttt{gpt-4.1}~\citep{gpt-4-1} & 3730.90&2979.57&1754.18&759.64&7912.81&2728.24&3035.45&2092.14&3124.12 \\
\texttt{gpt-o4-mini}~\citep{o4-mini} &  2005.11&1688.73&1534.84&1306.49&7586.32&2193.82&50768.08&2858.79&8742.77 \\
\midrule
\multicolumn{10}{l}{\textit{Output}} \\
\midrule
\texttt{gpt-4o-mini}~\citep{hurst2024gpt} & 1206.00&714.72&918.45&379.28&4206.73&4170.56&1479.87&10484.53&2945.02 \\
\texttt{gpt-4o}~\citep{hurst2024gpt} & 840.16&852.41&696.61&537.09&2821.00&4144.91&7278.49&9127.14&3287.23 \\
\texttt{gpt-4.1-mini}~\citep{gpt-4-1} & 952.68&991.78&880.43&1000.06&2892.98&3328.07&1308.73&23276.67&4328.93 \\
\texttt{gpt-4.1}~\citep{gpt-4-1} & 771.91&781.86&753.88&787.45&2051.20&2846.58&1627.78&5163.57&1848.03 \\
\texttt{gpt-o4-mini}~\citep{o4-mini} &  1586.65&1392.11&893.76&2407.87&1718.22&3144.74&1952.88&8083.71&2647.49 \\
\bottomrule
\end{tabular}%
}
\label{tab:cost}
\end{table}

%% file: table/tab-sft-abs.tex
\begin{table}[ht]
\centering
\renewcommand\arraystretch{0.92}
\caption{Effect of SFT stage in two-stage finetuning framework.
}
\resizebox{1.01\linewidth}{!}{ %
\begin{tabular}{l | c c c| c c| c c| c | c c }
\toprule
% & \multicolumn{9}{c|}{{In-distribution}} & \multicolumn{6}{c}{{Out-of-distribution}} \\
% \cmidrule(lr){2-10} \cmidrule(lr){11-16} 
{Datasets ($\rightarrow$)} &  {{MIMIC-III}} & {{eICU}} & {{TREQS}} & {{MedCalc.}} & {MedAgent.} & 
 {BioCoder} & {BioDS.} & {EHRSHOT} & {Avg.} & {{$\Delta$}} \\
{Base ($\downarrow$) / Metrics ($\rightarrow$)} & SR & SR & SR & SR & SR & SR  & SR & Acc & \multicolumn{2}{c}{Score} \\ \midrule
% \texttt{Qwen3-1.7B}  \\
% \quad +SFT  & & & & & & & & & & \\
% \texttt{Qwen3-4B}  \\
% \quad +SFT  & & & & & & & & & & \\
% \texttt{Qwen3-8B}  \\
% \quad +SFT  & & & & & & & & & & \\ 
\rowcolor{Gray!16} \texttt{Qwen2.5-7B-Instruct} & 13.08 & 15.57 &	12.76	& 25.91	& 30.36	& 21.79 & 10.20 & 5.42 & 16.89 & --\\
% \quad +SFT  & 57.83 & 61.48	& {72.66}	& 89.06	& 50.85	& 28.33	& 55.10	& 15.62 & 53.87 & {(\emph{+36.98})}\\
\quad +DPO w/o SFT & 49.59 & 43.61	& 46.68	& 49.20	& 45.25	& 30.13	& 69.39	& 26.43	& 45.04 & {(\emph{+28.15})}\\
\quad +DPO  & {64.13} & {66.91} & 72.02 & {90.06} & {52.54} & {34.62} & {69.39} & {29.55}	& {59.90} & {(\emph{+43.02})}\\
% +iDPO \\
\midrule
% \hdashline
% \quad +PPO  & 66.10	& 67.25	& {73.88} & 74.52	& 51.33	& 32.71	& 65.47	& 32.40 & 57.96 & {(\emph{+41.07})}\\
% \rowcolor{Peach!16} \quad +GRPO  & {68.21} & {68.73} & {70.50}	& {92.33}	& {55.87}	& {37.40}	& {71.11} & {33.18 }&{ 62.17} & {(\emph{+45.28})}\\
% \midrule
\rowcolor{Gray!16} \texttt{Qwen2.5-14B-Instrust} & 17.21 & 14.07	& 16.43	& 27.40	& 35.59	& 29.49	& 16.33	& 4.45	& 20.12 & --\\
% \quad +SFT  & 61.45	& 62.46	& {76.38}	& {94.36}	& 52.54	& 39.80	& 89.80	& 34.58 & 63.92 & {(\emph{+43.80})} \\
\quad +DPO w/o SFT  & 57.49	 & 59.18 & 70.45 & 71.32 & 47.46 & 42.95 & 91.84 & 41.33 &	60.25 & {(\emph{+40.13})} \\
\quad +DPO  & {64.54} & {63.52} & 76.08 & 92.45 & {54.32} & {43.56} & {92.96}	& {43.56} &	{66.37} & {(\emph{+46.25})}\\
% \rowcolor{JungleGreen!16} {(Qwen3-14B)}  \\
% +iDPO \\
% \midrule
% \hdashline
% \quad +PPO  &  {(\emph{+xxx})}\\
% \rowcolor{SeaGreen!16} \quad +GRPO &  {(\emph{+xxx})}\\
\bottomrule
\end{tabular}%
}
\label{tab:sft-abs}
\end{table}

%% file: table/tab-appendix-ext.tex
\begin{table}[ht]
\centering
\renewcommand\arraystretch{0.98}
\caption{Difficulty analysis on external sets for model generalization.}
\resizebox{0.8\linewidth}{!}{%
\begin{tabular}{l | c c | c}
\toprule
\textbf{Model} & \textbf{MIMIC-Extract (raw)} & \textbf{MIMIC-Extract (processed)} & \textbf{BixBench} \\ 
\midrule
gpt-4.1-mini & 5.62 & 23.47 & 26.01 \\
gpt-4.1 & 10.41 & 28.94 & 32.09 \\
Qwen-2.5-7B-Instruct & 1.34 & 17.06 & 18.92 \\
\rowcolor{Peach!16} \ours (7B) & 2.14 & 25.88 & 28.72 \\
Qwen-2.5-14B-Instruct & 4.51 & 18.52 & 20.61 \\
\rowcolor{SeaGreen!16} \ours (14B) & 2.75 & 28.66 & 29.39 \\
\bottomrule
\end{tabular}%
}
\label{tab:mimic_bixbench}
\end{table}

%% file: table/tab-appendix-human-study.tex
\begin{table}[ht]
    \centering
    \caption{
Human evaluation on structured and open-ended tasks from \method{}. }
    \label{tab:human_performance}
    \begin{threeparttable}
    \renewcommand\arraystretch{0.98}
\resizebox{\linewidth}{!}{ %
    \begin{tabular}{l|ccccc}
    \toprule
    \textbf{Dataset ($\downarrow$)} & \textbf{\# Attempt} & \textbf{\# Correct} & \textbf{SR} & \textbf{Total Time (min)} & \textbf{Avg Time (min)}\\
    \midrule
    \multicolumn{6}{l}{\textit{Structured}} \\
    \midrule
    MIMIC-III~\citep{johnson2016mimic,lee2022ehrsql}       & 10 & 8  & 80\%   & 74  & 7.40        \\
    eICU~\citep{pollard2018eicu,lee2022ehrsql}           & 8  & 5  & 63\%   & 63  & 7.88      \\
    TREQS~\citep{wang2020text}          & 10 & 7  & 70\%   & 39  & 3.90        \\
    EHR-SeqSQL~\citep{ryu2024ehr}     & 10 & 8  & 80\%   & 67  & 6.70        \\
    MedCalcBench~\citep{khandekar2024medcalc}        & 7  & 5  & 71\%   & 57  & 8.14  \\
    N-PowerAI~\citep{ruan2025n}      & 7  & 6  & 86\%   & 96  & 13.7  \\
    \rowcolor{RoyalPurple!6}
    \textbf{Structured Task (Total)} & \textbf{52} & \textbf{39} & \textbf{75\%} & \textbf{396} & \textbf{7.62} \\
    \midrule
    \multicolumn{6}{l}{\textit{Open-ended}} \\
    \midrule
    MedAgentBench~\citep{jiang2025medagentbench}       & 6  & 6  & 100\%       & 89  & 14.833  \\
    EHRCon~\citep{kwon2024ehrcon}         & 6  & 1  & 17\% & 241 & 40.17  \\
    BioDSBench~\citep{wang2024can}          & 3  & 0  & 0\%         & 195 & 65.00          \\
    BioCoder~\citep{tang2024biocoder}       & 8  & 2  & 25\%      & 142 & 17.75       \\
    EHRSHOT~\citep{wornow2023ehrshot}        & 5  & -- & 89\% & 185 & 37.00          \\
    MIMIC-Extract~\citep{MIMICExtract}  & 3  & -- & 94\%  & 215 & 71.67       \\
    \rowcolor{RoyalPurple!6}
    \textbf{Open-ended Task (Total)} & \textbf{31} & \textbf{--} & \textbf{45\%} & \textbf{1067} & \textbf{34.419} \\
    \bottomrule
    \end{tabular}
    }
\end{threeparttable}
\end{table}

%% file: appendix/app-prompt.tex
\newpage
\section{Prompt Details}
\label{app:prompt}
\subsection{MIMIC-III Prompts}
We include prompt details for MIMIC-III tasks as follows:
\begin{tcolorbox}[
    colback=RoyalPurple!6,          % background color
    colframe=RoyalPurple!48,        % border color
    title=\bfseries MIMIC-III Prompt - Main,
    fonttitle=\small\sffamily,
    sharp corners,            % remove rounding if you like
    boxrule=0.8pt,
    top=1mm, bottom=1mm, left=1mm, right=1mm
]
\begin{lstlisting}[style=prompt]
You are a biomedical expert in handling EHR data and answer questions. 
Your objective is to solve a coding problem with given EHR data, with the goal of finally give a concrete answer to the question.
Assume you have knowledge of several tables:
(1) Tables are linked by identifiers which usually have the suffix 'ID'. For example, SUBJECT_ID refers to a unique patient, HADM_ID refers to a unique admission to the hospital, and ICUSTAY_ID refers to a unique admission to an intensive care unit.
(2) Charted events such as notes, laboratory tests, and fluid balance are stored in a series of 'events' tables. For example the outputevents table contains all measurements related to output for a given patient, while the labevents table contains laboratory test
(3) Tables prefixed with 'd_' are dictionary tables and provide definitions for identifiers. For example, every row of chartevents is associated with a single ITEMID which represents the concept measured, but it does not contain the actual name of the measurement. By joining chartevents and d_items on ITEMID, it is possible to identify the concept represented by a given ITEMID.
(4) For the databases, four of them are used to define and track patient stays: admissions, patients, icustays, and transfers. Another four tables are dictionaries for cross-referencing codes against their respective definitions: d_icd_diagnoses, d_icd_procedures, d_items, and d_labitems. 
\end{lstlisting}
\end{tcolorbox}

\begin{tcolorbox}[
    colback=RoyalPurple!6,          % background color
    colframe=RoyalPurple!48,        % border color
    title=\bfseries MIMIC-III Prompt - Table information,
    fonttitle=\small\sffamily,
    sharp corners,            % remove rounding if you like
    boxrule=0.8pt,
    top=1mm, bottom=1mm, left=1mm, right=1mm
]
\begin{lstlisting}[style=prompt]
For different tables, they contain the following information:
(1) ADMISSIONS.csv: ROW_ID, SUBJECT_ID, HADM_ID, ADMITTIME, DISCHTIME, ADMISSION_TYPE, ADMISSION_LOCATION, DISCHARGE_LOCATION, INSURANCE, LANGUAGE, MARITAL_STATUS, ETHNICITY, AGE
(2) CHARTEVENTS.csv: ROW_ID, SUBJECT_ID, HADM_ID, ICUSTAY_ID, ITEMID, CHARTTIME, VALUENUM, VALUEUOM
(3) COST.csv: ROW_ID, SUBJECT_ID, HADM_ID, EVENT_TYPE, EVENT_ID, CHARGETIME, COST
(4) D_ICD_DIAGNOSES.csv: ROW_ID, ICD9_CODE, SHORT_TITLE, LONG_TITLE
(5) D_ICD_PROCEDURES.csv: ROW_ID, ICD9_CODE, SHORT_TITLE, LONG_TITLE
(6) D_ITEMS.csv: ROW_ID, ITEMID, LABEL, LINKSTO
(7) D_LABITEMS.csv: ROW_ID, ITEMID, LABEL
(8) DIAGNOSES_ICD.csv: ROW_ID, SUBJECT_ID, HADM_ID, ICD9_CODE
(9) ICUSTAYS.csv: ROW_ID, SUBJECT_ID, HADM_ID, ICUSTAY_ID, FIRST_CAREUNIT, LAST_CAREUNIT, FIRST_WARDID, LAST_WARDID, INTIME
(10) INPUTEVENTS_CV.csv: ROW_ID, SUBJECT_ID, HADM_ID, ICUSTAY_ID, CHARTTIME, ITEMID, AMOUNT
(11) LABEVENTS.csv: ROW_ID, SUBJECT_ID, HADM_ID, ITEMID, CHARTTIME, VALUENUM, VALUEUOM
(12) MICROBIOLOGYEVENTS.csv: RROW_ID, SUBJECT_ID, HADM_ID, CHARTTIME, SPEC_TYPE_DESC, ORG_NAME
(13) OUTPUTEVENTS.csv: ROW_ID, SUBJECT_ID, HADM_ID, ICUSTAY_ID, CHARTTIME, ITEMID, VALUE
(14) PATIENTS.csv: ROW_ID, SUBJECT_ID, GENDER, DOB, DOD
(15) PRESCRIPTIONS.csv: ROW_ID, SUBJECT_ID, HADM_ID, STARTDATE, ENDDATE, DRUG, DOSE_VAL_RX, DOSE_UNIT_RX, ROUTE
(16) PROCEDURES.csv: ROW_ID, SUBJECT_ID, HADM_ID, ICD9_CODE, CHARTTIME
(17) TRANSFERS.csv: ROW_ID, SUBJECT_ID, HADM_ID, ICUSTAY_ID, EVENTTYPE, CAREUNIT, WARDID, INTIME, OUTTIME

All the tabls are saved in the data directory {}.

\end{lstlisting}
\end{tcolorbox}

\subsection{eICU Prompts}
We include prompt details for eICU tasks as follows:
\begin{tcolorbox}[
    colback=RoyalPurple!6,          % background color
    colframe=RoyalPurple!48,        % border color
    title=\bfseries eICU Prompt -- Main,
    fonttitle=\small\sffamily,
    sharp corners,            % remove rounding if you like
    boxrule=0.8pt,
    top=1mm, bottom=1mm, left=1mm, right=1mm
]
\begin{lstlisting}[style=prompt]
You are a biomedical expert in handling EHR data and answer questions.
Your objective is to solve a coding problem with given EHR data, with the goal of finally give a concrete answer to the question.
Assume you have knowledge of several tables:
(1) Tables are linked by identifiers whose name usually ends 'ID'. For example, PATIENTUNITSTAYID refers to a unique patient, LABID refers to a unique lab test, and ALLERGYID refers to a unique incidence of allergy occurence.
(2) Four tables are related to measurements. First, the lab table contains laboratory measurements of chemicals such as chloride or albumin. Secondly, the intake and output (intakeoutput) table records all fluid-related measurements such as administered normal saline (ns) and urination. Thirdly, the microlab table records measurements of culture of microorganisms. Fourth, the vitalperiod table describes the patients' vitals during their stay.
(3) The remaining tables (allergy, cost, diagnosis, medication, patient and treatment) contain other critical information, and the table names are self-explanatory.

{EHR_tables}
\end{lstlisting}
\end{tcolorbox}

\begin{tcolorbox}[
    colback=RoyalPurple!6,          % background color
    colframe=RoyalPurple!48,        % border color
    title=\bfseries eICU Prompt -- Table Information,
    fonttitle=\small\sffamily,
    sharp corners,            % remove rounding if you like
    boxrule=0.8pt,
    top=1mm, bottom=1mm, left=1mm, right=1mm
]
\begin{lstlisting}[style=prompt]
For different tables, they contain the following information:
(1) allergy.csv: ALLERGYID, PATIENTUNITSTAYID, DRUGNAME, ALLERGYNAME, ALLERGYTIME
(2) cost.csv: COSTID, UNIQUEPID, PATIENTHEALTHSYSTEMSTAYID, EVENTTYPE, EVENTID, CHARGETIME, COST
(3) diagnosis.csv: DIAGNOSISID, PATIENTUNITSTAYID, ICD9CODE, DIAGNOSISNAME, DIAGNOSISTIME
(4) intakeoutput.csv: INTAKEOUTPUTID, PATIENTUNITSTAYID, CELLPATH, CELLLABEL, CELLVALUENUMERIC, INTAKEOUTPUTTIME
(5) lab.csv: LABID, PATIENTUNITSTAYID, LABNAME, LABRESULT, LABRESULTTIME
(6) medication.csv: MEDICATIONID, PATIENTUNITSTAYID, DRUGNAME, DOSAGE, ROUTEADMIN, DRUGSTARTTIME, DRUGSTOPTIME
(7) microlab.csv: MICROLABID, PATIENTUNITSTAYID, CULTURESITE, ORGANISM, CULTURETAKENTIME
(8) patient.csv: PATIENTUNITSTAYID, PATIENTHEALTHSYSTEMSTAYID, GENDER, AGE, ETHNICITY, HOSPITALID, WARDID, ADMISSIONHEIGHT, HOSPITALADMITSOURCE, HOSPITALDISCHARGESTATUS, ADMISSIONWEIGHT, DISCHARGEWEIGHT, UNIQUEPID, HOSPITALADMITTIME, UNITADMITTIME, UNITDISCHARGETIME, HOSPITALDISCHARGETIME
(9) treatment.csv: TREATMENTID, PATIENTUNITSTAYID, TREATMENTNAME, TREATMENTTIME
(10) vitalperiod.csv: VITALPERIODICID, PATIENTUNITSTAYID, TEMPERATURE, SAO2, HEARTRATE, RESPIRATION, SYSTEMICSYSTOLIC, SYSTEMICDIASTOLIC, SYSTEMICMEAN, OBSERVATIONTIME

All the tabls are saved in the data directory {data_directory}.
\end{lstlisting}
\end{tcolorbox}

\subsection{MedCalcBench Prompts}
We include prompt details for MedCalcBench tasks as follows:
\begin{tcolorbox}[
    colback=RoyalPurple!6,          % background color
    colframe=RoyalPurple!48,        % border color
    title=\bfseries MedCalcBench Prompt,
    fonttitle=\small\sffamily,
    sharp corners,            % remove rounding if you like
    boxrule=0.8pt,
    top=1mm, bottom=1mm, left=1mm, right=1mm
]
\begin{lstlisting}[style=prompt]
You work in a hospital, and a common task in your work is to calculate some biological values of your patients. 
To do this, you need to identify from clinical notes what information is relevant, before using your clinical knowledge to calculate.
And then write a Python code to calculate the value.
In the code, please use the variable 'answer' to store the answer of the code.
In the main function, please print the final answer of the code without any other text.
\end{lstlisting}
\end{tcolorbox}

\subsection{TREQS Prompts}
We include prompt details for TREQS tasks as follows:
\begin{tcolorbox}[
    colback=RoyalPurple!6,          % background color
    colframe=RoyalPurple!48,        % border color
    title=\bfseries TREQS Prompt,
    fonttitle=\small\sffamily,
    sharp corners,            % remove rounding if you like
    boxrule=0.8pt,
    top=1mm, bottom=1mm, left=1mm, right=1mm
]
\begin{lstlisting}[style=prompt]
You are an biomedical expert in handling EHR data and answer questions accordingly. 
Your objective is to solve a coding problem with given EHR data, with the goal of finally give a concrete answer to the question.
Assume you have knowledge of several tables:
(1) Tables are linked by identifiers which usually have the suffix 'ID'. For example, SUBJECT_ID refers to a unique patient. HADM_ID refers to a unique admission to the hospital, and ICUSTAY_ID refers to a unique admission to an intensive care unit.
(2) All tables contain SUBJECT_ID (patient identifier) and HADM_ID (hospital admission identifier).
(3) The table names are self-explanatory.

For different tables, they contain the following information:
(1) DEMOGRAPHIC.csv: SUBJECT_ID, HADM_ID, NAME, MARITAL_STATUS, AGE, DOB, GENDER, LANGUAGE, RELIGION, ADMISSION_TYPE, DAYS_STAY, INSURANCE, ETHNICITY, EXPIRE_FLAG, ADMISSION_LOCATION, DISCHARGE_LOCATION, DIAGNOSIS, DOD, DOB_YEAR, DOD_YEAR, ADMITTIME, DISCHTIME, ADMITYEAR
(2) DIAGNOSES.csv: SUBJECT_ID, HADM_ID, ICD9_CODE, SHORT_TITLE, LONG_TITLE
(3) LAB.csv: SUBJECT_ID, HADM_ID, ITEMID, CHARTTIME, FLAG, VALUE_UNIT, LABEL, FLUID, CATEGORY
(4) PRESCRIPTIONS.csv: SUBJECT_ID, HADM_ID, ICUSTAY_ID, DRUG_TYPE, DRUG, FORMULARY_DRUG_CD, ROUTE, DRUG_DOSE
(5) PROCEDURES.csv: SUBJECT_ID, HADM_ID, ICD9_CODE, SHORT_TITLE, LONG_TITLE

All the tabls are saved in the data directory {data_directory}.
\end{lstlisting}
\end{tcolorbox}

\subsection{MedAgentBench Prompts}
We include prompt details for MedAgentBench tasks as follows:
\begin{tcolorbox}[
    colback=RoyalPurple!6,          % background color
    colframe=RoyalPurple!48,        % border color
    title=\bfseries MedAgentBench Prompt -- Part I,
    fonttitle=\small\sffamily,
    sharp corners,            % remove rounding if you like
    boxrule=0.8pt,
    top=1mm, bottom=1mm, left=1mm, right=1mm
]
\begin{lstlisting}[style=prompt]
You are an expert in using FHIR functions to assist medical professionals.
In FHIR, there are a few common HTTP GET or POST requests to interact with the server. The descriptions of requests are listed here: {fhir_function_description}.

You are given a question and a set of possible functions. 
Based on the question, you will need to write a python code to achieve the purpose. 
    1. Write a python script to invoke a GET function of the FHIR server, you MUST put it in the format of\nGET url?param_name1=param_value1&param_name2=param_value2...
    2. Write a python script to invoke a POST function of the FHIR server, you MUST put it in the format of\nPOST url\n[your payload data in JSON format]
    3. If you have got answers for all the questions and finished all the requested tasks, you MUST save the final answers in the format of {answer_format} (make sure the list is JSON loadable.)

\end{lstlisting}
\end{tcolorbox}

\begin{tcolorbox}[
    colback=RoyalPurple!6,          % background color
    colframe=RoyalPurple!48,        % border color
    title=\bfseries MedAgentBench Prompt -- Part II,
    fonttitle=\small\sffamily,
    sharp corners,            % remove rounding if you like
    boxrule=0.8pt,
    top=1mm, bottom=1mm, left=1mm, right=1mm
]
\begin{lstlisting}[style=prompt]    
You SHOULD NOT include any other text in the response.
Please write the python code and use the variable 'answer' to store the answer of the code.
Question: {question}\n. The FHIR server base URL is {fhir_api_base}. Do not directly write the GET and POST requests.
\end{lstlisting}
\end{tcolorbox}

\begin{tcolorbox}[
    colback=RoyalPurple!6,          % background color
    colframe=RoyalPurple!48,        % border color
    title=\bfseries MedAgentBench Prompt -- Answer Format,
    fonttitle=\small\sffamily,
    sharp corners,            % remove rounding if you like
    boxrule=0.8pt,
    top=1mm, bottom=1mm, left=1mm, right=1mm
]
\begin{lstlisting}[style=prompt]
answer = {"GET": ["60","S2874099"], "POST": ["http://localhost:8080/fhir/Observation", "payload]}
The answers to the questions are listed in "GET" instead of the get commands, while the post url and payload are listed in "POST".
\end{lstlisting}
\end{tcolorbox}

\subsection{Biocoder Prompts}
We include prompt details for Biocoder tasks as follows:
\begin{tcolorbox}[
    colback=RoyalPurple!6,          % background color
    colframe=RoyalPurple!48,        % border color
    title=\bfseries Biocoder Prompt,
    fonttitle=\small\sffamily,
    sharp corners,            % remove rounding if you like
    boxrule=0.8pt,
    top=1mm, bottom=1mm, left=1mm, right=1mm
]
\begin{lstlisting}[style=prompt]
You are an biomedical expert in writing bioinformatics code and answer questions accordingly. 
Your objective is to write a python function to solve the given question.
Please only write the function, do not include any other text.

Please write a Python function with the following signature:
{signature}
\end{lstlisting}
\end{tcolorbox}

\subsection{BioDSBench Prompts}
We include prompt details for BioDSBench tasks as follows:
\begin{tcolorbox}[
    colback=RoyalPurple!6,          % background color
    colframe=RoyalPurple!48,        % border color
    title=\bfseries BioDSBench Prompt,
    fonttitle=\small\sffamily,
    sharp corners,            % remove rounding if you like
    boxrule=0.8pt,
    top=1mm, bottom=1mm, left=1mm, right=1mm
]
\begin{lstlisting}[style=prompt]
You are an biomedical expert in writing bioinformatics code and answer questions accordingly. 
Your objective is to write a python code to solve the given question.
Please only write the code, do not include any other text.
All the required data are stored in the directory: 
{dataset_path}
\end{lstlisting}
\end{tcolorbox}

\subsection{EHRShot Prompts}
We include prompt details for EHRShot tasks as follows:
\begin{tcolorbox}[
    colback=RoyalPurple!6,          % background color
    colframe=RoyalPurple!48,        % border color
    title=\bfseries BioDSBench Prompt -- Main,
    fonttitle=\small\sffamily,
    sharp corners,            % remove rounding if you like
    boxrule=0.8pt,
    top=1mm, bottom=1mm, left=1mm, right=1mm
]
\begin{lstlisting}[style=prompt]
You are an biomedical expert in writing machine learning code to solve EHR-relevant tasks.
Your objective is to solve a machine learning task based on the given data, with the goal of maximizing the performance of the model in limited steps.
You must use Machine Learning/Deep Learning methods to solve the problem, the score of random guess or without any ML/DL methods will be canclled finally.
You are likely to train models according to specific task requirements.
You have access to a GPU and several CPUs for training DL/ML models.
Use CUDA and PyTorch for faster training if needed.

Code requirements:
    - Read all data files from data_dir={data_dir}
    - Save all the predictions given by the model to a file named 'predictions-{task_name}.csv' in the './cache/ehrshot/{model}/' directory.
    - Don't add, delete, or modify any files in data_dir
    - Use "print" to output information in the feedback
    - No plotting or visualization is allowed
    - Code should be self-contained and not rely on any variables or state outside
    - Code must be completely runnable, otherwise it will be considered as failed
    - Optimize your Model/Parameters/Data Processing/Algorithm for continuous improvement
    - The prediction file should be a csv file with the following format, where the prediction should be predicted labels instead of predicted probabilities:
patient_id, prediction
115967096, 8192
...

{feature_information}

{label_information}
\end{lstlisting}
\end{tcolorbox}

\begin{tcolorbox}[
    colback=RoyalPurple!6,          % background color
    colframe=RoyalPurple!48,        % border color
    title=\bfseries BioDSBench Prompt -- Feature Information,
    fonttitle=\small\sffamily,
    sharp corners,            % remove rounding if you like
    boxrule=0.8pt,
    top=1mm, bottom=1mm, left=1mm, right=1mm
]
\begin{lstlisting}[style=prompt]
The corresponding features are stored in the following directories:
{feature_directory_train}: training features for the task
{feature_directory_val}: validation features for the task
{feature_directory_test}: test features for the task
Each of the feature files is a dictionary, containing the following keys:
    - data_matrix: the feature vectors of the visits, where each row is a embedded vector, representing a single visit of a patient
    - patient_ids: the identifiers of the patients, where each row is a visit and the corresponding patient id
    - labeling_time: the time of the visit, where each row is a visit and the corresponding time
\end{lstlisting}
\end{tcolorbox}

\begin{tcolorbox}[
    colback=RoyalPurple!6,          % background color
    colframe=RoyalPurple!48,        % border color
    title=\bfseries BioDSBench Prompt -- Label Information,
    fonttitle=\small\sffamily,
    sharp corners,            % remove rounding if you like
    boxrule=0.8pt,
    top=1mm, bottom=1mm, left=1mm, right=1mm
]
\begin{lstlisting}[style=prompt]
The corresponding labels are stored in the following directories:
{label_directory_train}: training labels for the task
{label_directory_val}: validation labels for the task
{label_directory_test}: test labels for the task
Each of the label files contain the following columns:
    - patient_id: the identifier of the patient
    - value: the label value of the patient on the {task_name} task
    - label_type: the type of the label, which can be 'categorical'/'boolean', etc.
    - prediction_time: only the features before this time can be used to predict the label, used in data processing stage
\end{lstlisting}
\end{tcolorbox}

\subsection{EHR-SeqSQL Prompts}
We include prompt details for EHR-SeqSQL tasks as follows:
\begin{tcolorbox}[
    colback=RoyalPurple!6,          % background color
    colframe=RoyalPurple!48,        % border color
    title=\bfseries EHR-SeqSQL Prompt -- Part I,
    fonttitle=\small\sffamily,
    sharp corners,            % remove rounding if you like
    boxrule=0.8pt,
    top=1mm, bottom=1mm, left=1mm, right=1mm
]
\begin{lstlisting}[style=prompt]
You are an biomedical expert in handling EHR data and answer questions accordingly. 
Your objective is to solve a coding problem with given EHR data, with the goal of finally give a concrete answer to the question.
Assume you have knowledge of several tables:
(1) Tables are linked by identifiers which usually have the suffix 'ID'. For example, SUBJECT_ID refers to a unique patient, HADM_ID refers to a unique admission to the hospital, and ICUSTAY_ID refers to a unique admission to an intensive care unit.
(2) Charted events such as notes, laboratory tests, and fluid balance are stored in a series of 'events' tables. For example the outputevents table contains all measurements related to output for a given patient, while the labevents table contains laboratory test results for a patient.
(3) Tables prefixed with 'd_' are dictionary tables and provide definitions for identifiers. For example, every row of chartevents is associated with a single ITEMID which represents the concept measured, but it does not contain the actual name of the measurement. By joining chartevents and d_items on ITEMID, it is possible to identify the concept represented by a given ITEMID.
(4) For the databases, four of them are used to define and track patient stays: admissions, patients, icustays, and transfers. Another four tables are dictionaries for cross-referencing codes against their respective definitions: d_icd_diagnoses, d_icd_procedures, d_items, and d_labitems. The remaining tables, including chartevents, cost, inputevents_cv, labevents, microbiologyevents, outputevents, prescriptions, procedures_icd, contain data associated with patient care, such as physiological measurements, caregiver observations, and billing information.

For different tables, they contain the following information:
(1) ADMISSIONS.csv: ROW_ID, SUBJECT_ID, HADM_ID, ADMITTIME, DISCHTIME, ADMISSION_TYPE, ADMISSION_LOCATION, DISCHARGE_LOCATION, INSURANCE, LANGUAGE, MARITAL_STATUS, ETHNICITY, AGE
(2) CHARTEVENTS.csv: ROW_ID, SUBJECT_ID, HADM_ID, ICUSTAY_ID, ITEMID, CHARTTIME, VALUENUM, VALUEUOM
\end{lstlisting}
\end{tcolorbox}

\begin{tcolorbox}[
    colback=RoyalPurple!6,          % background color
    colframe=RoyalPurple!48,        % border color
    title=\bfseries EHR-SeqSQL Prompt -- Part II,
    fonttitle=\small\sffamily,
    sharp corners,            % remove rounding if you like
    boxrule=0.8pt,
    top=1mm, bottom=1mm, left=1mm, right=1mm
]
\begin{lstlisting}[style=prompt]
(3) COST.csv: ROW_ID, SUBJECT_ID, HADM_ID, EVENT_TYPE, EVENT_ID, CHARGETIME, COST
(4) D_ICD_DIAGNOSES.csv: ROW_ID, ICD9_CODE, SHORT_TITLE, LONG_TITLE
(5) D_ICD_PROCEDURES.csv: ROW_ID, ICD9_CODE, SHORT_TITLE, LONG_TITLE
(6) D_ITEMS.csv: ROW_ID, ITEMID, LABEL, LINKSTO
(7) D_LABITEMS.csv: ROW_ID, ITEMID, LABEL
(8) DIAGNOSES_ICD.csv: ROW_ID, SUBJECT_ID, HADM_ID, ICD9_CODE, CHARTTIME
(9) ICUSTAYS.csv: ROW_ID, SUBJECT_ID, HADM_ID, ICUSTAY_ID, FIRST_CAREUNIT, LAST_CAREUNIT, FIRST_WARDID, LAST_WARDID, INTIME, OUTTIME
(10) INPUTEVENTS_CV.csv: ROW_ID, SUBJECT_ID, HADM_ID, ICUSTAY_ID, CHARTTIME, ITEMID, AMOUNT
(11) LABEVENTS.csv: ROW_ID, SUBJECT_ID, HADM_ID, ITEMID, CHARTTIME, VALUENUM, VALUEUOM
(12) MICROBIOLOGYEVENTS.csv: RROW_ID, SUBJECT_ID, HADM_ID, CHARTTIME, SPEC_TYPE_DESC, ORG_NAME
(13) OUTPUTEVENTS.csv: ROW_ID, SUBJECT_ID, HADM_ID, ICUSTAY_ID, CHARTTIME, ITEMID, VALUE
(14) PATIENTS.csv: ROW_ID, SUBJECT_ID, GENDER, DOB, DOD
(15) PRESCRIPTIONS.csv: ROW_ID, SUBJECT_ID, HADM_ID, STARTDATE, ENDDATE, DRUG, DOSE_VAL_RX, DOSE_UNIT_RX, ROUTE
(16) PROCEDURES.csv: ROW_ID, SUBJECT_ID, HADM_ID, ICD9_CODE, CHARTTIME
(17) TRANSFERS.csv: ROW_ID, SUBJECT_ID, HADM_ID, ICUSTAY_ID, EVENTTYPE, CAREUNIT, WARDID, INTIME, OUTTIME

All the tabls are saved in the data directory {data_directory}.
\end{lstlisting}
\end{tcolorbox}

\subsection{EHRCon Prompts}
We include prompt details for EHRCon tasks as follows:
\begin{tcolorbox}[
    colback=RoyalPurple!6,          % background color
    colframe=RoyalPurple!48,        % border color
    title=\bfseries EHRCon Prompt -- Part I,
    fonttitle=\small\sffamily,
    sharp corners,            % remove rounding if you like
    boxrule=0.8pt,
    top=1mm, bottom=1mm, left=1mm, right=1mm
]
\begin{lstlisting}[style=prompt]
You are an biomedical expert in handling EHR data and answer questions accordingly. 
Your objective is to solve a coding problem with given EHR data, with the goal of finally give a concrete answer to the question.
Assume you have knowledge of several tables:
(1) Tables are linked by identifiers which usually have the suffix 'ID'. For example, SUBJECT_ID refers to a unique patient, HADM_ID refers to a unique admission to the hospital, and ICUSTAY_ID refers to a unique admission to an intensive care unit.
(2) Charted events such as notes, laboratory tests, and fluid balance are stored in a series of 'events' tables. For example the outputevents table contains all measurements related to output for a given patient, while the labevents table contains laboratory test results for a patient.
(3) Tables prefixed with 'd_' are dictionary tables and provide definitions for identifiers. For example, every row of chartevents is associated with a single ITEMID which represents the concept measured, but it does not contain the actual name of the measurement. By joining chartevents and d_items on ITEMID, it is possible to identify the concept represented by a given ITEMID.

\end{lstlisting}
\end{tcolorbox}

\begin{tcolorbox}[
    colback=RoyalPurple!6,          % background color
    colframe=RoyalPurple!48,        % border color
    title=\bfseries EHRCon Prompt -- Part II,
    fonttitle=\small\sffamily,
    sharp corners,            % remove rounding if you like
    boxrule=0.8pt,
    top=1mm, bottom=1mm, left=1mm, right=1mm
]
\begin{lstlisting}[style=prompt]
(4) For the databases, four of them are used to define and track patient stays: admissions, patients, icustays, and transfers. Another four tables are dictionaries for cross-referencing codes against their respective definitions: d_icd_diagnoses, d_icd_procedures, d_items, and d_labitems. The remaining tables, including chartevents, cost, inputevents_cv, labevents, microbiologyevents, outputevents, prescriptions, procedures_icd, contain data associated with patient care, such as physiological measurements, caregiver observations, and billing information.

For different tables, they contain the following information:
(1) ADMISSIONS.csv: ROW_ID, SUBJECT_ID, HADM_ID, ADMITTIME, DISCHTIME, ADMISSION_TYPE, ADMISSION_LOCATION, DISCHARGE_LOCATION, INSURANCE, LANGUAGE, MARITAL_STATUS, ETHNICITY, AGE
(2) CHARTEVENTS.csv: ROW_ID, SUBJECT_ID, HADM_ID, ICUSTAY_ID, ITEMID, CHARTTIME, VALUENUM, VALUEUOM
(3) COST.csv: ROW_ID, SUBJECT_ID, HADM_ID, EVENT_TYPE, EVENT_ID, CHARGETIME, COST
(4) D_ICD_DIAGNOSES.csv: ROW_ID, ICD9_CODE, SHORT_TITLE, LONG_TITLE
(5) D_ICD_PROCEDURES.csv: ROW_ID, ICD9_CODE, SHORT_TITLE, LONG_TITLE
(6) D_ITEMS.csv: ROW_ID, ITEMID, LABEL, LINKSTO
(7) D_LABITEMS.csv: ROW_ID, ITEMID, LABEL
(8) DIAGNOSES_ICD.csv: ROW_ID, SUBJECT_ID, HADM_ID, ICD9_CODE, CHARTTIME
(9) ICUSTAYS.csv: ROW_ID, SUBJECT_ID, HADM_ID, ICUSTAY_ID, FIRST_CAREUNIT, LAST_CAREUNIT, FIRST_WARDID, LAST_WARDID, INTIME, OUTTIME
(10) INPUTEVENTS_CV.csv: ROW_ID, SUBJECT_ID, HADM_ID, ICUSTAY_ID, CHARTTIME, ITEMID, AMOUNT
(11) LABEVENTS.csv: ROW_ID, SUBJECT_ID, HADM_ID, ITEMID, CHARTTIME, VALUENUM, VALUEUOM
(12) MICROBIOLOGYEVENTS.csv: RROW_ID, SUBJECT_ID, HADM_ID, CHARTTIME, SPEC_TYPE_DESC, ORG_NAME
(13) OUTPUTEVENTS.csv: ROW_ID, SUBJECT_ID, HADM_ID, ICUSTAY_ID, CHARTTIME, ITEMID, VALUE
(14) PATIENTS.csv: ROW_ID, SUBJECT_ID, GENDER, DOB, DOD
(15) PRESCRIPTIONS.csv: ROW_ID, SUBJECT_ID, HADM_ID, STARTDATE, ENDDATE, DRUG, DOSE_VAL_RX, DOSE_UNIT_RX, ROUTE
(16) PROCEDURES.csv: ROW_ID, SUBJECT_ID, HADM_ID, ICD9_CODE, CHARTTIME
(17) TRANSFERS.csv: ROW_ID, SUBJECT_ID, HADM_ID, ICUSTAY_ID, EVENTTYPE, CAREUNIT, WARDID, INTIME, OUTTIME

All the tables are saved in the a .db file at {db_location}.

In addition, you have access to a csv containing the clinical notes with the matching subject ids and hospital admission ids: ROW_ID, SUBJECT_ID, HADM_ID, CHARTDATE, CHARTTIME, STORETIME, CATEGORY, DESCRIPTION, CGID, ISERROR, TEXT, ADMITTIME

This clinical note csv is at {note_csv}.
\end{lstlisting}
\end{tcolorbox}

\subsection{MIMIC-Extract Prompts}
We include prompt details for MIMIC-EXTRACT tasks as follows:
\begin{tcolorbox}[
    colback=RoyalPurple!6,          % background color
    colframe=RoyalPurple!48,        % border color
    title=\bfseries MIMIC-EXTRACT Prompt -- PART I,
    fonttitle=\small\sffamily,
    sharp corners,            % remove rounding if you like
    boxrule=0.8pt,
    top=1mm, bottom=1mm, left=1mm, right=1mm
]
\begin{lstlisting}[style=prompt]
You are an biomedical expert in writing machine learning code to solve EHR-relevant tasks.
Your objective is to solve a machine learning task based on the given data, with the goal of maximizing the performance of the model in limited steps.
You must use Machine Learning/Deep Learning methods to solve the problem, the score of random guess or without any ML/DL methods will be canceled finally.
You are likely to train models according to specific task requirements.
You have access to a GPU and several CPUs for training DL/ML models.
Use CUDA and PyTorch for faster training if needed.

Code requirements:
    - Read all data files from data_dir={data_dir}
    - Save all the predictions given by the model to a file named 'predictions-{task_name}.csv' in the './cache/ehrshot/{model}/' directory.
    - Don't add, delete, or modify any files in data_dir
    - Use "print" to output information in the feedback
    - No plotting or visualization is allowed
    - Code should be self-contained and not rely on any variables or state outside
    - Code must be completely runnable, otherwise it will be considered as failed
    - Optimize your Model/Parameters/Data Processing/Algorithm for continuous improvement
    - The prediction file should be a csv file with the following format, where the prediction should be predicted labels instead of predicted probabilities:

You have the data splits based on hospital admission ids. You are asked to use longitudinal EHR data within each admission instance to predict a two types of tasks:
(1) Classification associated with the entire duration of admission: mortality inside hospital, mortality inside ICU, length of stay beyond 3 days, length of stay beyond 7 days. All 4 are binary classification tasks using lab features only.
For the first task, the output csv should have two columns:
subject_id, prediction
9923, 0
...

(2) Classification associated with hourly measurements: intervention of vasopressor in ICU, and intervention of ventilator in ICU. Use the past 6 hours of lab measurements and static demographics (matching patient id) to predict the 4 intervention statuses during the 4-hour period after 6 hours. 
For the second task, the output csv should have three colums instead:
subject_id, window_idx, prediction
140, 4, 3
...

The corresponding features are stored in the following directories:
{feature_directory_train}: training features for the task
{feature_directory_val}: validation features for the task
{feature_directory_test}: test features for the task
\end{lstlisting}
\end{tcolorbox}

\begin{tcolorbox}[
    colback=RoyalPurple!6,          % background color
    colframe=RoyalPurple!48,        % border color
    title=\bfseries MIMIC-EXTRACT Prompt -- PART II,
    fonttitle=\small\sffamily,
    sharp corners,            % remove rounding if you like
    boxrule=0.8pt,
    top=1mm, bottom=1mm, left=1mm, right=1mm
]
\begin{lstlisting}[style=prompt]
Each of the feature files is a pickled pandas dataframe:
    - subject_id: the unique ID of the subject
    - hadm_id: the unique ID of the hospital admission 
    - icustay_id: the unique ID of the ICU session
    - hours_in: the number of hours since hospital admission. Counting from 0
    - The rest of the columns are organized in groups of three, where the outer level specifies the type of measurements (e.g. alanine aminotransferase and ph urine), and the inner level lists the count, mean and std of the measurements, respectively. The table has been imputed.

{feature_information}

{label_information}
\end{lstlisting}
\end{tcolorbox}

\begin{tcolorbox}[
    colback=RoyalPurple!6,          % background color
    colframe=RoyalPurple!48,        % border color
    title=\bfseries MIMIC-EXTRACT Prompt -- Lab Feature,
    fonttitle=\small\sffamily,
    sharp corners,            % remove rounding if you like
    boxrule=0.8pt,
    top=1mm, bottom=1mm, left=1mm, right=1mm
]
\begin{lstlisting}[style=prompt]
The corresponding features are stored in the following directories:
{feature_directory_train}: training features for the task
{feature_directory_val}: validation features for the task
{feature_directory_test}: test features for the task
Each of the feature files is a pickled pandas dataframe:
    - subject_id: the unique ID of the subject
    - hadm_id: the unique ID of the hospital admission 
    - icustay_id: the unique ID of the ICU session
    - hours_in: the number of hours since hospital admission. Counting from 0
    - The rest of the columns are organized in groups of three, where the outer level specifies the type of measurements (e.g. alanine aminotransferase and ph urine), and the inner level lists the count, mean and std of the measurements, respectively. The table has been imputed.
\end{lstlisting}
\end{tcolorbox}

\begin{tcolorbox}[
    colback=RoyalPurple!6,          % background color
    colframe=RoyalPurple!48,        % border color
    title=\bfseries MIMIC-EXTRACT Prompt -- Static Feature,
    fonttitle=\small\sffamily,
    sharp corners,            % remove rounding if you like
    boxrule=0.8pt,
    top=1mm, bottom=1mm, left=1mm, right=1mm
]
\begin{lstlisting}[style=prompt]
The corresponding features are stored in the following directories:
{feature_directory_train}: demographic training features for the task
{feature_directory_val}: demographic validation features for the task
{feature_directory_test}: demographic test features for the task
Each of the feature files is a pickled pandas dataframe:
    - subject_id: the unique ID of the subject
    - hadm_id: the unique ID of the hospital admission
    - icustay_id: the unique ID of the ICU session
    - intime: the total number of hours in the associated admission
    - gender_F and gender_M: one-hot boolean columns for gender
    - Age 1.0, Age 2.0, Age 3.0, Age 4.0: one-hot boolean columns for ages groups of 10-30, 30-50, 50-70, and >70, respectively
    - Ethnicity columns: one-hot boolean columns for ethnicity (American Indian, Asian, Black, Hispano, Other, White)
    - First care columns: one-hot boolean columns for first admitted care unit (CCU, CSRU, MICU, SICU, TSICU)
\end{lstlisting}
\end{tcolorbox}

\begin{tcolorbox}[
    colback=RoyalPurple!6,          % background color
    colframe=RoyalPurple!48,        % border color
    title=\bfseries MIMIC-EXTRACT Prompt -- Mor Los Label,
    fonttitle=\small\sffamily,
    sharp corners,            % remove rounding if you like
    boxrule=0.8pt,
    top=1mm, bottom=1mm, left=1mm, right=1mm
]
\begin{lstlisting}[style=prompt]
The corresponding labels are stored in the following directories:
{label_directory_train}: training labels for the task
{label_directory_val}: validation labels for the task
{label_directory_test}: test labels for the task
Each of the label csv files contain the following columns:
    - subject_id: the unique ID of the subject
    - hadm_id: the unique ID of the hospital admission
    - mort_icu or mort_hosp or los_3 or los_7: the boolean label for whether the patient died in the ICU, died in hospital, the length of stay exceeding 3 days, and LOS exceeding 7 days, respectively
    - label_type: the type of the label, which can be 'categorical'/'boolean', etc.
\end{lstlisting}
\end{tcolorbox}

\begin{tcolorbox}[
    colback=RoyalPurple!6,          % background color
    colframe=RoyalPurple!48,        % border color
    title=\bfseries MIMIC-EXTRACT Prompt -- Ventilator Vasopressor Label,
    fonttitle=\small\sffamily,
    sharp corners,            % remove rounding if you like
    boxrule=0.8pt,
    top=1mm, bottom=1mm, left=1mm, right=1mm
]
\begin{lstlisting}[style=prompt]
The corresponding labels are stored in the following directories:
{label_directory_train}: training labels for the task
{label_directory_val}: validation labels for the task
{label_directory_test}: test labels for the task
Each of the label csv files contain the following columns:
    - subject_id: the unique ID of the subject
    - 6_hour_window_id: the 6 hour predicted window counted since the patient is admitted to hospital.
    - intervention_category: one of the four scenarios: Label 1 "CONTROL": No intervention throughout the prediction window. Label 2 "ON INTERVENTION": The intervention persists throughout the prediction window. Label 3 "ONSET": Intervention starts within the prediction window. Label 4 "WEAN": Intervention ends within the prediction window.
    - label_type: the type of the label, which can be 'categorical'/'boolean', etc.
\end{lstlisting}
\end{tcolorbox}

\subsection{N-PowerAI Prompts}
We include prompt details for NPowerAI tasks as follows:
\begin{tcolorbox}[
    colback=RoyalPurple!6,          % background color
    colframe=RoyalPurple!48,        % border color
    title=\bfseries NPowerAI Prompt,
    fonttitle=\small\sffamily,
    sharp corners,            % remove rounding if you like
    boxrule=0.8pt,
    top=1mm, bottom=1mm, left=1mm, right=1mm
]
\begin{lstlisting}[style=prompt]
You are a scientist conducting biomedical research and constantly facing statistical problems. Sometimes, you need to find the minimum sample size to achieve a specific power. In other times, you would like to know the statistical power given a population size.
\end{lstlisting}
\end{tcolorbox}